\relax
\documentclass[a4]{article}
\usepackage[a4paper,left=0.5in,right=0.5in,top=0.5in,bottom=0.5in,%
footskip=.25in]{geometry}
\usepackage{titling}
\usepackage{helvet}

\usepackage[hyphens]{url}
\usepackage{graphicx}
\usepackage{caption}
\usepackage[superscript]{cite}
\usepackage{enumitem}%
\usepackage[justification=centering]{caption}

\usepackage{stfloats}
\usepackage[hang,flushmargin, norule]{footmisc}
\usepackage{marvosym}

\usepackage[dvipsnames]{xcolor}

\usepackage{titlesec}
\usepackage{multicol}
\usepackage{cite}
\usepackage{amsmath,amssymb,amsfonts}
\usepackage{algcompatible}
\usepackage{algorithm,algpseudocode}
\usepackage{graphicx}
\usepackage{booktabs, multirow, soul}
\usepackage[flushleft]{threeparttable}
\usepackage{xspace, mathtools, comment}
\usepackage{caption}
\usepackage{subcaption}
\usepackage[resetlabels,labeled]{multibib}
\usepackage{etoolbox}

\usepackage{multicol}
\usepackage{lipsum}
\titlespacing*{\section}{0pt}{1\baselineskip}{0.1\baselineskip}
\titlespacing*{\subsection}{0pt}{1\baselineskip}{0.1\baselineskip}
\titlespacing*{\subsubsection}{0pt}{1\baselineskip}{0.1\baselineskip}

\titleformat*{\section}{\normalfont\sffamily\large\bfseries}
\titleformat*{\subsection}{\normalfont\sffamily\normalsize\bfseries}
\titleformat*{\subsubsection}{\normalfont\sffamily\small\bfseries}

\algnewcommand{\LineComment}[1]{\Statex \(\triangleright\) #1}
\algnewcommand{\ShortLineComment}[1]{\Statex \hspace{1.8em}\(\triangleright\) #1}
\algnewcommand{\ShortShortLineComment}[1]{\Statex \hspace{3.1em}\(\triangleright\) #1}
\algnewcommand{\ShortShortShortLineComment}[1]{\Statex \hspace{4.4em}\(\triangleright\) #1}

\DeclareCaptionLabelSeparator{bar}{\space\textbar\space}

\makeatletter
\renewcommand\@biblabel[1]{#1.}
\makeatother

\makeatletter
\renewenvironment{table*}%
{\renewcommand\familydefault\sfdefault
	\@float{table}}
{\end@float}
\makeatother

\makeatletter
\renewcommand{\scriptsize}{\@setfontsize\scriptsize{7}{7}}
\renewcommand{\footnotesize}{\@setfontsize\footnotesize{8}{9}}
\makeatother

\patchcmd{\thebibliography}
{\settowidth}
{\setlength{\parsep}{0pt}\setlength{\itemsep}{0pt plus 0pt}\settowidth}
{}{}
\apptocmd{\thebibliography}
{\small}
{}{}

\renewcommand{\footnoterule}{%
	\kern -3pt
	\hrule width 0.5\textwidth height 0.5pt
	\kern 2pt
}

\title{A Deep Generative Model for Molecule Optimization via One Fragment Modification}
\date{\vspace{-5ex}}

\author{
	Ziqi Chen\textsuperscript{\rm 1}, Martin Renqiang Min\textsuperscript{\rm 2}, 
	Srinivasan Parthasarathy\textsuperscript{\rm 1,3}, Xia Ning\textsuperscript{\rm 1,3,4 \Letter}\\
}

\newcommand{\Address}{
	\textsuperscript{\rm 1}Computer Science and Engineering, The Ohio State University, Columbus, OH 43210.
	\textsuperscript{\rm 2}Machine Learning Department, NEC Labs America, Princeton, NJ 08540.
	\textsuperscript{\rm 3}Translational Data Analytics Institute, The Ohio State University, Columbus, OH 43210.
	\textsuperscript{\rm 4}Biomedical Informatics, The Ohio State University, Columbus, OH 43210.
	\textsuperscript{\Letter}ning.104@osu.edu
}

\newcommand{\myabstract}[2][1]{%
	\renewcommand\maketitlehookd{
		\Address
		\vspace{10pt}
		\mbox{}\medskip\par
		\centering
		\begin{minipage}{#1\textwidth}
			\textbf{
				{\fontfamily{phv}\selectfont
					#2
				}
			}
		\end{minipage}
	}
}

\newcommand{\etal}{\mbox{\emph{et al.}}\xspace}

\newcommand{\OM}{\mbox{$\mathsf{OM}$}\xspace}

\newcommand{\srone}{\mbox{$\mathsf{OM\text{-}pic}$}\xspace}
\newcommand{\srtwo}{\mbox{$\mathsf{OM\text{-}trn}$}\xspace}
\newcommand{\rate}{\mbox{rate\%}\xspace}

\newcommand{\graph}{\mbox{$\mathcal{G}$}\xspace}
\newcommand{\graphx}{\mbox{$\mathcal{G}_x$}\xspace}
\newcommand{\graphy}{\mbox{$\mathcal{G}_y$}\xspace}
\newcommand{\tree}{\mbox{$\mathcal{T}$}\xspace}
\newcommand{\treex}{\mbox{$\mathcal{T}_x$}\xspace}
\newcommand{\treey}{\mbox{$\mathcal{T}_y$}\xspace}

\newcommand{\atoms}{\mbox{$\mathcal{A}$}\xspace}
\newcommand{\bonds}{\mbox{$\mathcal{B}$}\xspace}
\newcommand{\vertices}{\mbox{$\mathcal{V}$}\xspace}
\newcommand{\verticesx}{\mbox{$\mathcal{V}_x$}\xspace}
\newcommand{\verticesy}{\mbox{$\mathcal{V}_y$}\xspace}
\newcommand{\edges}{\mbox{$\mathcal{E}$}\xspace}

\newcommand{\neigh}{\mbox{$\mathcal{N}$}\xspace}
\newcommand{\latent}{\mbox{$\vect{z}$}\xspace}
\newcommand{\matchSet}{\mbox{$\mathcal{M}$}\xspace}
\newcommand{\deleteSet}{\mbox{$\mathcal{D}$}\xspace}
\newcommand{\addSet}{\mbox{$\mathcal{J}$}\xspace}

\newcommand{\add}{+}
\newcommand{\delete}{-}

\newcommand{\atom}{\mbox{$a$}\xspace}
\newcommand{\atomi}{\mbox{$a_i$}\xspace}
\newcommand{\atomj}{\mbox{$a_j$}\xspace}
\newcommand{\atomEmb}{\mbox{$\mathbf{a}$}\xspace}

\newcommand{\atomEmbj}{\mbox{$\mathbf{a}_j$}\xspace}
\newcommand{\atomEmbp}{\mbox{$\mathbf{a}_p^{*}$}\xspace}
\newcommand{\atomEmbc}{\mbox{$\mathbf{a}_c^{*}$}\xspace}

\newcommand{\bondij}{\mbox{$b_{ij}$}\xspace}
\newcommand{\node}{\mbox{$n$}\xspace}
\newcommand{\nodeu}{\mbox{$n_u$}\xspace}
\newcommand{\nodev}{\mbox{$n_v$}\xspace}
\newcommand{\nodeb}{\mbox{$n_d$}\xspace}
\newcommand{\nodex}{\mbox{$n_x$}\xspace}
\newcommand{\nodey}{\mbox{$n_y$}\xspace}
\newcommand{\nodeEmb}{\mbox{$\mathbf{n}$}\xspace}
\newcommand{\nodeEmbu}{\mbox{$\mathbf{n}_u$}\xspace}
\newcommand{\nodeEmbv}{\mbox{$\mathbf{n}_v$}\xspace}
\newcommand{\nodeEmbx}{\mbox{$\mathbf{n}_x$}\xspace}
\newcommand{\nodeEmby}{\mbox{$\mathbf{n}_y$}\xspace}

\newcommand{\edge}{\mbox{$e$}\xspace}

\newcommand{\attatomp}{\mbox{$a_p^{*}$}\xspace}
\newcommand{\attatomc}{\mbox{$a_c^{*}$}\xspace}

\newcommand{\atype}{\mbox{$\mathbf{{x}}$}\xspace}
\newcommand{\atypei}{\mbox{$\mathbf{{x}}_i$}\xspace}

\newcommand{\btypeij}{\mbox{$\mathbf{{x}}_{ij}$}\xspace}
\newcommand{\ntype}{\atype}

\newcommand{\GMPN}{\mbox{$\mathsf{GMPN}$}\xspace}
\newcommand{\TMPN}{\mbox{$\mathsf{TMPN}$}\xspace}
\newcommand{\mess}{\mbox{$\mathbf{m}$}\xspace}
\newcommand{\mol}{\mbox{$M$}\xspace}
\newcommand{\molx}{\mbox{$M_x$}\xspace}
\newcommand{\moly}{\mbox{$M_y$}\xspace}

\newcommand{\nema}{\mbox{$\vect{s}$}\xspace}
\newcommand{\DE}{\mbox{$\mathsf{DE}$}\xspace}
\newcommand{\encoder}{\mbox{Modof\text{-}encoder}\xspace}
\newcommand{\decoder}{\mbox{Modof\text{-}decoder}\xspace}

\newcommand{\plogp}{\mbox{\text{p}$\log$P}\xspace}
\newcommand{\clogp}{\mbox{Crippen $\log$P}\xspace}
\newcommand{\logp}{\mbox{$\log$P}\xspace}
\newcommand{\prop}{\mbox{property}\xspace}

\newcommand{\drd}{\mbox{\text{DRD2}}\xspace}
\newcommand{\chembl}{\mbox{\text{CheMBL}}\xspace} 
\newcommand{\qed}{\mbox{\text{QED}}\xspace} 
\newcommand{\Fbpp}{\mbox{$f_d$}\xspace}

\newcommand{\Frfp}{\mbox{$f_r$}\xspace}

\newcommand{\Fcp}{\mbox{$f_c$}\xspace}

\newcommand{\Fntp}{\mbox{${f}_l$}\xspace}
\newcommand{\Fpapp}{\mbox{${g}_{p}$}\xspace}

\newcommand{\Fcapp}{\mbox{${g}_{c}$}\xspace}

\newcommand{\molmod}{\mbox{Modof}\xspace}
\newcommand{\pipeline}{\mbox{Modof\text{-}pipe}\xspace}
\newcommand{\pipelineF}{\mbox{Modof\text{-}pipe$^m$}\xspace}
\newcommand{\gcpn}{\mbox{GCPN}\xspace}
\newcommand{\jtvae}{\mbox{JT-VAE}\xspace}
\newcommand{\jtnn}{\mbox{JTNN}\xspace}
\newcommand{\jtnnSTAR}{\mbox{JTNN(m)}\xspace}
\newcommand{\moldqn}{\mbox{MolDQN}\xspace}

\newcommand{\graphaf}{\mbox{GraphAF}\xspace}
\newcommand{\hiergtog}{\mbox{HierG2G}\xspace}
\newcommand{\hiergtogSTAR}{\mbox{HierG2G(m)}\xspace}
\newcommand{\moflow}{\mbox{MoFlow}\xspace}
\newcommand{\gegl}{\mbox{GEGL}\xspace}
\newcommand{\tspolish}{\mbox{T\&S polish}\xspace}

\newcommand{\bpp}{\mbox{$\mathsf{DSP}$}\xspace}
\newcommand{\rfp}{\mbox{$\mathsf{RFP}$}\xspace}
\newcommand{\imr}{\mbox{$\mathsf{IMR}$}\xspace}
\newcommand{\nfa}{\mbox{$\mathsf{NFA}$}\xspace}
\newcommand{\cp}{\mbox{$\mathsf{NFA\text{-}cp}$}\xspace}
\newcommand{\ntp}{\mbox{$\mathsf{NFA\text{-}ntp}$}\xspace}
\newcommand{\app}{\mbox{$\mathsf{NFA\text{-}app}$}\xspace}
\newcommand{\papp}{\mbox{$\mathsf{NFA\text{-}app\text{-}p}$}\xspace}
\newcommand{\capp}{\mbox{$\mathsf{NFA\text{-}app\text{-}c}$}\xspace}

\newcommand{\vect}[1]{\mathbf{#1}}
\newcommand{\matr}[1]{\uppercase{#1}}

\newcommand{\relu}{\mbox{$\text{ReLU}$}\xspace}

\newcommand{\ziqi}[1]{\textcolor{blue}{#1}}
\newcommand{\xia}[1]{\textcolor{red}{#1}}

\begin{document}

\myabstract{Molecule optimization is a critical step in drug development to improve desired properties of drug candidates through chemical modification. 
We developed a novel deep generative model \molmod 
over molecular graphs for molecule optimization.
\molmod modifies a given molecule through the 
prediction of a single site of disconnection at the molecule and the removal and/or addition of fragments
at that site.
A pipeline of multiple, identical {\molmod} models is implemented into \pipeline
to modify an input molecule at multiple disconnection sites. 
Here we show that \pipeline is able to retain major molecular scaffolds, allow controls over 
intermediate optimization steps and better constrain molecule similarities. 
\pipeline outperforms the state-of-the-art methods on benchmark datasets:
without molecular similarity constraints, \pipeline achieves 81.2\% improvement in octanol-water partition coefficient 
penalized by synthetic accessibility and ring size; and 51.2\%, 25.6\% and 9.2\% improvement if the optimized molecules are 
at least 0.2, 0.4 and 0.6 similar to those before optimization, respectively. 
\pipeline is further enhanced into \pipelineF to allow modifying one molecule to multiple optimized ones.
\pipelineF achieves additional performance improvement as at least 17.8\% better than \pipeline. 
}

\maketitle


Molecule optimization is a critical step in drug discovery to improve desired properties of drug 
candidates through chemical modification. 
For example, in lead (molecules showing both activity and selectivity towards a given target) optimization~\cite{jorgensen2009efficient}, 
the chemical structures of the lead molecules can be altered to improve their selectivity and specificity. 
Conventionally, such molecule optimization process is planned based on knowledge and experiences from medicinal chemists, 
and is done via fragment-based screening or 
synthesis~\cite{verdonk2004structure, de2020silico, hoffer2018integrated, gerry2018chemical}. 
Thus, it is not scalable or automated. 
Recent \emph{in silico} approaches using deep learning have enabled alternative computationally 
generative processes to accelerate the conventional paradigm. 
These deep-learning methods learn from string-based molecule representations 
(SMILES)~\cite{Sattarov2019,Sanchez-Lengeling2018} 
or molecular graphs~\cite{jin2018junction, you2018graph}, and generate new ones accordingly 
(e.g., via connecting atoms and bonds) with better properties. 
While computationally attractive, these methods do not conform to the \emph{in vitro} molecule 
optimization process in one very important aspect: molecule optimization needs to retain the major
scaffold of a molecule, but generating entire, new molecular structures may not reproduce the scaffold.  
Therefore, these methods are limited in their potentials to inform and direct \emph{in vitro} molecule optimization.

We propose a novel generative model for molecule optimization that better 
approximates \emph{in silico} chemical modification. 
Our method is referred to \underline{mod}ifier with \underline{o}ne \underline{f}ragment, 
denoted as \molmod. 
Following the idea of fragment-based drug design~\cite{Murray2009, hajduk2007decade}, 
\molmod predicts a single site of disconnection at a molecule, and modifies the molecule by changing the fragments 
(e.g., ring systems, linkers, side chains) at that site. 
Distinctly from existing molecule optimization approaches that encode and decode whole molecular graphs,
\molmod learns from and encodes the difference between molecules before and after optimization at one disconnection site. 
To modify a molecule, \molmod generates only one fragment that instantiates the expected difference 
by decoding a sample drawn from the latent `difference' space.
Then, \molmod removes the original fragment at the disconnection site, and 
attaches the generated fragment at the site. 
By sampling multiple times, \molmod is able to generate multiple optimized candidates. 
A pipeline of multiple, identical \molmod models, denoted as \pipeline, 
is implemented to optimize molecules at multiple disconnection sites through different \molmod models iteratively, 
with the output molecule from one \molmod model as the input to the next. 
\pipeline is further enhanced into \pipelineF to allow modifying one molecule into multiple optimized ones as the 
final output.

\molmod has the following advantages: 
\begin{itemize}
	\item \molmod modifies one fragment at a time. 
	It better approximates the \emph{in vitro} chemical modification and retains the majority of 
	molecular scaffolds. Thus, it potentially better informs and directs \emph{in vitro} molecule optimization. 
	\item \molmod only encodes and decodes the fragment that needs modification and facilitates better modification performance.   
	\item \pipeline modifies multiple fragments at different disconnection sties iteratively.
	It enables easier control over and intuitive deciphering of the intermediate modification steps, 
	and facilitates better interpretability of the entire modification process. 
	\item \molmod is less complex compared to the state of the art (SOTA). 
	It has at least 40\% fewer parameters and uses 26\% less training data. 
	\item \pipeline outperforms the SOTA methods on benchmark datasets
	in optimizing octanol-water partition coefficient penalized by synthetic accessibility and ring size, with 81.2\% improvement 
	without molecular similarity constraints on the optimized molecules, 
	and 51.2\%, 25.6\% and 9.2\% improvement if the optimized molecules need to 
	be at least 0.2, 0.4 and 0.6 similar (in Tanimoto over 2,048-dimension Morgan fingerprints with radius 2) to those before optimization, respectively.
	\item 
	\pipelineF further improves over \pipeline by at least 17.8\%. 
	\item 
	\pipelineF and \pipeline also show superior performance on two other benchmarking tasks optimizing 
	molecule binding affinities against the dopamine D2 receptor, 
	and improving the drug-likeness estimated by quantitative measures.
\end{itemize}


\section*{Related Work}
\label{sec:related}
%
A variety of deep generative models have been developed to generate molecules of desired properties.
These generative models include reinforcement learning (RL)-based models,
generative adversarial networks (GAN)-based models,
flow-based generative models,
and variational autoencoder (VAE)-based models, among others.
Among RL-based models, You \etal~\cite{you2018graph} 
developed a graph convolutional policy network (\gcpn) 
to sequentially add new atoms and corresponding bonds to construct new molecules.
%
%
In the flow-based models, Shi \etal~\cite{shi2020graphaf} developed an autoregressive model  
(\graphaf), in which they learned an invertible mapping between Gaussian distribution 
and molecule structures, and applied RL to fine tune the generation process. 
Zang and Wang~\cite{zang2020} developed a flow-based method (\moflow), in which they utilized bond flow to learn an 
invertible mapping between bond adjacency tensors and Gaussian distribution, and then applied a graph conditional flow to 
generate an atom-type matrix given the bond adjacency tensors.
Variational autoencoder (VAE)-based generative models are also very popular in molecular graph generation.  
%
Jin \etal~\cite{jin2018junction} first decomposed a molecular graph into a junction tree
of chemical substructures, and then used a junction tree VAE (\jtvae) 
to generate and assemble new molecules. 
Jin \etal~\cite{jin2019learning}
developed a junction tree-based encoder-decoder neural model (\jtnn), which
learns a translation mapping between a pair of molecules to 
optimize one into another. 
Jin \etal~\cite{jin2020hierarchical} replaced the small chemical substructures used 
in \jtvae with larger graph motifs, and modified \jtnn
into an autoregressive hierarchical encoder-decoder model (\hiergtog).
%
%
Additional related work including fragment-based VAE~\cite{podda2020},  Teacher and Student polish (\tspolish)~\cite{ji2020graph}, 
scaffold-based VAE~\cite{lim2020} and other genetic algorithm-based methods~\cite{ahn2020guiding, nigam2020} are 
discussed in Section 
S1\footnote{Section references starting with ``S" refer to a Supplementary Information Section.}.

The existing generative methods typically encode the entire molecular graphs, and generate whole, new molecules from
an empty or a randomly selected structure. 
Different from these methods, \molmod learns from and encodes the \emph{difference} between molecules before and after 
optimization. 
Thus, the learning and generative processes is less complex, and are able to retain major molecular scaffolds.
\section*{Problem Definition}
\label{sec:notations:problem}
%
Following Jin \etal~\cite{jin2018junction}, 
we focus on the optimization of the partition coefficients (\logp) measured by \clogp~\cite{logp1999} and penalized by 
synthetic accessibility~\cite{ertl2009estimation} and ring size.
\clogp is a predicted value of experimental \logp using the Wildman and Crippen approach~\cite{logp1999}, 
and has been demonstrated to have a strong correlation (e.g., $r^2$=0.918~\cite{logp1999}) with  
experimental \logp.
Since it is impractical to measure the experimental {\logp} values for a large set of molecules, 
such as our training set (Section S3), 
or for \emph{\mbox{in silico}} generated molecules, using
\clogp will enable the scalable learning from a large set of molecules, and effective yet accurate evaluation 
on \emph{\mbox{in silico}} optimized molecules. 
The combined measurement of \logp, synthetic accessibility (SA) and ring size 
is referred to as \emph{penalized} \logp, denoted as \plogp.
Higher \plogp values indicate higher molecule concentrations in the lipid phase with potentially 
good synthetic accessibility and simple ring structures.
Note that \molmod can be used to optimize other properties as well, with the property of interest used 
instead of \plogp.
Optimizing other properties is discussed in the Section S11. 
Optimizing multiple properties simultaneously is discussed in the
Section S12.
In the rest of this document,  ``property" is by default referred to \plogp.

\textbf{Problem Definition}: 
Given a molecule \molx, molecule optimization aims to modify \molx into another molecule $\moly$ such that 
1) $\moly$ is similar to $\molx$ in its molecular structures (similarity constraint), that is, 
\mbox{$\text{sim}(\molx, \moly)\!\geq\!\delta$} ($\delta$ is a threshold); 
and 
2) $\moly$ is than better than \molx in the property of interest
(e.g., $\plogp(\moly) > \plogp(\molx)$) (property constraint). 
%

\section*{Materials}
\label{sec:expriments}

\subsection*{Data}
\label{sec:experiments:data}
%
We used the benchmark training dataset provided by Jin \etal~\cite{jin2020hierarchical}. 
This dataset was extracted from ZINC dataset~\cite{sterling2015zinc,gomez2018automatic} and contains 75K pairs of molecules. 
Every two paired molecules are similar in their molecule structures 
but different in their \plogp values. 
Using DF-GED~\cite{abu2015exact} algorithm, we then extracted 55,686 pairs of molecules 
from Jin's training dataset such that each extracted pair has only one disconnection site. 
That is, our training data is 26\% less than that in Jin's.  
We used these extracted pairs of molecules
(104,708 unique molecules) as our training data. 
Details about the training data generation are discussed as follows. 
We used Jin's validation set for parameter tuning 
and tested on Jin's test dataset of 800 molecules. 
More details about the training data are available in Section S3.
%

\subsubsection*{Training Data Generation}
\label{app:data}
%
We used a pair of molecules $(\molx, \moly)$ as a training instance in \molmod, where $\molx$ and $\moly$ satisfy 
both the similarity and property constraints, 
and \moly is different from \molx in only one fragment at one disconnection site. 
We constructed such training instances as follows. 
We first quantified the difference between  $\molx$ and $\moly$ using the optimal graph edit distance~\cite{sanfeliu1983} between 
{their junction tree representations} $\treex$ and $\treey$, 
and derived the optimal edit paths to transform $\treex$ to $\treey$.
Such quantification also identified disconnection sites at $\molx$ during its graph comparison.
Details about this process is available in Section S4. 
Identified molecule pairs satisfying similarity and property constraints with only one site of disconnection were used as
training instances. 
For a pair of molecules with a high similarity (e.g., above 0.6), it is very likely that 
they have only one disconnection site as demonstrated in Section S5. 

\subsubsection*{{Molecule Similarity Calculation}}
\label{app:sim}

We used 2,048-dimension binary Morgan fingerprints with radius 2
to represent molecules, and used Tanimoto coefficient to measure molecule similarities.

\subsection*{Baseline Methods}
\label{sec:experiments:baseline}

We compared \molmod with the state-of-the-art baseline methods for the molecule optimization, including 
\jtvae~\cite{jin2018junction}, \gcpn~\cite{you2018graph}, \jtnn~\cite{jin2019learning},
\hiergtog~\cite{jin2020hierarchical}, \graphaf~\cite{shi2020graphaf} and \moflow~\cite{zang2020}. 
\begin{itemize}
\item \jtvae encodes and decodes junction trees, and assembles new, entire molecular graphs based on decoded junction trees.  
\item \gcpn applies a graph convolutional policy network
and iteratively generates molecules by adding atoms and bonds one by one.
\item 
\jtnn learns from molecule pairs and performs molecule optimization as to translate molecular graphs. 
\item 
\hiergtog encodes molecular graphs in a hierarchical fashion, and generates new molecules
via generating and connecting structural motifs. 
\item 
\graphaf learns an invertible mapping between a prior distribution and molecular structures, and 
uses reinforcement learning to fine-tune the model for molecule optimization.
\item 
\moflow learns an 
invertible mapping between bond adjacency tensors and Gaussian distribution, and then applies a graph conditional flow to 
generate an atom-type matrix as the representation of a new molecule from the mapping.
\end{itemize}
%
%

\section*{Experimental Results}
\label{sec:results}

\subsection*{Overall Comparison {on {\plogp} Optimization}}
\label{sec:results:overall}
%
\begin{table*}[h] 
	\centering
	\captionsetup{justification=raggedright}
	\captionof{table}{\textbf{Overall Comparison on Optimizing \plogp}}
	\label{tbl:overall}  
	\begin{threeparttable}
		\begin{scriptsize}
			\begin{tabular}{
					@{\hspace{10pt}}l@{\hspace{10pt}}
					@{\hspace{5pt}}c@{\hspace{5pt}}
					@{\hspace{5pt}}c@{\hspace{5pt}}
					@{\hspace{5pt}}c@{\hspace{5pt}}
					@{\hspace{5pt}}c@{\hspace{5pt}}
					@{\hspace{5pt}}c@{\hspace{5pt}}
					@{\hspace{5pt}}c@{\hspace{5pt}}
					@{\hspace{5pt}}c@{\hspace{5pt}}
					@{\hspace{5pt}}c@{\hspace{5pt}}
					@{\hspace{5pt}}c@{\hspace{5pt}}
					@{\hspace{5pt}}c@{\hspace{5pt}}
					@{\hspace{5pt}}c@{\hspace{5pt}}
				}
				\toprule
				\multirow{2}{*}{model} & \multicolumn{2}{c}{$\delta$ $=$ 0.0} &&  \multicolumn{2}{c}{$\delta$ $=$ 0.2} && \multicolumn{2}{c}{$\delta$ $=$ 0.4} && \multicolumn{2}{c}{$\delta$ $=$ 0.6} \\
				\cmidrule(r){2-3} \cmidrule(r){5-6} \cmidrule(r){8-9} \cmidrule(r){11-12}
				& imprv$\pm$std & sim$\pm$std && imprv$\pm$std & sim$\pm$std && imprv$\pm$std & sim$\pm$std && imprv$\pm$std & sim$\pm$std \\ 
				\midrule
				\jtvae        & 1.91$\pm$2.04 & 0.28$\pm$0.15 & & {1.68}$\pm${1.85} & {0.33}$\pm${0.13} & & 0.84$\pm$1.45 & 0.51$\pm$0.10 	&& 0.21$\pm$0.71 & 0.69$\pm$0.06 \\ 
				\gcpn       & 4.20$\pm$1.28 & 0.32$\pm$0.12 & & {4.12}$\pm${1.19} & {0.34}$\pm${0.11} & & 2.49$\pm$1.30 & 0.47$\pm$0.08  && 0.79$\pm$0.63 & 0.68$\pm$0.08 \\ 
				\jtnn         & -                        & - 			      & & - & - && 3.55$\pm$1.54 & {0.46$\pm$0.06}
				&& 2.33$\pm$1.19 & {0.66$\pm$0.05} \\ 
				\hiergtog  & - 			   & -  	 	      & & - & - && 3.98$\pm$1.46 & {0.46$\pm$0.06} 
				&& 2.49$\pm$1.09 & {0.66$\pm$0.05} \\ 
				\graphaf & 2.94$\pm$1.55 & 0.31$\pm$0.15 & & 2.65$\pm$1.29 & 0.35$\pm$0.12 & & 1.62$\pm$1.16 & 0.51$\pm$0.10  && 0.34$\pm$0.46 & 0.69$\pm$0.06  \\
				\moflow  & 2.39$\pm$1.47 & 0.54$\pm$0.22 & & 2.26$\pm$1.37 & 0.59$\pm$0.17 & & 2.04$\pm$1.24 & 0.65$\pm$0.12  && 1.46$\pm$1.09 & 0.71$\pm$0.07  \\
				\pipeline  &  {{7.61}$\pm${2.30}} & {{0.21}$\pm${0.15}} & & {{6.23}$\pm${1.77}} & {{0.34}$\pm${0.12}} & & {5.00$\pm$1.53} & {0.48$\pm$0.09}  & & {2.72$\pm$1.53} & {0.65$\pm$0.05} \\
				\pipelineF  &  {{9.37}$\pm${2.04}} & {{0.12}$\pm${0.08}} & & {{7.58}$\pm${1.65}} & {{0.27}$\pm${0.07}} & & {5.89$\pm$1.57} & {0.46$\pm$0.06}  & & {3.14$\pm$1.77} & {0.65$\pm$0.05} \\
				\bottomrule
			\end{tabular}
			
			\begin{tablenotes}[normal,flushleft]
				\item 
				\!\!Columns represent: ``imprv": the average improvement in \plogp; 
				``std": the standard deviation; ``sim": the similarity between the original molecules \molx and 
				optimized molecules \moly; ``-": not reported in literature. 
				{We calculated ``sim$\pm$std" for {\jtnn} and {\hiergtog} 
					using the optimized molecules provided by {\jtnn} and our reproduced results for {\hiergtog}, respectively}. \par
				\par
			\end{tablenotes}
		\end{scriptsize}
	\end{threeparttable}
\end{table*}

%
Table~\ref{tbl:overall} presents the overall comparison among \pipeline and \pipelineF, both with a maximum of 5 iterations, 
and the baseline methods on \plogp optimization.
Note that \pipelineF outputs 20 optimized molecules as \jtnn and \hiergtog do. 
Following \gcpn, an additional constraint of molecule size is imposed into \pipeline to limit the size of 
optimized molecules to be at most 38.
As \clogp tends to be large on large molecules, 
this additional constraint also prevents \pipeline from improving \logp by simply increasing molecule size.
When there is no similarity constraint ($\delta$=0), that is, it is not required to produce similar molecules out of
the optimization, \pipeline is able to generate highly optimized molecules with substantially better \plogp improvement (7.61$\pm$2.30), 
with 81.2\%
improvement from the best baseline \gcpn (4.20$\pm$1.28), although with lower similarities between the molecules 
before and after the optimization. 
\pipelineF achieves even better performance with \plogp improvement 9.37$\pm$2.04, that is, 123.1\%
better than \gcpn.
When the similarity constraint takes effect (e.g., $\delta$=0.2, $0.4$ and $0.6$), \pipeline consistently produces 
molecules that are both similar to those before optimization and also with better properties. 
At $\delta$=0.2, 0.4 and 0.6,
\pipeline achieves better property improvement 
(6.23$\pm$1.77, 5.00$\pm$1.53 and 2.72$\pm$1.68, respectively)
than all the best baselines 
(\gcpn with 4.12$\pm$1.19 at $\delta$=0.2, \hiergtog with 3.98$\pm$1.47 at $\delta$=0.4 and 2.49$\pm$1.09 at $\delta$=0.6), 
with 51.2\%, 25.6\% and 9.2\% improvement over the baselines, respectively, 
though the baselines generate more similar molecules than \pipeline; 
\pipelineF achieves the best performance on property improvement 
(7.58$\pm$1.65, 5.89$\pm$1.57, 3.14$\pm$1.77, respectively) with 84.0\%, 48.0\% and 26.1\% improvement over the 
best baselines, respectively.

When $\delta$ is large, we could observe that \jtnn and \hiergtog tend to decode more aromatic rings, 
leading to large molecules with over-estimated
similarities. Instead, \molmod tends to stop if there are many aromatic rings, and thus, 
produces more drug-like molecules~\cite{lipinski2004lead, Ghose1999}. 
Issues related to similarity calculation that will affect optimization performance are discussed
in Section S7. 
Still, the overall comparison demonstrates that \pipeline and \pipelineF
outperform or at least achieve similar performance as the state-of-the-art methods. 

It is worth noting that our performance is reported on the exact benchmark test set.
In our study, we observed some issues of unfair comparison in the existing baseline methods. 
For example, some baseline methods compared and reported results on 
a different test set rather than the benchmark test set. 
%
Some reinforcement learning methods used the test molecules to either directly train a model or 
fine-tune a pre-trained model to optimize the test molecules, which may lead to artificially high performance~\cite{whiteson2011,zhang2018}. 
%
%
Detailed discussions on comparison fairness are available in 
Section S8.

Additional experimental results are available in Section S9, 
such as overall \pipeline performance, 
transformation over chemical spaces, and retaining of molecule scaffolds. 
Specifically, we compared model complexities (Section S9.7), 
which shows that \molmod uses at least 40\% fewer parameters and 26\% less training data 
but outperforms or achieves very comparable results as these state-of-the-art baselines. 
For reproducibility purposes, detailed parameters are reported in Section S9.8.

\begin{figure}[!h]
	\centering
	\captionsetup{justification=raggedright}
	\includegraphics[width=.9\textwidth]{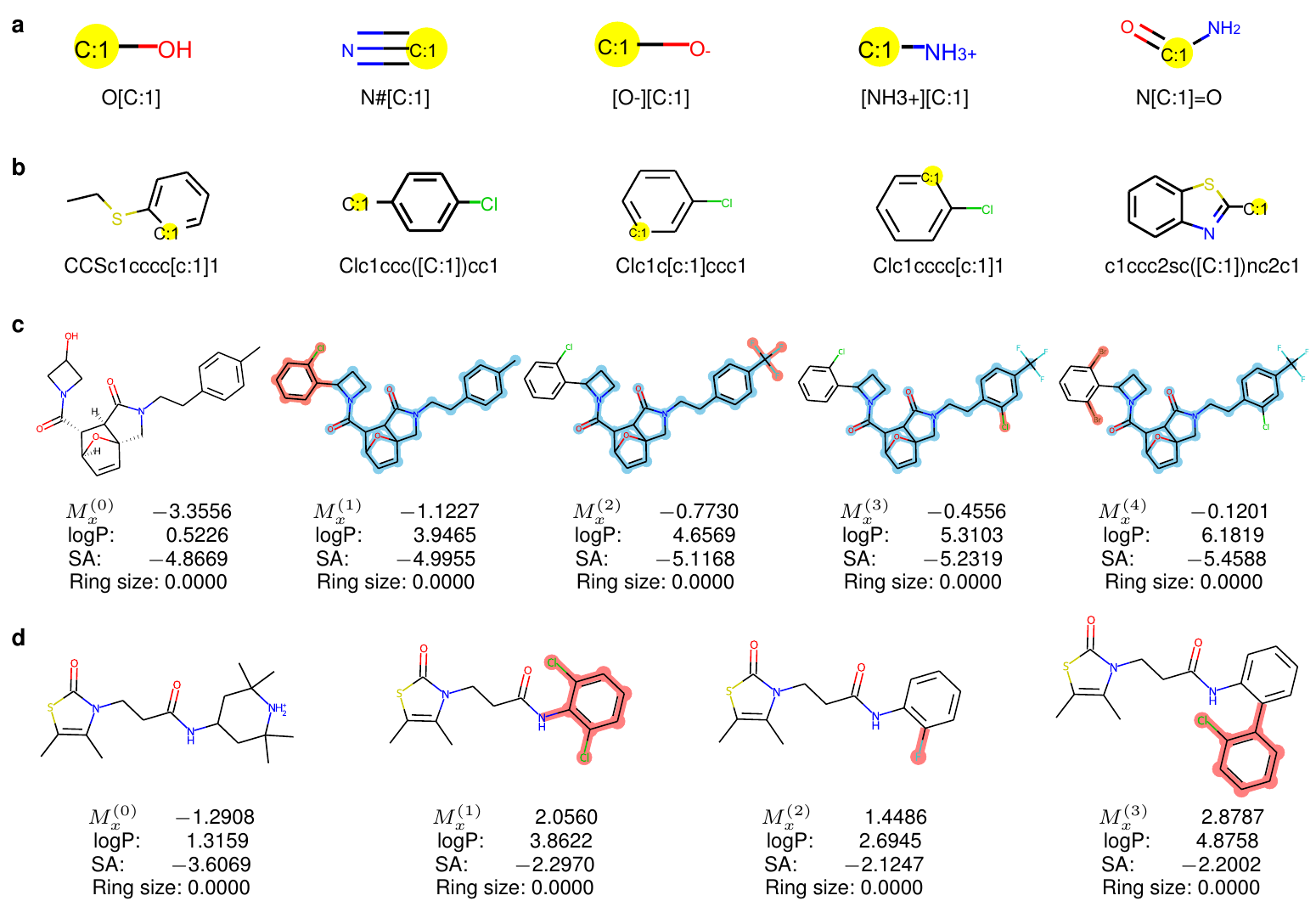}
	\caption{\textbf{\pipeline Examples for \plogp Optimization.} 
		\textbf{a,} Visualization of popular removal fragments.
		\textbf{b,} Visualization of popular attaching fragments.
		\textbf{c,} \pipeline optimization example with multiple disconnection
		sites and multiple \molmod iterations. 
		\textbf{d,} Local optimization.}
	\label{fig:example_logp}
\end{figure}

\subsection*{Case Study}
\label{sec:results:case}

Among training molecules, the top-5 most popular fragments that have been removed from \molx are 
presented in Fig.~\ref{fig:example_logp}{a} with their canonical SMILE strings; 
the top-5 most popular fragments to be attached to generate \moly are
presented in Fig.~\ref{fig:example_logp}{b}.
Overall, the removal fragments in training data are on average of 2.85 atoms and the new attached fragments 
are of 7.55 atoms, that is, the optimization is typically done via removing small fragments and then 
attaching larger fragments.

Fig.~\ref{fig:example_logp}{c} presents an example of molecule \molx (i.e., $\mol^{(0)}_x$)
 being optimized via four iterations in \pipeline 
into another molecule $\mol^{(4)}_x$ under \mbox{$\delta$=0.4}.
At each iteration, only one, small fragment (highlighted in red in 
the figure) is modified from its input, and \plogp value (below each molecule) is improved. 
In the first iteration, $\mol^{(1)}_x$ is modified from $\mol^{(0)}_x$ via the removal of the 
hydroxyl group in {$\mol^{(0)}_x$} and the addition of the 2-chlorophenyl group.
The hydroxyl group is polar and tends to increase water solubility of the molecules, while the 2-chlorophenyl 
group is non-polar and thus more hydrophobic. 
In addition, the increase in molecular weight brought by the chlorophenyl substituent would contribute to the lower water solubility as well.
Thus, the modification from the hydroxyl group to the chlorophenyl group induces the \logp increase (from $0.5226$ to $3.9465$). 
Meanwhile, the introduction of the 2-chlorophenyl group to the cyclobutyl group adds complexity to the synthesis, 
in addition to possible steric effects due to the \textit{ortho}-substitution on the aromatic ring, and induces a 
decrease in synthetic accessibility (SA) (from $-4.8669$ to $-4.9955$).
In the second iteration, the methyl group in $\mol^{(1)}_x$ is replaced by a trifluoromethyl group. 
The trifluoromethyl group is more hydrophobic than the methyl group, and thus increases the {\logp} value 
of $\mol^{(2)}_x$ over $\mol^{(1)}_x$ (from $3.9465$ to $4.6569$). Meanwhile, the slightly larger molecule $\mol^{(2)}_x$ 
has slightly worse SA (from $-4.9955$ to $-5.1168$).
If \logp is preferred to be lower than 5 as proposed in the Lipinski's Rule of Five~\cite{Lipinski2001},
\pipeline can be stopped at this iteration; otherwise,  
in the following two iterations, more halogens are added to the aromatic ring, 
which could make the aromatic ring less polar and 
further decrease water solubility and increase \logp values~\cite{Rokitskaya2019}. 
These four iterations highlight the interpretability of \pipeline corresponding to chemical knowledge. Please note
that all the modifications in \molmod are learned in an end-to-end fashion from data 
without any chemical rules or templates imposed \emph{a priori}, emphasizing the power of \molmod
in learning from molecules. 

In Fig.~\ref{fig:example_logp}{c}, the molecule similarities between $\mol^{(t)}_x$ (\mbox{$t$=1,..,4}) and $\mol^{(0)}_x$ are 
0.630, 0.506, 0.421, 0.4111, respectively. 
This example also shows that \molmod is able to retain the major scaffold of a molecule and optimizes
at different disconnection sites during the iterative optimization process.
Additional analysis on fragments is available in Section S10.

\begin{figure}[!h]
	\centering
	\captionsetup{justification=raggedright}
	\includegraphics[width=.9\textwidth]{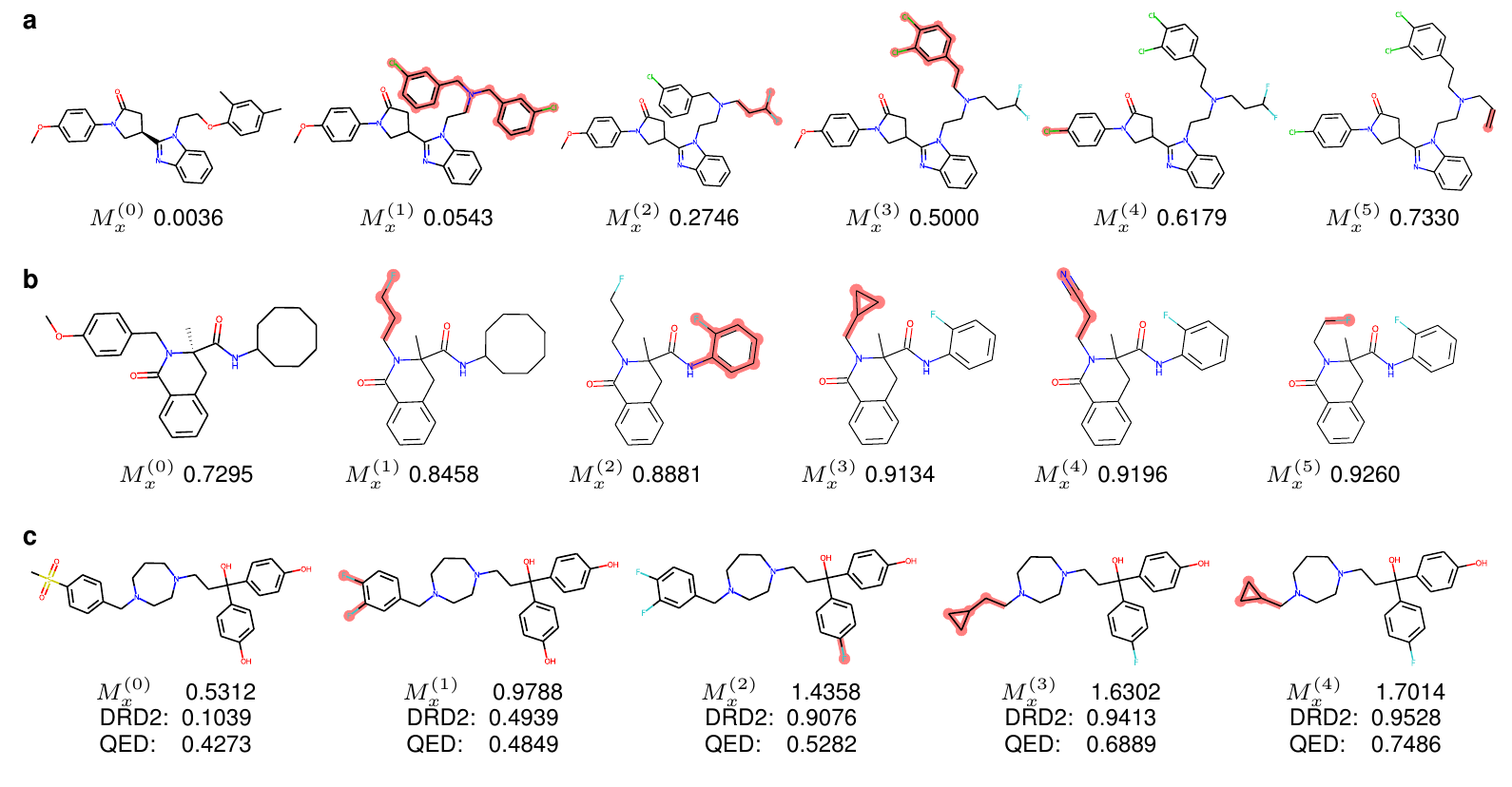}
	\caption{\textbf{\pipeline examples for \drd, \qed and multi-property optimization.}
		\textbf{a,} \pipeline examples for \drd optimization.
		\textbf{b,} \pipeline examples for \qed optimization.
		\textbf{c,} \pipeline examples for multi-property optimization of \drd and \qed.
	}
	\label{fig:example_other}
\end{figure}

\begin{center}
	\begin{table*}[h]
		\centering
		\captionof{table}{\textbf{Overall Comparison on Optimizing \drd and \qed}}
		\label{tbl:drd_qed}
		\begin{small}
			\begin{threeparttable}
				\begin{scriptsize}
					\begin{tabular}{
							@{\hspace{0pt}}l@{\hspace{3pt}}
							@{\hspace{2pt}}c@{\hspace{2pt}}
							@{\hspace{2pt}}c@{\hspace{0pt}}
							@{\hspace{-1pt}}c@{\hspace{2pt}}						
							@{\hspace{0pt}}c@{\hspace{0pt}}
							@{\hspace{2pt}}c@{\hspace{2pt}}
							@{\hspace{2pt}}c@{\hspace{2pt}}
							@{\hspace{2pt}}c@{\hspace{2pt}}						
							@{\hspace{3pt}}c@{\hspace{3pt}}
							@{\hspace{2pt}}c@{\hspace{2pt}}
							@{\hspace{2pt}}c@{\hspace{0pt}}
							@{\hspace{-1pt}}c@{\hspace{2pt}}						
							@{\hspace{0pt}}c@{\hspace{0pt}}
							@{\hspace{2pt}}c@{\hspace{2pt}}
							@{\hspace{2pt}}c@{\hspace{2pt}}
							@{\hspace{2pt}}c@{\hspace{2pt}}						
						}
						\toprule
						& \multicolumn{7}{c}{Optimizing \drd} && \multicolumn{7}{c}{Optimizing \qed} \\
						\cmidrule(r){2-8} \cmidrule(r){10-16}
						\multirow{2}{*}{model} &  \multicolumn{3}{c}{\srone($\drd(\moly)\geq0.5$)} && \multicolumn{3}{c}{\srtwo(imprv $\geq$ 0.2)} &
						&  \multicolumn{3}{c}{\srone($\qed(\moly)\geq$0.9)} && \multicolumn{3}{c}{\srtwo(imprv$\geq$0.1)}\\
						\cmidrule(r){2-4} \cmidrule(r){6-8}\cmidrule(r){10-12}\cmidrule(r){14-16} 
						& rate\% & imprv$\pm$std & sim$\pm$std && rate\% & imprv$\pm$std & sim$\pm$std & 
						& rate\% & imprv$\pm$std & sim$\pm$std && rate\% & imprv$\pm$std & sim$\pm$std\\ 
						\midrule
						
						\jtnn                & 78.10 & 0.83$\pm$0.17 & 0.44$\pm$0.05 && 78.30 & 0.83$\pm$0.17 & 0.44$\pm$0.05 & 
						& 60.50 & 0.17$\pm$0.03 & 0.47$\pm$0.06 && 67.38 & 0.17$\pm$0.03 & 0.47$\pm$0.07 \\
						
						\hiergtog          & 82.00 & 0.83$\pm$0.16 & 0.44$\pm$0.05 && 84.00 & 0.82$\pm$0.18 & 0.44$\pm$0.05 &
						& \textbf{75.12} & 0.18$\pm$0.03 & 0.46$\pm$0.06 && 82.38 & 0.17$\pm$0.03 & 0.46$\pm$0.06 \\	
						\cmidrule(r){2-4} \cmidrule(r){6-8}\cmidrule(r){10-12}\cmidrule(r){14-16} 
						\jtnnSTAR        & 43.50 & 0.77$\pm$0.15 & 0.49$\pm$0.08 && 61.60 & 0.65$\pm$0.24 & 0.49$\pm$0.08 & 
						& 40.50 & 0.17$\pm$0.03 & 0.54$\pm$0.09 && 68.50 & 0.15$\pm$0.03 & 0.54$\pm$0.09  \\
						
						\hiergtogSTAR & 51.80 & 0.78$\pm$0.15 & 0.49$\pm$0.08 && 70.20 & 0.66$\pm$0.24 & 0.49$\pm$0.08 & 
						& 37.12 & 0.17$\pm$0.03 & 0.52$\pm$0.09 && 65.88 & 0.15$\pm$0.03 & 0.53$\pm$0.10\\
						
						\pipeline           & 74.90 & 0.83$\pm$0.14 & 0.48$\pm$0.07 && 89.00 & 0.75$\pm$0.22 & 0.48$\pm$0.07 & 
						& 40.00 & 0.17$\pm$0.03 & 0.51$\pm$0.08 && 70.00 & 0.16$\pm$0.03 & 0.51$\pm$0.08 \\
						
						\pipelineF         & \textbf{88.60} & 0.88$\pm$0.12 & 0.46$\pm$0.05 && \bf{95.90} & 0.84$\pm$0.18 & 0.46$\pm$0.05 & 
						& 66.25 & 0.18$\pm$0.03 & 0.48$\pm$0.07 && \textbf{87.62} & 0.17$\pm$0.03 & 0.48$\pm$0.07  \\
						
						\bottomrule
					\end{tabular}
					
					\begin{tablenotes}[normal,flushleft]		
						\item
						\!Columns represent: 
						\srone: the optimized molecules that achieve a certain property improvement: 
						(1) for {\drd}, the optimized molecules {\moly} should have {\drd} score no less than 0.5; 
						(2) for {\qed}, the optimized molecules {\moly} should have {\qed} score no less than 0.9.
						\srtwo: the optimized molecules that 	achieve a property improvement in a similar 
						degree as in training data:
						(1) for {\drd}, the optimized molecules {\moly} should satisfy \mbox{$\drd(\moly)-\drd(\molx)\geq0.2$};
						(2) for {\qed}, the optimized molecules {\moly} should satisfy {\qed} scores $\qed(\moly)-\qed(\molx)\geq0.1$.
						``{\rate}": the percentage of optimized molecules in each group ({\OM}, {\srone}, {\srtwo})
						over all test molecules;
						``imprv": the average property improvement;  
						``std": the standard deviation; ``sim": the similarity between the original molecules {\molx} and 
						optimized molecules {\moly}. 
						Best {\rate} values are in \bf{bold}. 
						\par
					\end{tablenotes}
				\end{scriptsize}
			\end{threeparttable}
		\end{small}	
	\end{table*}
\end{center}

\subsection*{Performance on \drd and \qed Optimization}
\label{sec:results:drdqed}
%
In addition to improving \plogp, another two popular benchmarking tasks for molecule optimization 
include improving molecule binding affinities against the dopamine D2 receptor (\drd), 
and improving the drug-likeness estimated by quantitative measures (\qed)~\cite{bickerton2012}.
Specifically, given a molecule that doesn't bind well to the \drd receptor (e.g., with low binding affinities), 
the objective of optimizing \drd property is to modify the molecule into another one that will better bind to \drd. 
In the \qed task, given a molecule that is not much drug-like, the objective of optimizing 
\qed property is to modify this molecule into a more ``drug-like" molecule.
Table~\ref{tbl:drd_qed} presents the major results in success rates, property improvement and similarity comparison
under the similarity constraint $\delta$=0.4.
The results demonstrate that \pipelineF significantly outperforms or is comparable to the baseline methods in optimizing \drd and 
\qed, when the success rates are measured using either the benchmark metrics~\cite{jin2019learning,jin2020hierarchical} (\srone in Table~\ref{tbl:drd_qed}) 
or based on training data (\srtwo in Table~\ref{tbl:drd_qed}). 
Fig.~\ref{fig:example_other}{a} and Fig.~\ref{fig:example_other}{b} present
two examples of molecule optimization for \drd and \qed property improvement.  
Particularly, as in Fig.~\ref{fig:example_other}{b}, in the first iteration, 
a 4-methoxyphenyl group is removed and a small chain of 2-fluoroethyl group is added, and thus, 
the number of aromatic rings and the number of hydrogen bond acceptors are reduced, which 
makes the compound more drug-like than its predecessor.
In the second iteration, a cyclooctyl group is removed from $\molx^{(1)}$ and a 2-fluorophenyl group 
is added. This modification may induce reduced flexibility  -- another preferred property of a successful drug. 
In the following iterations, some commonly used fragments in drug design are used to further modify the molecule 
into more drug-likeness. 
Note that, again, \qed optimization is completely learned from data in an end-to-end fashion without any medicinal chemistry knowledge 
imposed by experts. 
The meaningful optimization in the example in Fig.~\ref{fig:example_other}{b} demonstrates the interpretability of \pipeline. 
More details about these two optimization tasks and results are available in the Section S11.

We also conducted 
experiments to optimize both \drd and \qed properties of molecules simultaneously, that is, a multi-property optimization task. 
%
%
Details on this multi-property task and results are available in Section S12. 
Fig.~\ref{fig:example_other}{c} presents an example of multi-property molecule optimization, 
in which both the \drd and \qed scores of the molecule are consistently increased with the iterations of optimization.

\section*{Discussions and Conclusions}
\label{sec:discussion}
\subsection*{Molecule Optimization using Simulated Properties}
\label{sec:discussion:simulate}
%

Most of the molecule properties considered in our experiments 
are based on simulated or predicted values rather than experimentally measured. 
That is, an independent simulation or machine learning model
is first used to generate the property values for the benchmark dataset.  
For example, \clogp is estimated via the Wildman and Crippen approach~\cite{logp1999}; 
synthesis accessibility is calculated using a scoring function over predefined fragments~\cite{ertl2009estimation}; 
the \drd property is predicted using a support vector machine classifier~\cite{olivecrona2017}; 
and the \qed property is predicted using a non-linear classifier combining multiple desirability functions of 
molecular properties~\cite{bickerton2012}. 
While all the existing generative models for molecule optimization~\cite{kusner2017grammar,you2018graph,de2018molgan,jin2018junction, zhou2019optimization,jin2019learning,zang2020,jin2020hierarchical,podda2020,ahn2020guiding} use such simulated 
properties, there are both challenges and opportunities. 
Challenges arise when the simulation or machine learning models
for those property predictions are not sufficiently accurate due to various 
reasons (e.g., limited or biased training molecules), the generative models learned from the 
inaccurate property values would also be inaccurate or incorrect, resulting in generated molecules 
that could negatively impact the downstream drug development tasks significantly. 
However, the opportunities due to the property simulation or prediction can be immense in fully 
unleashing the power of large-scale, data-driven learning paradigms to stimulate drug development as we continue to 
improve these simulations and predictions. 
Specifically, most deep learning-based models for drug development purposes, 
many of which have been demonstrated to be very promising~\cite{Wainberg2018}, 
are not possible without large-scale training data. 
While it is impractical, if ever possible, to experimentally measure the 
interested properties for a large set of molecules (e.g., more than 100K molecules as in our benchmark training data), 
the property 
simulation or prediction of the molecules enables large training data and 
makes the development of such deep learning methodologies possible. 
Fortunately, property prediction simulations or models have become more accurate (e.g., 98\% accuracy for \drd~\cite{olivecrona2017})
due to the accumulation of experimental measurements~\cite{Kim2020} and the strong 
learning power of innovative computational approaches. 
The accurate property simulation or prediction over large-scale molecule data and the powerful learning 
capability of generative models from such molecule data will together have strong potentials to further advance 
\emph{in silico} drug development.

%
\subsection*{Synthesizability and Retrosynthesis}
Our experiments show that \molmod is also able to improve synthesis accessibility (Section S9.4). 
However, it does not necessarily mean that the generated 
molecules can be easily synthesized. This limitation of \molmod is actually common for almost all the computational 
approaches for molecule generation. 
A recent study shows that many molecules generated via deep learning are not easily synthesizable~\cite{gao2020},
which significantly limits the translational potentials of the generative models in making real impacts in drug development. 
On the other hand, retrosynthesis prediction via deep learning, which aims to identify a feasible synthesis path for a given 
molecule through learning and searching from a large collection of synthesis paths, has been an active 
research area~\cite{segler2018,kishimoto2019}. 
Optimizing molecules towards not only better properties but also better synthesizability, particularly 
with explicit synthesis paths identified simultaneously, could be a highly interesting and challenging future research direction. 
Ultimately, we would like to develop a comprehensive computational framework that could generate synthesizable molecules 
with preferable properties. This would require not only a substantial amount of data to train sophisticated models, but 
also necessary domain knowledge and human experts looped in the learning process. 
\subsection*{\emph{In vitro} Validation}

Testing the \emph{\mbox{in silico}} generated molecules in a laboratory will be needed ultimately to validate the computational
methods. 
While currently most existing computational methods are developed in academic environments and thus cannot be 
easily tested on purchasable or proprietary molecule libraries, or cannot be easily synthesized as we discussed 
earlier, 
a few successful stories~\cite{stoke2020} have demonstrated that powerful computational methods have high potentials to 
truly make new discoveries that can succeed in laboratory validation.
Analogous to this molecule optimization and discovery process using deep learning approaches is from 
\mbox{AlphaFold}~\cite{Jumper2021}, a deep learning method predicting protein folding structures. 
The breakthrough from AlphaFold 
in solving a 50-year-old grand challenge in biology offers a strong evidence showing the tremendous power of modern learning approaches, 
which should not be underestimated. 
Still, collaborations with pharmaceutical industry and \emph{\mbox{in vitro}} test are highly needed to truly translate the computational 
methods into real impact. 
In addition, effective sampling and/or prioritization of generated molecules in order 
to identify a feasible, small set of molecules for small-scale \emph{\mbox{in vitro}} validation could be a practical 
solution; it will require the development of new sampling schemes over molecule subspaces, and/or 
the learning of molecule prioritization~\cite{Liu2017Multi,Liu2017Diff} within the molecule generation process. Meanwhile, 
large-scale \emph{\mbox{in vitro}} validation of \emph{\mbox{in silico}} generated molecules represents 
a challenging but interesting future research direction.


\subsection*{Other Issues in Computational Molecule Optimization}

A limitation of {\pipeline} is that it 
employs a local greedy optimization strategy: in each iteration, 
the input molecules to \molmod will be optimized to the best, and if the optimized molecules 
do not have better properties, they will not go through additional \molmod iterations. 
Detailed discussions on local greedy optimization are available in Section S13.1. 
In addition to partition coefficient,
there are a lot of factors (e.g., toxicity, synthesizability) that need to be considered in order 
to develop a molecule into a drug. 
Discussions on multi-property optimization are available in Section S13.2. 
%
Target-specific molecule optimization are also discussed in Section S13.3. 
The \molmod framework could also be used for compounds or substance property
optimization in other application areas (e.g., melting or boiling points for volatiles).
Related discussions are available in Section S13.4. 
\subsection*{Conclusions}
\label{sec:discussion:future}

\molmod optimizes molecules at one disconnection site at a time by learning 
the difference between molecules before and after optimization. With a much less complex model,
it achieves significantly better or similar performance compared to the states of the art.  
In addition to the limitations and corresponding future research directions that have been discussed above,
another limitation with \molmod is that in \molmod,
the modification happens at the periphery of molecules. 
Although this is very common in \emph{in vitro} lead optimization, 
we are currently investigating how
\molmod can be enhanced to  modify the internal regions of molecules, if needed, 
by learning from proper training data with such regions.
%
%
%
Additionally, we hope to integrate domain-specific knowledge in the \molmod learning process facilitating 
increased explainability in the learning and generative process.

\begin{figure}[!h]
	\centering
	\includegraphics[width=.95\textwidth]{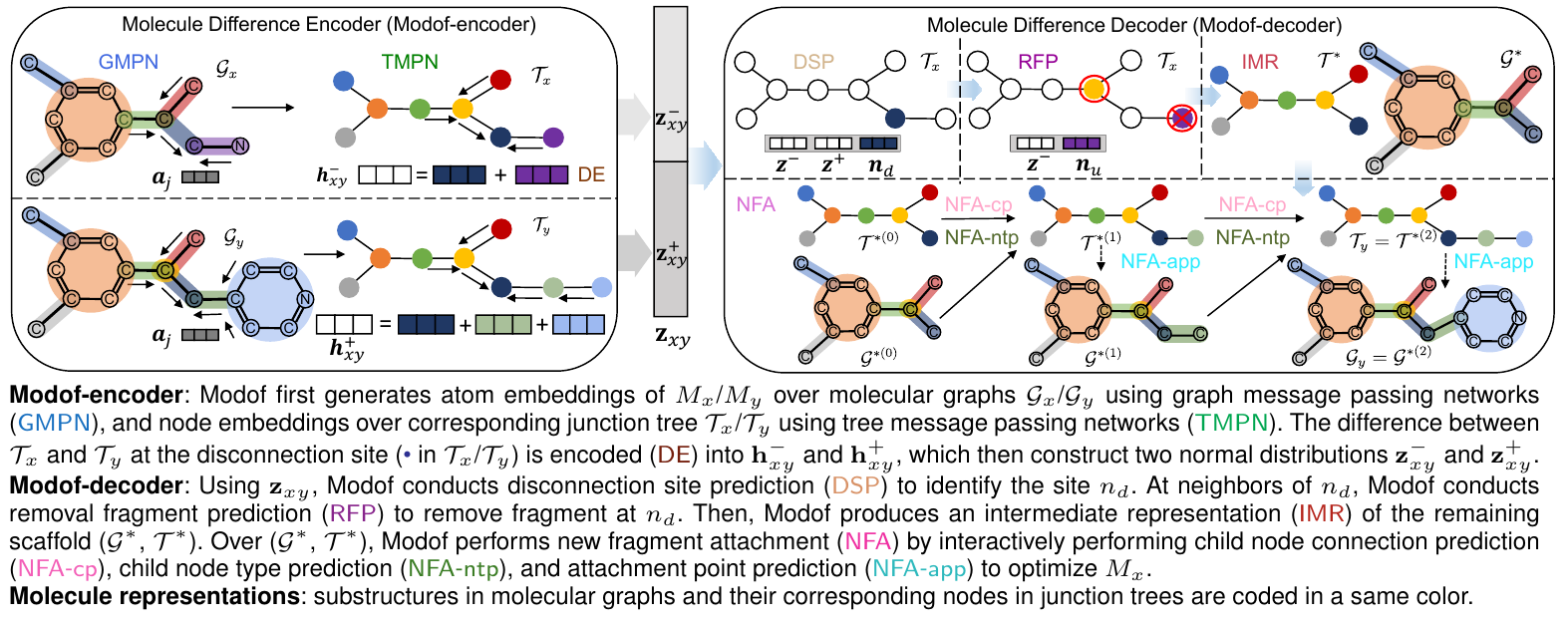}
	\captionsetup{justification=raggedright}
	\caption{\textbf{\molmod Model Overview.}}
	\label{fig:overview}
\end{figure}

\section*{Methods}
\label{sec:method}

\molmod
modifies one fragment (e.g., a ring system, a linker, a side chain) of a molecule at a time, and thus only encodes 
and decodes the fragment that needs modification. 
The site of \mol where the fragment is modified is 
referred to as the site of disconnection
and denoted as \nodeb, which 
corresponds to a node in the junction tree representation ({discussed in "Molecule Representations and Notations"}). 
Fig.~\ref{fig:overview} presents an overview of \molmod. 
All the algorithms are presented in Section S14. 
Discussions on the single-disconnection-site rationale are presented in Section S5. 

\subsection*{Molecule Representations and Notations}
\label{sec:notations:representation}

We represent a molecule $\molx$ using a molecular graph $\graphx$ and a junction tree $\treex$.
$\graphx$ is denoted as \mbox{$\graphx\!=\!(\atoms_x, \bonds_x)$}, where $\atoms_x$ is the set of atoms in $\molx$, and 
$\bonds_x$ is the set of corresponding bonds. 
In the junction tree representation \mbox{$\treex\!=\!(\verticesx,\edges_x)$}~\cite{jin2018junction}, 
all the rings and bonds in $\molx$ are extracted as nodes in $\verticesx$; nodes with common atoms are connected 
with edges in $\edges_x$. 
Thus, each node \mbox{$\node\!\in\!\verticesx$} is a substructure (e.g., a ring, a bond and its connected atoms) in $\graphx$. 
We denote the atoms included in node $\node$ as \mbox{$\atoms_x(\node)$} and refer to the nodes connected 
to \node in $\treex$ as its neighbors, denoted as \mbox{$\neigh_x(\node)$}. 
Thus, each edge \mbox{$(\nodeu,\nodev)\!\in\!\edges_x$} actually corresponds to the common atoms 
\mbox{$\atoms_x(\nodeu)\!\cap\!\atoms_x(\nodev)$} between $\nodeu$ and $\nodev$.
When no ambiguity arises, we will eliminate subscript $x$ in the notations. 
Note that atoms and bonds are the terms used for molecular graph representations, and 
nodes and edges are used for junction tree representations. 
In this manuscript, all the embedding vectors are by default column vectors, represented by lower-case bold letters; all
the matrices are represented by upper-case letters. Key notations are listed in Table~\ref{tbl:notation}.

\begin{center}
	\begin{table*}[!h]
		\caption{\textbf{Notations}}
		\label{tbl:notation}
		\centering
		\begin{threeparttable}
			\begin{scriptsize}
				\begin{tabular}{
						@{\hspace{2pt}}l@{\hspace{2pt}}
						@{\hspace{2pt}}l@{\hspace{2pt}}          
					}
					\toprule
					notation & meaning \\
					\midrule
					$\mol=(\graph, \tree)$              & molecule represented by \graph and  \tree\\
					$\graph=(\atoms, \bonds)$            & molecular graph with atoms \atoms and bonds \bonds \\
					$\tree=(\vertices,\edges)$             & junction tree with nodes \vertices and edges \edges \\
					\atom     & an atom in \graph \\
					\bondij & a bond connecting atoms \atomi and \atomj in \graph \\
					\node 		& a node  in \tree \\
					\mbox{$\edge_{uv}$} &  an edge connecting nodes \nodeu and \nodev in \tree \\
					\nodeb & site of disconnection \\
					$\atoms(\node)$, $\neigh(\node)$ & atoms included in a tree node \node, \node's neighbors \\
					\atype & atom type embedding \\
					$\mess^{(1\cdots t)}$ & concatenation of $\mess^{(1)}$, $\mess^{(2)}$, ... $\mess^{(t)}$ \\
					
					\bottomrule
				\end{tabular}
			\end{scriptsize}
		\end{threeparttable}
		\vspace{-5pt}
	\end{table*}	
\end{center}

\subsection*{\mbox{Molecular Difference Encoder (\encoder)}}
\label{sec:method:encoder}
%
Given two molecules $(\molx, \moly)$, \molmod (Algorithm S1 in Section S14)
learns and encodes the difference between $\molx$ and $\moly$ using 
message passing networks~\cite{gilmer2017neural} over graphs $\graphx$ and $\graphy$, denoted as \GMPN, 
and over junction trees $\treex$ and $\treey$, denoted as \TMPN, via three steps. 
%

\subsubsection*{Step 1. Atom Embedding over Graphs (\GMPN)}
\label{sec:method:encoder:atom}
%
\molmod first represents atoms using embeddings to capture atom types and their local neighborhood structures
by propagating messages along bonds over molecular graphs.
\molmod uses an one-hot encoding $\atypei$ to represent the type of atom $\atomi$, and 
an one-hot encoding $\btypeij$ to represent the type of bond $\bondij$ connecting
$\atomi$ and $\atomj$. 
Each bond $\bondij$ is associated with two messages $\mess_{ij}$ and $\mess_{ji}$ 
encoding the messages propagating from atom $\atomi$ to $\atomj$ and vice versa.
The $\mess_{ij}^{(t)}$  in $t$-th iteration of \GMPN is updated as follows: 
\begin{equation*}
\mess_{ij}^{(t)}=\relu(\matr{W}_1^a\atypei + \matr{W}_2^a\btypeij + 
\matr{W}_3^a\sum_{\mathclap{\scriptsize{\atom_k\in \neigh(\atomi)\setminus \{\atomj\}}}}\mess_{ki}^{(t-1)}),  
\end{equation*}
where $\mess_{ki}^{(0)}$ is initialized as zero, and $\matr{W}_i^a$'s ($i$=1,2,3) are the learnable parameter matrices.
Thus, the message $\mess_{ij}^{(t)}$ encodes the information of  
all length-$t$ paths passing through $\bondij$ to $\atomj$ in the graph.
After $t_a$ iterations of message passing, the atom embedding $\atomEmbj$ is updated as follows:
%
\begin{equation*}
\label{eqn:mol_rep}
\atomEmbj = \relu(\matr{U}_1^a\vect{x}_j 
+ \matr{U}_2^a\sum_{\mathclap{\scriptsize{\atomi\in \neigh(\atomj)}}}\mess_{ij}^{(1 \cdots t_a)}), 
\end{equation*}
\noindent
where $\mess_{ij}^{(1 \cdots t_a)}$ is the 
concatenation of message vectors from all iterations, and 
$\matr{U}_1^a$ and $\matr{U}_2^a$ are learnable parameter matrices.  
Thus, the atom embedding $\atomEmbj$ aggregates information from \atomj's $t_a$-hop neighbors, 
similarly to Xu \etal~\cite{xu2019powerful}, to improve the  atom embedding 
representation power.
%

\subsubsection*{Step 2. Node Embedding over Junction Trees (\TMPN)}
\label{sec:method:encoder:substructure}

\molmod encodes nodes in junction trees into embeddings to capture their local neighborhood structures 
by passing messages along the tree edges. 
To produce rich representations of nodes, 
\molmod first aggregates the information of atoms within a node \nodeu into an embedding $\nema_u$, 
and the information of atoms shared by a tree edge $\edge_{uv}$ into an embedding $\nema_{uv}$
through the following pooling:  
\begin{minipage}{0.45\linewidth}
	\begin{equation}
	\label{eqn:tree_atom}
	\nema_u=\sum_{\mathclap{\scriptsize{\atomi\in \atoms(\nodeu)}}}\atomEmb_i,
	\end{equation} 
\end{minipage}%
\hfill
\begin{minipage}{0.45\linewidth}
	\begin{equation}
	\label{eqn:tree_edge}
	\nema_{uv}=\sum_{\mathclap{\scriptsize{\atomi\in \atoms(\nodeu)\cap\atoms(\nodev)}}}\atomEmb_i. 
	\end{equation}
\end{minipage}
\vskip0.5em
\noindent
\molmod also uses a learnable embedding $\ntype_u$ to represent the type of node \nodeu. 
Thus, $\mess_{uv}^{(t)}$ from node $\nodeu$ to $\nodev$ in $t$-th iteration of \TMPN is updated as follows:
\begin{eqnarray*}
	\begin{aligned}
		\mess_{uv}^{(t)} = \relu( \matr{W}_1^n \relu(\matr{W}_2^n[{\ntype}_u; \nema_u])
		+\matr{W}_3^n\nema_{uv} 
		 +\matr{W}_4^n\sum_{\mathclap{\scriptsize{\node_w\in \neigh(\nodeu)\setminus \{\nodev\}}}}\mess_{wu}^{(t-1)}), 
	\end{aligned}
\end{eqnarray*}
where $[{\ntype}_u; \nema_u]$ is a concatenation of  $\ntype_u$ and $\nema_u$
so as to represent comprehensive node information,
and $\matr{W}_i^n$'s ($i$=1,2, 3,4) are learnable parameter matrices. 
Similarly to the messages in \GMPN, $\mess_{uv}^{(t)}$ encodes the information of all length-$t$ paths 
passing through  edge $\edge_{uv}$ to $\nodev$ in the tree.
After $t_n$ iterations, the node embedding $\nodeEmbv$  is updated as follows:
\begin{equation}
\label{eqn:tree_rep}
\vect{n}_v= \relu(\matr{U}_1^n\relu(\matr{U}_2^n[{\ntype}_v; \nema_v])+
\matr{U}_3^n\sum_{\mathclap{\scriptsize{\nodeu\in \neigh(\nodev)}}}\mess_{uv}^{(1\cdots t_n)}),
\end{equation}
where $\matr{U}_i^n$'s ($i$=1,2,3) are the learnable parameter matrices.
%

\subsubsection*{Step 3. Difference Embedding (\DE)}
\label{sec:method:encoder:difference}

The difference embedding between  $\molx$ and $\moly$ is calculated by pooling the node embeddings 
from $\treex$ and $\treey$ as follows: 

\noindent
\begin{minipage}{0.5\linewidth}
	\begin{equation*}
	\vect{h}_{xy}^{\delete} ~~=~~ \sum_{\mathclap{\scriptsize{\nodex\in \{\verticesx \setminus \verticesy \}\cup \{\nodeb \in \verticesx\}}}}\nodeEmbx,
	\end{equation*} 
\end{minipage}%
\hfill
\begin{minipage}{0.5\linewidth}
	\begin{equation*}
	\vect{h}_{xy}^{\add}  ~~=~~ \sum_{\mathclap{\scriptsize{\nodey\in \{\verticesy \setminus \verticesx\} \cup \{\nodeb \in \verticesy \}}}}\nodeEmby,
	\end{equation*}
\end{minipage}
\vskip0.5em

\noindent
where 
$\nodeEmbx$'s/$\nodeEmby$'s are the embeddings of nodes only appearing in and learned from 
$\treex$/$\treey$ via \TMPN. 
Note that \nodeb in the above equations is the site of disconnection, 
and both \treex and \treey have the common node \nodeb. 
Thus, $\vect{h}_{xy}^{\delete}$ essentially represents the fragment that should be removed 
from $\molx$ at \nodeb 
and 
$\vect{h}_{xy}^{\add}$ represents the fragment that should be attached to $\molx$ at \nodeb afterwards
in order to modify $\molx$ into $\moly$.
We will discuss how to identify \nodeb, and the removed and new attached fragments at \nodeb in $\molx$ and $\moly$
later in Section "Molecular Difference Decoder (\decoder)".

As in VAE \cite{kingma2013auto}, we map the two difference embeddings $\vect{h}_{xy}^{\delete}$ and $\vect{h}_{xy}^{\add}$ 
into two normal distributions by computing the mean and log variance with fully connected layers $\mu(\cdot)$ and $\Sigma(\cdot)$.
We then sample the latent vectors $\latent_{xy}^{\delete}$ and $\latent_{xy}^{\add}$ from these two distributions
and concatenate them into one latent vector $\latent_{xy}$, that is, 

\noindent
\begin{minipage}{0.5\linewidth}
	\begin{equation*}
	\latent_{xy}^{\delete}\sim N(\mu^{\delete}(\vect{h}_{xy}^{\delete}), \Sigma^{\delete}(\vect{h}_{xy}^{\delete})),
	\end{equation*} 
\end{minipage}%
\hfill
\begin{minipage}{0.5\linewidth}
	\begin{equation*}
	\latent_{xy}^{\add}\sim N(\mu^{\add}(\vect{h}_{xy}^{\add}), \Sigma^{\add}(\vect{h}_{xy}^{\add})),
	\end{equation*}
\end{minipage}
\vskip-0.3em
\begin{equation}
\label{eqn:latentz}
\latent_{xy} = [\latent_{xy}^{\delete}; \latent_{xy}^{\add}]. 
\end{equation}
Thus, $\latent_{xy}$ encodes the difference between \molx and \moly. 
%

\subsection*{\mbox{Molecular Difference Decoder (\decoder)}}
\label{sec:method:decoder}
%
Following the autoencoder idea, 
\molmod decodes the difference embedding  $\latent_{xy}$  
(Eqn~\ref{eqn:latentz}) into edit operations that change $\molx$ into $\moly$. 
Specifically, \molmod first predicts a node \nodeb in $\treex$ as the disconnection site. 
This node will split $\treex$ into several fragments, and the number of the resulted fragments 
depends on the number of \nodeb's neighboring nodes \mbox{$\neigh(\nodeb)$}.
\molmod then predicts which fragments to remove from $\molx$, and merges the remaining fragments 
with \nodeb into an intermediate representation \mbox{$\mol^*\!=\!(\graph^*\!,\tree^*)$}.   
After that, \molmod attaches new fragments sequentially starting from \nodeb to $(\graph^*\!,\tree^*)$.
The decoding process (Algorithm S2 in Section S14) has the following 4 steps. 
%

\subsubsection*{Step 1. Disconnection Site Prediction (\bpp)}
\label{sec:method:decoder:bpp}

\molmod predicts a disconnection score for each $\treex$'s node $\nodeu$  as follows,
\begin{equation}
\label{eqn:target}
\Fbpp(\nodeu) = (\vect{w}^d)^{\mathsf{T}}\tanh\vect(\matr{W}^d_1\nodeEmbu+\matr{W}^d_2\vect{\latent})\text{, }\forall \nodeu \in \verticesx,
\end{equation}
where $\nodeEmbu$ is ${\node}_u$'s embedding  (Eqn~\ref{eqn:tree_rep}) in $\treex$, 
$\vect{w}^d$ and $\matr{W}_i^d$'s ($i$=1,2) are learnable parameter vector and 
matrices, respectively. 
The node with the largest disconnection score is predicted as the disconnection site \nodeb. 
Intuitively, \molmod considers the neighboring or local structures of \nodeu (in $\nodeEmbu$) and ``how likely" edit 
operations (represented by \latent)
can be applied at \nodeu. 
To learn \Fbpp, \molmod uses the negative log likelihood of ground-truth disconnection site 
in tree $\treex$ as the loss function.

\subsubsection*{Step 2. Removal Fragment Prediction (\rfp)}
\label{sec:method:decoder:rsp}
Next, \molmod predicts which fragments separated by \nodeb should be removed from \treex. 
For each node \nodeu connected to \nodeb, \molmod predicts a removal score as follows, 
\begin{equation}
\label{eqn:deletion}
\Frfp(\nodeu)=\sigma((\vect{w}^r)^{\mathsf{T}}\relu(\matr{W}^r_1\vect{n}_u+\matr{W}^r_2\latent^{\delete}))\text{, }
\forall \edge_{ud} \in \edges_x,
\end{equation}
where $\sigma(\cdot)$ is sigmoid function, 
$\vect{w}^r$ and $\matr{W}_i^r$'s ($i$=1,2) are learnable parameter vector and 
matrices, respectively. The fragment with a removal score greater than 0.5 is predicted to be removed. 
Thus, there could be multiple or no fragments removed. 
Intuitively, \molmod considers the local structures of the fragment (i.e., $\nodeEmbu$) and ``how likely" this fragment should be removed 
(represented by $\latent^{\delete}$).
To learn \Frfp, \molmod minimizes 
binary cross entropy loss to maximize the predicted scores of ground-truth removed fragments in $\treex$. 
%

\subsubsection*{Step 3. Intermediate Representation (\imr)}
\label{sec:method:decoder:imr}
%
After fragment removal, 
\molmod merges the remaining fragments together with the disconnection site \nodeb into an
intermediate representation \mbox{$\mol^*\!=\!(\graph^*,\tree^*)$}.
$\mol^*$ may not be a valid molecule after some fragments are removed (some bonds are broken).
It represents the scaffold of $\molx$ that should remain unchanged during the optimization. 
\molmod first removes a fragment in order to identify such a scaffold and then adds a fragment to the scaffold to modify the 
molecule. 

\subsubsection*{Step 4. New Fragment Attachment (\nfa)}
\label{sec:method:decoder:nfa}
%
\molmod modifies $\mol^*$ into the optimized $\moly$ 
by attaching a new fragment (Algorithm S3 in Section S14). 
\molmod uses the following four predictors to sequentially attach new nodes to $\tree^*$.  
The predictors will be applied iteratively, starting from \nodeb, on each newly attached node in $\tree^*$. 
The attached 
new node in the $t$-th step is denoted as $\node^{*(t)}$ (\mbox{$\node^{*(0)}\!=\!\nodeb$}), and 
the corresponding molecular graph and tree are denoted as $\graph^{*(t)}$ (\mbox{$\graph^{*(0)}\!=\!\graph^{*}$}) 
and $\tree^{*(t)}$ (\mbox{$\tree^{*(0)}\!=\!\tree^*$}), respectively. 
%

\ul{Step 4.1. Child Connection Prediction (\cp)} 
\label{sec:method:decoder:nfa:cp}
%
\molmod first predicts whether $\node^{*(t)}$ 
should have a new child node attached to it, 
with the probability calculated as follows: 
\begin{equation}
\label{eqn:child}
\Fcp(\node^{*(t)}) = \sigma((\vect{w}^c)^{\mathsf{T}}\relu(\matr{W}^c_1{\nodeEmb}^{*(t)}+\matr{W}^c_2\latent^{\add})), 
\end{equation}
where ${\nodeEmb}^{*(t)}$ is the embedding of node $\node^{*(t)}$ learned over $(\tree^{*(t)},\graph^{*(t)})$ (Eqn~\ref{eqn:tree_rep}), 
$\latent^{\add}$ (Eqn~\ref{eqn:latentz}) indicates ``how much" 
$\tree^{*(t)}$ should be expanded, and  $\vect{w}^c$ and $\matr{W}_i^c$'s ($i$=1,2) are learnable parameter vector and matrices. 
If $\Fcp(\node^{*(t)})$ is above 0.5, \molmod predicts that $\node^{*(t)}$ should have a new child node  
and thus child node type prediction will follow;
otherwise, the optimization process stops at $\node^{*(t)}$. 
To learn \Fcp, \molmod minimizes a binary cross entropy loss to 
maximize the probabilities of ground-truth child nodes. 
Note that $\node^{*(t)}$ may have multiple children, and therefore, once a child is generated as in the following steps 
and attached to $\tree^{*(t)}$, 
another child connection prediction will be conducted at $\node^{*(t)}$
with the updated embedding $\nodeEmb^{*(t)}$ over the expanded 
$(\tree^{*(t)},\graph^{*(t)})$. 
The above process will be iterated until $\node^{*(t)}$ is predicted to have no more children. 

\ul{Step 4.2. Child Node Type Prediction (\ntp)} 
\label{sec:method:decoder:nfa:ntp}
%
The new child node of  $\node^{*(t)}$ is denoted as $\node_c$. 
\molmod predicts the type of $\node_c$ by calculating the probabilities of all types of the nodes that can be 
attached to $\node^{*(t)}$ as follows:  
\begin{equation}
\label{eqn:nodetype}
\Fntp(\node_c) = \text{softmax}(\matr{U}^l\times
\relu(\matr{W}^l_1{\nodeEmb}^{*(t)}+\matr{W}^l_2\latent^{\add})), 
\end{equation}
where $\text{softmax}(\cdot)$ converts a vector of values into probabilities, 
$\matr{U}^l$ and $\matr{W}_i^l$'s ($i$=1,2)
are learnable matrices. 
\molmod assigns the new child $\node_c$ the node type $\ntype_c$
corresponding to the highest probability.
\molmod learns \Fntp by minimizing cross entropy
to maximize the likelihood of true child node types. 

\ul{Step 4.3. Attachment Point Prediction (\app)}
\label{sec:method:decoder:nfa:app}
%
If node $\node^{*(t)}$ is predicted to have a child node $\node_c$, 
the next step is to connect $\node^{*(t)}$ and $\node_c$.  
If $\node^{*(t)}$ and $\node_c$ share one or multiple atoms 
(e.g., $\node^{*(t)}$ and $\node_c$ form a fused ring and thus share two adjacent atoms)
that can be unambiguously determined as the attachment point(s) based on chemical rules, 
\molmod will connect $\node^{*(t)}$ and $\node_c$ via the atom(s).
Otherwise, if $\node^{*(t)}$ and $\node_c$ 
have multiple connection configurations, 
\molmod predicts the attachment atoms at 
$\node^{*(t)}$ and $\node_c$, respectively.

\noindent
\emph{\ul{Step 4.3.1. Attachment Point Prediction at Parent Node (\papp)}}~
\label{sec:method:decoder:nfa:app:parent}
%
\molmod scores each candidate attachment point at parent node $\node^{*(t)}$, denoted as \attatomp, 
as follows, 
%
\begin{equation}
\label{eqn:attach:parent}
\begin{aligned}
\Fpapp(\attatomp)  = (\vect{w}^p)^{\mathsf{T}} \tanh \vect(
\matr{W}_1^p\atomEmbp 
+ \matr{W}_2^p\ntype_c 
 + \matr{W}_3^p\times{\relu({\matr{U}_2^n}[{\ntype}^{*(t)};\tilde{\nema}^{*(t)}])}
+ \matr{W}_4^p{\latent}^{\add}), 
\end{aligned}
\end{equation}
where \mbox{${\atomEmbp}\!=\!\sum_{\scriptsize{\atomi\in \attatomp}}\tilde{\atomEmb}_i$} represents the embedding of \attatomp
(\attatomp could be an atom or a bond), 
$\tilde{\atomEmb}_i$ is calculated by \GMPN over $\graph^{*(t)}$; 
$\matr{U}^n_2$ is as in Eqn~\ref{eqn:tree_rep}; 
$\tilde{\mathbf{\nema}}^{*(t)}$ is the sum of the embeddings of all atoms in $\node^{*(t)}$ (Eqn~\ref{eqn:tree_atom}); 
and $\vect{w}^p$ and $\matr{W}_i^p$ ($i$=1,2,3,4) are learnable vector and matrices. 
\molmod intuitively measures ``how likely'' \attatomp can be attached to $\node_c$ 
by looking at \attatomp its own (i.e., $\atomEmbp$),
its context in $\node^{*(t)}$ (i.e., $\ntype^{*(t)}$ 
and neighbors $\tilde{{\nema}}^{*(t)}$),
its connecting node $\node_c$ (i.e., ${\ntype_c}$) and 
``how much''  $\node^{*(t)}$ should be expanded (represented by $\vect{\latent}^+$). 
The candidate with the highest score is selected as the attachment point in $\node^{*(t)}$. 
\molmod learns \Fpapp by minimizing the  negative log likelihood
of ground-truth attachment points.

\noindent
\emph{\ul{Step 4.3.2. Attachment Point Prediction at Child Node (\capp)}}~
\label{sec:method:decoder:nfa:app:child}
%
\molmod scores each candidate attachment point at the child node $\node_c$, 
denoted as \attatomc,  as follows: 
%
%
\begin{equation}
\label{eqn:attach:child}
\begin{aligned}
\Fcapp(\attatomc)=(\vect{w}^o)^{\mathsf{T}}\tanh \vect(\matr{W}_1^o\atomEmb_c^{*} 
+ \matr{W}_2^o {\ntype}_c
+ \matr{W}_3^o\atomEmbp
+\matr{W}_4^o\vect{\latent}^{\add}),
\end{aligned}                      
\end{equation}
where \mbox{${\atomEmbc}=\sum_{\scriptsize{\atomi \in \attatomc}}\tilde{\atomEmb}_i$} represents the embedding 
of \attatomc (\attatomc could be an atom or a bond) and $\tilde{\atomEmb}_i$ is \atomi's embedding 
calculated over $\node_c$ via \GMPN; $\vect{w}^o$ and $\matr{W}_i^o$'s ($i$=1,2,3,4) are learnable parameters. 
\molmod intuitively measures ``how likely'' candidate \attatomc can be attached to \attatomp at $\node^{*(t)}$
by looking at \attatomc its own (i.e., $\atomEmbc$),
the features of \attatomp (i.e., $\atomEmbp$), 
its context in $\node_c$ (i.e., ${\ntype_c}$) and ``how much'' $\node^{*(t)}$ should be expanded 
(i.e., ${\latent}^+$). 
The candidate with the highest score is selected as the attachment point in $\node_c$. 
\molmod learns  $\Fcapp$ by minimizing the  negative log likelihood
of ground-truth attachment points.

\noindent
\emph{\ul{Valence Checking}}
\label{sec:method:decoder:nfa:app:checking}
%
In \app, \molmod incorporates valence check to only generate and predict legitimate 
candidate attachment points that do not violate valence laws.

\noindent
\emph{\underline{Molecule size constraint}}
\label{sec:method:decoder:nfa:app:size}
%
{Following You {\etal}~{\cite{you2018graph}}, for {\plogp} optimization, we limit the size of optimized molecules 
to at most 38 (38 is the maximum number of atoms in the molecules in the ZINC dataset~{\cite{sterling2015zinc}}).}
{With this molecule size constraint,
{\molmod} can avoid increasing {\plogp} by trivially increasing molecule size, which may have the efforts of improving
{\plogp}~{\cite{wildman1999}}}. 
%

\subsubsection*{Sampling Schemes}
\label{sec:method:decoder:sample}

In the decoding process, for each $\mol_x$, \molmod
samples twenty times from the latent space of \latent and optimize $\mol_x$ accordingly. 
Among all decoded molecules satisfying the similarity constraint with \molx, 
\molmod selects the one of best property as its output.
%

\subsection*{\molmod Pipelines}
\label{sec:method:overall}

%
A pipeline of \molmod models, denoted as \pipeline (Algorithm S4 in Section S14), 
is constructed with a series of identical \molmod models, with the output molecule from one {\molmod} model as the input to the next. 
Given an input molecule $\mol^{(t)}$ to the \mbox{$t$-th} \molmod model  ($\mol^{(0)}$=\mol),
\molmod first optimizes $\mol^{(t)}$ into \mbox{$\mol^{(t+1)}$} as the  output of this model. 
$\mol^{(t+1)}$ is then fed into the \mbox{($t$+$1$)}-th model if it satisfies the similarity constraint  
\mbox{$\text{sim}({\mol^{(t+1)}\!,\mol})\!>\!\delta$} and property constraint 
\mbox{$\plogp(\mol^{(t+1)}\!)\!>\!\plogp(\mol^{(t)}\!)$}. 
Otherwise, $\mol^{(t)}$ is output as the final result and \pipeline stops.
In addition to \pipeline, which outputs one optimized molecule for each input molecule, \pipelineF is developed to output multiple 
optimized molecules for each input molecule. 
Details about \pipelineF are available in Section S2. 

The advantages of this iterative, one-fragment-at-one-time optimization process include that
1) it is easier to control intermediate optimization steps so as to result in optimized molecules of desired similarities and 
properties; 
2) it is easier to optimize multiple fragments in a molecule that are far apart; 
and
3) it follows a rational molecule design process~\cite{hajduk2007decade} and thus could enable more insights and
inform \emph{in vitro} lead optimization.
%

\subsection*{Model Training}
\label{sec:method:training}

%
During model training, we apply teacher forcing to feed the ground truth instead of the prediction results to 
the sequential decoding process.
Following the idea of variational autoencoder, we minimize the following loss 
function to maximize the likelihood $P(\moly|\vect{z},\molx)$. Thus, the optimization problem is formulated as follows,
\begin{equation}
\begin{aligned}
\label{eqn:opt_prob}
\min_{\boldsymbol{\Theta}} -\beta D_{\text{KL}}(q_{\boldsymbol{\phi}}(\latent|\molx,\moly)\|p_{\boldsymbol{\theta}}(\latent))
+E_{\scriptsize{q_{\boldsymbol{\phi}}(\latent|\molx,\moly)}}[\log p_{\boldsymbol{\theta}}(\moly|\latent,\molx)], 
\end{aligned}
\end{equation}
where $\boldsymbol{\Theta}$ is the set of parameters; $q_{\boldsymbol{\phi}}()$ is an estimated posterior 
probability function (\encoder);
$p_{\boldsymbol{\theta}}(\moly|\latent,\molx)$ is the probabilistic decoder representing the likelihood of generating {\moly} 
given the latent embedding \latent and \molx; and the prior $p_{\boldsymbol{\theta}}(\latent)$ follows $\mathcal{N}(\vect{0},\vect{I})$. 
In the above problem, 
\mbox{$D_{\text{KL}}()$} is the KL divergence between 
$q_{\boldsymbol{\phi}}()$ and
$p_{\boldsymbol{\theta}}()$. 
Specifically, the second term represents the prediction or empirical error, 
defined as the sum of all the loss functions in the above six predictions (Eqn~~\ref{eqn:target}-\ref{eqn:attach:child}). 
We use AMSGRAD~\cite{reddi2019convergence} to optimize the learning objective.

\section*{Data Availability}
\label{sec:data_availability}

The data used in this manuscript is made publicly available at Chen \etal~\cite{Ziqi2021} and the link https://github.com/ziqi92/Modof.

\section*{Code Availability}
\label{sec:code_availability}

The code for \molmod, \pipeline and \pipelineF is made publicly available at Chen \etal~\cite{Ziqi2021} and
 the link \\ https://github.com/ziqi92/Modof. 

\section*{Acknowledgements}

This project was made possible, in part, by support from
the National Science Foundation grant numbers (IIS-1855501, X.N.; 
IIS-1827472, X.N.; IIS-2133650, X.N., S.P.; \mbox{OAC-2018627}, S.P.), and the National Library of Medicine (grant
numbers 1R01LM012605-01A1, X.N.; and 1R21LM013678-01, X.N.), an AWS Machine Learning Research Award (X.N.) 
and The Ohio State University President’s Research Excellence program (X.N.).
Any opinions, findings,
and conclusions or recommendations expressed in this material
are those of the authors, and do not necessarily reflect
the views of the funding agencies.
We thank Dr. Xiaoxue Wang and Dr. Xiaolin Cheng for their constructive comments.

\section*{Author Contributions}

X.N. conceived the research; X.N. and S.P. obtained funding for the research, and co-supervised Z.C.; 
Z.C., M.R.M., S.P. and X.N. designed the research; 
Z.C. and X.N. conducted the research, including data curation, formal analysis, methodology design and implementation, result 
analysis and visualization; 
Z.C. drafted the original manuscript; 
M.R.M. provided comments on the original manuscript; 
Z.C., X.N. and S.P. conducted the manuscript editing and revision; 
all authors reviewed the final manuscript. 

\section*{Competing Interests}

M.R.M. was employed by the company NEC Labs America. The remaining authors
declare that the research was conducted in the absence of any commercial or financial relationships that
could be construed as a potential conflict of interest.

\onecolumn

\addtolength{\rightmargin}{.5in}
\addtolength{\leftmargin}{.5in}

\twocolumn
{
	\footnotesize
	
}

\pagebreak

\setcounter{section}{0}
\renewcommand{\thesection}{S\arabic{section}}

\setcounter{table}{0}
\renewcommand{\thetable}{S\arabic{table}}

\setcounter{figure}{0}
\renewcommand{\thefigure}{S\arabic{figure}}

\setcounter{algorithm}{0}
\renewcommand{\thealgorithm}{S\arabic{algorithm}}

\setcounter{equation}{0}
\renewcommand{\theequation}{S\arabic{equation}}
\onecolumn

\begin{center}
\begin{minipage}{0.95\linewidth}
	\centering
	\LARGE 
	A Deep Generative Model for Molecule Optimization via One Fragment Modification (Supplementary Information)
\end{minipage}
\end{center}
\vspace{10pt}

\section{Additional Related Work}
\label{appendix:related_work}

There exists some limited work that follow the fragment-based drug design as \molmod does.
Podda \etal~\cite{podda2020} decomposed molecules into fragments using an algorithm 
that breaks bonds according to chemical reactions.
They then represented each fragment using its SMILES string and 
generated molecules via a VAE-based model, which sequentially decodes the SMILES strings of the fragments
and reassembles the decoded SMILES strings into complete molecules.

Similar to \molmod, a recent work named Teacher and Student polish (\tspolish)~\cite{ji2020graph} also proposed to retain the scaffolds 
and edit the molecules by removing and adding fragments.
%
However, \molmod is fundamentally different from \tspolish.
\tspolish employs a teacher component to identify the logic rules from training molecules that can transform one molecule to another
with better properties.
Thus, the logic rules describe an one-to-one mapping between the two molecules.
\tspolish then learns from these logic rules in the student component, and 
uses the student component to polish or modify new molecules.
The limitation of \tspolish is that it generates only one modified molecule for each input molecule. 
However, there could be multiple ways to optimize one molecule, and 
as suggested in Jin \etal~\cite{jin2019learning}, generative models should be able to generate 
such multiple, diverse optimized molecules for each input molecule.   
In contrast to \tspolish, \molmod samples from a latent `difference' distribution during testing and thus is able to generate multiple 
diverse optimized molecules.

In addition to \tspolish, Lim \etal~\cite{lim2020} also developed a scaffold-based method to generate the molecules 
from scaffolds. 
Their method takes a scaffold 
as input and completes the input scaffold into a molecule through sequentially adding atoms and bonds via a VAE-based 
model.
The limitation of their method is that the retained scaffolds must be cyclic skeletons
extracted from training data by removing side-chain atoms.
Due to this pre-defined scaffold vocabulary, their model is only able to add side-chain atoms to input scaffolds, and their 
generated molecules are limited within the chemical subspace shattered by their scaffold vocabulary. 
In contrast to their method, \molmod is able to learn to determine the scaffolds that need to be
retained from input test molecules, and completes the identified scaffolds with fragments that could be more complicated 
than just side-chain atoms.
Hence, \molmod has the potential to explore more widely in the chemical space for molecule optimization.
We could not compare \molmod with \tspolish as the \tspolish authors haven't published their code.
They also applied a different experimental setting (e.g., sample only once for all the baseline methods and thus
lead to underestimation of the baselines) so that we could not directly use their reported results.
Issues related to their parameter setting are discussed in Supplementary Information Section~\ref{appendix:discussion:para_set}.

In addition to deep generative models, some genetic algorithm-based methods are also developed to find 
molecules with better properties.
Ahn \etal~\cite{ahn2020guiding} developed a genetic expert-guided learning method  (\gegl), in which
they used an expert policy to modify molecules through mutation and crossover, 
and learned a parameterized apprentice policy from good molecules modified by the expert policy for imitation learning.
Nigam \etal~\cite{nigam2020} used a genetic algorithm to modify molecules with random mutations defined 
in Krenn \etal~\cite{krenn2019},
and employed a discriminator to prevent the genetic algorithm from searching the same or similar molecules repeatedly.
%

\section{{{\pipelineF} Protocol}}
\label{appendix:pipelineF}

{We further allow {\pipeline} to output multiple, diverse optimized molecules (the corresponding pipeline 
denoted as {\pipelineF}) following the below protocol: }
%
%
\begin{enumerate}[label=\arabic*), noitemsep,topsep=0pt]
\item {At the $t$-th iteration in {\pipelineF}, {\molmod} optimizes each input molecule into 20 output molecules
via 20 times of sampling and decoding. }
%
\item {Among all the output molecules from iteration $t$ that satisfy the similarity constraint, the top-5 molecules with the best properties 
are fed into the next, ($t$+1)-th iteration.
The remaining output molecules may still have better properties compared to the input molecule to {\pipelineF}, but they will not 
be further optimized by {\pipelineF}. 
Note that the top-5 molecules may not always have improved properties (e.g., when all the output molecules do not have improved properties), 
but they will still be further optimized in the downstream iterations.
} 
%
\item {The above two steps are conducted at each iteration up to five iterations, or until the iteration does not output any molecules 
(e.g., molecules cannot be decoded, similarity constraints are not satisfied), and then {\pipelineF} stops. }
\item {Once {\pipelineF} has stopped, all the unique molecules output at each iteration that are not further optimized
(either not fed into the next iteration, or output at the last iteration)
 are collected, and the top-20 molecules among them with the best properties will be the output, optimized molecules of {\pipelineF}. }
\end{enumerate}
{
Algorithm~{\ref{alg:pipelineF}} presents the {\pipelineF} algorithm. 
}

\section{Data Used in {{\plogp} Optimization }Experiments}
\label{appendix:data}

\begin{table}
	\centering
	\caption{Data Statistics for {\plogp} Optimization}
	\label{tbl:stats}
	
	\begin{threeparttable}
		\scriptsize{
		\begin{tabular}{
				@{\hspace{2pt}}p{0.5\linewidth}@{\hspace{2pt}}
				@{\hspace{2pt}}r@{\hspace{2pt}}          
			}
			\toprule
			description & value \\
			\midrule
			\texttt{\#}training molecules        & 104,708\\
			\texttt{\#}training (\molx, \moly) pairs            & 55,686\\
			\texttt{\#}validation molecules            & 200\\
			\texttt{\#}test molecules            & 800\\
			\midrule
			average similarity of training (\molx, \moly) pairs & {{0.6654}}\\
			average pairwise similarity between training and test molecules & {{0.1070}}\\
			\midrule
			average training molecule size & 25.04\\
			average training $\{\molx\}$ size       & 22.75\\
			average training $\{\moly\}$ size       & 27.07\\
			average test molecule size     & 20.50\\
			\midrule
			average $\{\molx\}$ p$\log$P                     &  -0.7362 \\
			average $\{\moly\}$ p$\log$P                     & 1.1638 \\
			average test molecule p$\log$P             & -2.7468 \\
			average p$\log$P improvement in training \mbox{(\molx, \moly)} pairs     & {1.9000} \\
			\bottomrule
		\end{tabular}
		}
	\end{threeparttable}
	
\end{table}

Table~\ref{tbl:stats} presents the statistics of the data used in our experiments {for {\plogp} optimization}. 


\begin{figure*}
	\centering
	\begin{subfigure}{\linewidth}
		\centering
		\begingroup
		\makeatletter
		\providecommand\color[2][]{%
			\GenericError{(gnuplot) \space\space\space\@spaces}{%
				Package color not loaded in conjunction with
				terminal option `colourtext'%
			}{See the gnuplot documentation for explanation.%
			}{Either use 'blacktext' in gnuplot or load the package
				color.sty in LaTeX.}%
			\renewcommand\color[2][]{}%
		}%
		\providecommand\includegraphics[2][]{%
			\GenericError{(gnuplot) \space\space\space\@spaces}{%
				Package graphicx or graphics not loaded%
			}{See the gnuplot documentation for explanation.%
			}{The gnuplot epslatex terminal needs graphicx.sty or graphics.sty.}%
			\renewcommand\includegraphics[2][]{}%
		}%
		\providecommand\rotatebox[2]{#2}%
		\@ifundefined{ifGPcolor}{%
			\newif\ifGPcolor
			\GPcolortrue
		}{}%
		\@ifundefined{ifGPblacktext}{%
			\newif\ifGPblacktext
			\GPblacktexttrue
		}{}%
		\let\gplgaddtomacro\g@addto@macro
		\gdef\gplbacktext{}%
		\gdef\gplfronttext{}%
		\makeatother
		\ifGPblacktext
		\def\colorrgb#1{}%
		\def\colorgray#1{}%
		\else
		\ifGPcolor
		\def\colorrgb#1{\color[rgb]{#1}}%
		\def\colorgray#1{\color[gray]{#1}}%
		\expandafter\def\csname LTw\endcsname{\color{white}}%
		\expandafter\def\csname LTb\endcsname{\color{black}}%
		\expandafter\def\csname LTa\endcsname{\color{black}}%
		\expandafter\def\csname LT0\endcsname{\color[rgb]{1,0,0}}%
		\expandafter\def\csname LT1\endcsname{\color[rgb]{0,1,0}}%
		\expandafter\def\csname LT2\endcsname{\color[rgb]{0,0,1}}%
		\expandafter\def\csname LT3\endcsname{\color[rgb]{1,0,1}}%
		\expandafter\def\csname LT4\endcsname{\color[rgb]{0,1,1}}%
		\expandafter\def\csname LT5\endcsname{\color[rgb]{1,1,0}}%
		\expandafter\def\csname LT6\endcsname{\color[rgb]{0,0,0}}%
		\expandafter\def\csname LT7\endcsname{\color[rgb]{1,0.3,0}}%
		\expandafter\def\csname LT8\endcsname{\color[rgb]{0.5,0.5,0.5}}%
		\else
		\def\colorrgb#1{\color{black}}%
		\def\colorgray#1{\color[gray]{#1}}%
		\expandafter\def\csname LTw\endcsname{\color{white}}%
		\expandafter\def\csname LTb\endcsname{\color{black}}%
		\expandafter\def\csname LTa\endcsname{\color{black}}%
		\expandafter\def\csname LT0\endcsname{\color{black}}%
		\expandafter\def\csname LT1\endcsname{\color{black}}%
		\expandafter\def\csname LT2\endcsname{\color{black}}%
		\expandafter\def\csname LT3\endcsname{\color{black}}%
		\expandafter\def\csname LT4\endcsname{\color{black}}%
		\expandafter\def\csname LT5\endcsname{\color{black}}%
		\expandafter\def\csname LT6\endcsname{\color{black}}%
		\expandafter\def\csname LT7\endcsname{\color{black}}%
		\expandafter\def\csname LT8\endcsname{\color{black}}%
		\fi
		\fi
		\setlength{\unitlength}{0.0500bp}%
		\ifx\gptboxheight\undefined%
		\newlength{\gptboxheight}%
		\newlength{\gptboxwidth}%
		\newsavebox{\gptboxtext}%
		\fi%
		\setlength{\fboxrule}{0.5pt}%
		\setlength{\fboxsep}{1pt}%
		\begin{picture}(7200.00,720.00)%
		\gplgaddtomacro\gplbacktext{%
		}%
		\gplgaddtomacro\gplfronttext{%
			\csname LTb\endcsname
			\put(2876,547){\makebox(0,0)[r]{\strut{}\scriptsize{training molecules}}}%
			\csname LTb\endcsname
			\put(6239,547){\makebox(0,0)[r]{\strut{}\scriptsize{ZINC molecues}}}%
		}%
		\gplbacktext
		\put(0,0){\includegraphics{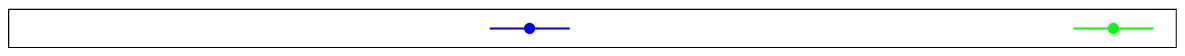}}%
		\gplfronttext
		\end{picture}%
		\endgroup
		\vspace{-10pt}
	\end{subfigure}%
	\\
	\begin{subfigure}{.3\linewidth}
		\centering
		\caption{}
		\vspace{-5pt}
		\begingroup
		\makeatletter
		\providecommand\color[2][]{%
			\GenericError{(gnuplot) \space\space\space\@spaces}{%
				Package color not loaded in conjunction with
				terminal option `colourtext'%
			}{See the gnuplot documentation for explanation.%
			}{Either use 'blacktext' in gnuplot or load the package
				color.sty in LaTeX.}%
			\renewcommand\color[2][]{}%
		}%
		\providecommand\includegraphics[2][]{%
			\GenericError{(gnuplot) \space\space\space\@spaces}{%
				Package graphicx or graphics not loaded%
			}{See the gnuplot documentation for explanation.%
			}{The gnuplot epslatex terminal needs graphicx.sty or graphics.sty.}%
			\renewcommand\includegraphics[2][]{}%
		}%
		\providecommand\rotatebox[2]{#2}%
		\@ifundefined{ifGPcolor}{%
			\newif\ifGPcolor
			\GPcolorfalse
		}{}%
		\@ifundefined{ifGPblacktext}{%
			\newif\ifGPblacktext
			\GPblacktexttrue
		}{}%
		\let\gplgaddtomacro\g@addto@macro
		\gdef\gplbacktext{}%
		\gdef\gplfronttext{}%
		\makeatother
		\ifGPblacktext
		\def\colorrgb#1{}%
		\def\colorgray#1{}%
		\else
		\ifGPcolor
		\def\colorrgb#1{\color[rgb]{#1}}%
		\def\colorgray#1{\color[gray]{#1}}%
		\expandafter\def\csname LTw\endcsname{\color{white}}%
		\expandafter\def\csname LTb\endcsname{\color{black}}%
		\expandafter\def\csname LTa\endcsname{\color{black}}%
		\expandafter\def\csname LT0\endcsname{\color[rgb]{1,0,0}}%
		\expandafter\def\csname LT1\endcsname{\color[rgb]{0,1,0}}%
		\expandafter\def\csname LT2\endcsname{\color[rgb]{0,0,1}}%
		\expandafter\def\csname LT3\endcsname{\color[rgb]{1,0,1}}%
		\expandafter\def\csname LT4\endcsname{\color[rgb]{0,1,1}}%
		\expandafter\def\csname LT5\endcsname{\color[rgb]{1,1,0}}%
		\expandafter\def\csname LT6\endcsname{\color[rgb]{0,0,0}}%
		\expandafter\def\csname LT7\endcsname{\color[rgb]{1,0.3,0}}%
		\expandafter\def\csname LT8\endcsname{\color[rgb]{0.5,0.5,0.5}}%
		\else
		\def\colorrgb#1{\color{black}}%
		\def\colorgray#1{\color[gray]{#1}}%
		\expandafter\def\csname LTw\endcsname{\color{white}}%
		\expandafter\def\csname LTb\endcsname{\color{black}}%
		\expandafter\def\csname LTa\endcsname{\color{black}}%
		\expandafter\def\csname LT0\endcsname{\color{black}}%
		\expandafter\def\csname LT1\endcsname{\color{black}}%
		\expandafter\def\csname LT2\endcsname{\color{black}}%
		\expandafter\def\csname LT3\endcsname{\color{black}}%
		\expandafter\def\csname LT4\endcsname{\color{black}}%
		\expandafter\def\csname LT5\endcsname{\color{black}}%
		\expandafter\def\csname LT6\endcsname{\color{black}}%
		\expandafter\def\csname LT7\endcsname{\color{black}}%
		\expandafter\def\csname LT8\endcsname{\color{black}}%
		\fi
		\fi
		\setlength{\unitlength}{0.0500bp}%
		\ifx\gptboxheight\undefined%
		\newlength{\gptboxheight}%
		\newlength{\gptboxwidth}%
		\newsavebox{\gptboxtext}%
		\fi%
		\setlength{\fboxrule}{0.2pt}%
		\setlength{\fboxsep}{0.5pt}%
		\begin{picture}(2160.00,1260.00)%
		\gplgaddtomacro\gplbacktext{%
			\csname LTb\endcsname
			\put(-66,226){\makebox(0,0)[r]{\strut{}\scriptsize{1.0}}}%
			\put(-66,428){\makebox(0,0)[r]{\strut{}\scriptsize{2.0}}}%
			\put(-66,630){\makebox(0,0)[r]{\strut{}\scriptsize{3.0}}}%
			\put(-66,832){\makebox(0,0)[r]{\strut{}\scriptsize{4.0}}}%
			\put(-66,1034){\makebox(0,0)[r]{\strut{}\scriptsize{5.0}}}%
			\put(-66,1236){\makebox(0,0)[r]{\strut{}\scriptsize{6.0}}}%
			\put(0,-88){\makebox(0,0){\strut{}\scriptsize{0}}}%
			\put(382,-88){\makebox(0,0){\strut{}\scriptsize{10}}}%
			\put(764,-88){\makebox(0,0){\strut{}\scriptsize{20}}}%
			\put(1147,-88){\makebox(0,0){\strut{}\scriptsize{30}}}%
			\put(1529,-88){\makebox(0,0){\strut{}\scriptsize{40}}}%
			\put(1911,-88){\makebox(0,0){\strut{}\scriptsize{50}}}%
		}%
		\gplgaddtomacro\gplfronttext{%
			\csname LTb\endcsname
			\put(-352,630){\rotatebox{-270}{\makebox(0,0){\strut{}\scriptsize{\% in cluster}}}}%
			\put(1079,-264){\makebox(0,0){\strut{}\scriptsize{cluster ID}}}%
		}%
		\gplbacktext
		\put(0,0){\includegraphics{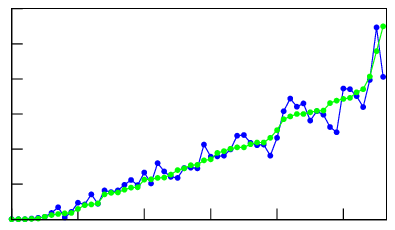}}%
		\gplfronttext
		\end{picture}%
		\endgroup
		\label{fig:coverage:train}
	\end{subfigure}
	\begin{subfigure}{.3\linewidth}
		\centering
		\caption{}
		\vspace{-5pt}
		\begingroup
		\makeatletter
		\providecommand\color[2][]{%
			\GenericError{(gnuplot) \space\space\space\@spaces}{%
				Package color not loaded in conjunction with
				terminal option `colourtext'%
			}{See the gnuplot documentation for explanation.%
			}{Either use 'blacktext' in gnuplot or load the package
				color.sty in LaTeX.}%
			\renewcommand\color[2][]{}%
		}%
		\providecommand\includegraphics[2][]{%
			\GenericError{(gnuplot) \space\space\space\@spaces}{%
				Package graphicx or graphics not loaded%
			}{See the gnuplot documentation for explanation.%
			}{The gnuplot epslatex terminal needs graphicx.sty or graphics.sty.}%
			\renewcommand\includegraphics[2][]{}%
		}%
		\providecommand\rotatebox[2]{#2}%
		\@ifundefined{ifGPcolor}{%
			\newif\ifGPcolor
			\GPcolorfalse
		}{}%
		\@ifundefined{ifGPblacktext}{%
			\newif\ifGPblacktext
			\GPblacktexttrue
		}{}%
		\let\gplgaddtomacro\g@addto@macro
		\gdef\gplbacktext{}%
		\gdef\gplfronttext{}%
		\makeatother
		\ifGPblacktext
		\def\colorrgb#1{}%
		\def\colorgray#1{}%
		\else
		\ifGPcolor
		\def\colorrgb#1{\color[rgb]{#1}}%
		\def\colorgray#1{\color[gray]{#1}}%
		\expandafter\def\csname LTw\endcsname{\color{white}}%
		\expandafter\def\csname LTb\endcsname{\color{black}}%
		\expandafter\def\csname LTa\endcsname{\color{black}}%
		\expandafter\def\csname LT0\endcsname{\color[rgb]{1,0,0}}%
		\expandafter\def\csname LT1\endcsname{\color[rgb]{0,1,0}}%
		\expandafter\def\csname LT2\endcsname{\color[rgb]{0,0,1}}%
		\expandafter\def\csname LT3\endcsname{\color[rgb]{1,0,1}}%
		\expandafter\def\csname LT4\endcsname{\color[rgb]{0,1,1}}%
		\expandafter\def\csname LT5\endcsname{\color[rgb]{1,1,0}}%
		\expandafter\def\csname LT6\endcsname{\color[rgb]{0,0,0}}%
		\expandafter\def\csname LT7\endcsname{\color[rgb]{1,0.3,0}}%
		\expandafter\def\csname LT8\endcsname{\color[rgb]{0.5,0.5,0.5}}%
		\else
		\def\colorrgb#1{\color{black}}%
		\def\colorgray#1{\color[gray]{#1}}%
		\expandafter\def\csname LTw\endcsname{\color{white}}%
		\expandafter\def\csname LTb\endcsname{\color{black}}%
		\expandafter\def\csname LTa\endcsname{\color{black}}%
		\expandafter\def\csname LT0\endcsname{\color{black}}%
		\expandafter\def\csname LT1\endcsname{\color{black}}%
		\expandafter\def\csname LT2\endcsname{\color{black}}%
		\expandafter\def\csname LT3\endcsname{\color{black}}%
		\expandafter\def\csname LT4\endcsname{\color{black}}%
		\expandafter\def\csname LT5\endcsname{\color{black}}%
		\expandafter\def\csname LT6\endcsname{\color{black}}%
		\expandafter\def\csname LT7\endcsname{\color{black}}%
		\expandafter\def\csname LT8\endcsname{\color{black}}%
		\fi
		\fi
		\setlength{\unitlength}{0.0500bp}%
		\ifx\gptboxheight\undefined%
		\newlength{\gptboxheight}%
		\newlength{\gptboxwidth}%
		\newsavebox{\gptboxtext}%
		\fi%
		\setlength{\fboxrule}{0.2pt}%
		\setlength{\fboxsep}{0.5pt}%
		\begin{picture}(2160.00,1260.00)%
		\gplgaddtomacro\gplbacktext{%
			\csname LTb\endcsname
			\put(-66,226){\makebox(0,0)[r]{\strut{}\scriptsize{1.0}}}%
			\put(-66,428){\makebox(0,0)[r]{\strut{}\scriptsize{2.0}}}%
			\put(-66,630){\makebox(0,0)[r]{\strut{}\scriptsize{3.0}}}%
			\put(-66,832){\makebox(0,0)[r]{\strut{}\scriptsize{4.0}}}%
			\put(-66,1034){\makebox(0,0)[r]{\strut{}\scriptsize{5.0}}}%
			\put(-66,1236){\makebox(0,0)[r]{\strut{}\scriptsize{6.0}}}%
			\put(0,-88){\makebox(0,0){\strut{}\scriptsize{0}}}%
			\put(382,-88){\makebox(0,0){\strut{}\scriptsize{10}}}%
			\put(764,-88){\makebox(0,0){\strut{}\scriptsize{20}}}%
			\put(1147,-88){\makebox(0,0){\strut{}\scriptsize{30}}}%
			\put(1529,-88){\makebox(0,0){\strut{}\scriptsize{40}}}%
			\put(1911,-88){\makebox(0,0){\strut{}\scriptsize{50}}}%
		}%
		\gplgaddtomacro\gplfronttext{%
			\csname LTb\endcsname
			\put(-352,630){\rotatebox{-270}{\makebox(0,0){\strut{}\scriptsize{\% in cluster}}}}%
			\put(1079,-264){\makebox(0,0){\strut{}\scriptsize{cluster ID}}}%
		}%
		\gplbacktext
		\put(0,0){\includegraphics{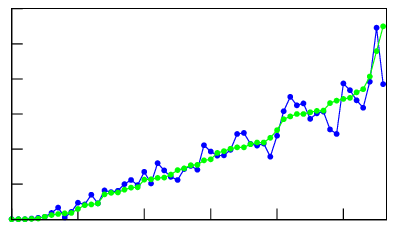}}%
		\gplfronttext
		\end{picture}%
		\endgroup
		\label{fig:coverage:Mx}
	\end{subfigure}%
	\begin{subfigure}{.3\linewidth}
		\centering
		\caption{}
		\vspace{-5pt}
		\begingroup
		\makeatletter
		\providecommand\color[2][]{%
			\GenericError{(gnuplot) \space\space\space\@spaces}{%
				Package color not loaded in conjunction with
				terminal option `colourtext'%
			}{See the gnuplot documentation for explanation.%
			}{Either use 'blacktext' in gnuplot or load the package
				color.sty in LaTeX.}%
			\renewcommand\color[2][]{}%
		}%
		\providecommand\includegraphics[2][]{%
			\GenericError{(gnuplot) \space\space\space\@spaces}{%
				Package graphicx or graphics not loaded%
			}{See the gnuplot documentation for explanation.%
			}{The gnuplot epslatex terminal needs graphicx.sty or graphics.sty.}%
			\renewcommand\includegraphics[2][]{}%
		}%
		\providecommand\rotatebox[2]{#2}%
		\@ifundefined{ifGPcolor}{%
			\newif\ifGPcolor
			\GPcolorfalse
		}{}%
		\@ifundefined{ifGPblacktext}{%
			\newif\ifGPblacktext
			\GPblacktexttrue
		}{}%
		\let\gplgaddtomacro\g@addto@macro
		\gdef\gplbacktext{}%
		\gdef\gplfronttext{}%
		\makeatother
		\ifGPblacktext
		\def\colorrgb#1{}%
		\def\colorgray#1{}%
		\else
		\ifGPcolor
		\def\colorrgb#1{\color[rgb]{#1}}%
		\def\colorgray#1{\color[gray]{#1}}%
		\expandafter\def\csname LTw\endcsname{\color{white}}%
		\expandafter\def\csname LTb\endcsname{\color{black}}%
		\expandafter\def\csname LTa\endcsname{\color{black}}%
		\expandafter\def\csname LT0\endcsname{\color[rgb]{1,0,0}}%
		\expandafter\def\csname LT1\endcsname{\color[rgb]{0,1,0}}%
		\expandafter\def\csname LT2\endcsname{\color[rgb]{0,0,1}}%
		\expandafter\def\csname LT3\endcsname{\color[rgb]{1,0,1}}%
		\expandafter\def\csname LT4\endcsname{\color[rgb]{0,1,1}}%
		\expandafter\def\csname LT5\endcsname{\color[rgb]{1,1,0}}%
		\expandafter\def\csname LT6\endcsname{\color[rgb]{0,0,0}}%
		\expandafter\def\csname LT7\endcsname{\color[rgb]{1,0.3,0}}%
		\expandafter\def\csname LT8\endcsname{\color[rgb]{0.5,0.5,0.5}}%
		\else
		\def\colorrgb#1{\color{black}}%
		\def\colorgray#1{\color[gray]{#1}}%
		\expandafter\def\csname LTw\endcsname{\color{white}}%
		\expandafter\def\csname LTb\endcsname{\color{black}}%
		\expandafter\def\csname LTa\endcsname{\color{black}}%
		\expandafter\def\csname LT0\endcsname{\color{black}}%
		\expandafter\def\csname LT1\endcsname{\color{black}}%
		\expandafter\def\csname LT2\endcsname{\color{black}}%
		\expandafter\def\csname LT3\endcsname{\color{black}}%
		\expandafter\def\csname LT4\endcsname{\color{black}}%
		\expandafter\def\csname LT5\endcsname{\color{black}}%
		\expandafter\def\csname LT6\endcsname{\color{black}}%
		\expandafter\def\csname LT7\endcsname{\color{black}}%
		\expandafter\def\csname LT8\endcsname{\color{black}}%
		\fi
		\fi
		\setlength{\unitlength}{0.0500bp}%
		\ifx\gptboxheight\undefined%
		\newlength{\gptboxheight}%
		\newlength{\gptboxwidth}%
		\newsavebox{\gptboxtext}%
		\fi%
		\setlength{\fboxrule}{0.5pt}%
		\setlength{\fboxsep}{1pt}%
		\begin{picture}(2160.00,1260.00)%
		\gplgaddtomacro\gplbacktext{%
			\csname LTb\endcsname
			\put(-66,226){\makebox(0,0)[r]{\strut{}\scriptsize{1.0}}}%
			\put(-66,428){\makebox(0,0)[r]{\strut{}\scriptsize{2.0}}}%
			\put(-66,630){\makebox(0,0)[r]{\strut{}\scriptsize{3.0}}}%
			\put(-66,832){\makebox(0,0)[r]{\strut{}\scriptsize{4.0}}}%
			\put(-66,1034){\makebox(0,0)[r]{\strut{}\scriptsize{5.0}}}%
			\put(-66,1236){\makebox(0,0)[r]{\strut{}\scriptsize{6.0}}}%
			\put(0,-88){\makebox(0,0){\strut{}\scriptsize{0}}}%
			\put(382,-88){\makebox(0,0){\strut{}\scriptsize{10}}}%
			\put(764,-88){\makebox(0,0){\strut{}\scriptsize{20}}}%
			\put(1147,-88){\makebox(0,0){\strut{}\scriptsize{30}}}%
			\put(1529,-88){\makebox(0,0){\strut{}\scriptsize{40}}}%
			\put(1911,-88){\makebox(0,0){\strut{}\scriptsize{50}}}%
		}%
		\gplgaddtomacro\gplfronttext{%
			\csname LTb\endcsname
			\put(-352,630){\rotatebox{-270}{\makebox(0,0){\strut{}\scriptsize{\% in cluster}}}}%
			\put(1079,-264){\makebox(0,0){\strut{}\scriptsize{cluster ID}}}%
		}%
		\gplbacktext
		\put(0,0){\includegraphics{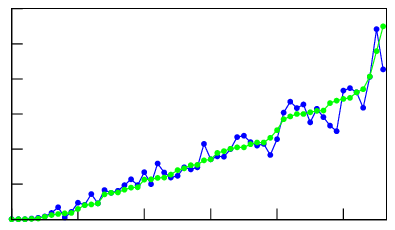}}%
		\gplfronttext
		\end{picture}%
		\endgroup
		\label{fig:coverage:My}
	\end{subfigure}%
	\vspace{10pt}
	\caption{\textbf{Molecule Representativeness for ZINC Chemical Space.} \textbf{a,} Training data representativeness. 
		\textbf{b,} $\{\molx\}_{\text{trn}}$ data representativeness. \textbf{c,} $\{\moly\}_{\text{trn}}$ data representativeness}
	\label{fig:coverage}
\end{figure*}

\subsection{{Training Data Representativeness}}
\label{appendix:data:coverage}
{%
The training data used in our experiments and the baseline methods 
are extracted from the widely used ZINC dataset. 
The ZINC dataset is relatively standard for problems including chemical property prediction~{\cite{ryu2018deeplylm}},
chemical synthesis~{\cite{hung6799}}, optimization~{\cite{jin2019learning, jin2020hierarchical}}, etc. 
Instead of using the entire ZINC or a random subset of ZINC, in {\molmod}, we used pairs of molecules in ZINC that 
satisfy a particular structural constraint: the molecules in a pair are only different in structures at one disconnection site. 
Whether {\molmod}'s training data well represent the ZINC chemical space will affect whether {\molmod} can be well 
generalized in the entire ZINC space. 

To analyze the representativeness of {\molmod}'s training data, we conducted the following analysis: 
We clustered the following three groups of molecules all together: 
(1) the molecules in {\molmod}'s training pairs that have bad properties, denoted as $\{\molx\}_{\text{trn}}$; 
(2) the molecules in {\molmod}'s training pairs that have good properties, denoted as $\{\moly\}_{\text{trn}}$; and
(3) the rest, all the ZINC molecules that are not in $\{\molx\}_{\text{trn}}$ and $\{\moly\}_{\text{trn}}$, denoted as $\{\mol\}_{\text{ZINC}}$. 
In total, we had 324,949 molecules to cluster. 
We first represented each molecule using its canonical SMILE string, and generated a 2,048-dimension binary Morgan fingerprint 
based on the SMILE string. 
We then clustered the 324,949 molecules using the CLUTO~{\cite{karypis2002cluto}}
clustering software. 
CLUTO constructs a graph among the molecules, 
in which each molecule is connected to its nearest neighbors defined by molecule similarities calculated 
via Tanimoto coefficient over molecule fingerprints. 
Please note that this is exactly the same molecule similarity calculation used in {\molmod}. }


{Fig.~{\ref{fig:coverage}} presents the results from 56 clusters (50 clusters and 6 disconnected components 
in the nearest-neighbor graphs). 
In Fig.~{\ref{fig:coverage}}, the clusters are sorted based on their size, and the $y$ axis represents the percentage of molecules 
in $\{M\}_{\text{ZINC}}$, training data (i.e., $\{\molx\}_{\text{trn}} \cup \{\moly\}_{\text{trn}}$), 
$\{\molx\}_{\text{trn}}$ or $\{\moly\}_{\text{trn}}$ that fall within in each cluster. 
Fig.~{\ref{fig:coverage}} shows that {\molmod}'s training data, 
$\{\molx\}_{\text{trn}}$ and $\{\moly\}_{\text{trn}}$  have data distributions similar to ZINC data
in each of the clusters. 
We also conducted six paired $t$-tests over the data distributions among all the group pairs between
$\{M\}_{\text{ZINC}}$, $\{\molx\}_{\text{trn}} \cup \{\moly\}_{\text{trn}}$, 
$\{\molx\}_{\text{trn}}$ and $\{\moly\}_{\text{trn}}$. The $t$-tests show no statistically significant difference in data distributions 
over clusters among the four groups of molecules, with all the $p$-values close to 1.0.
This indicates that {\molmod}'s training data actually well represent the entire ZINC data, and thus {\molmod}  
is generalizable and applicable to ZINC molecules outside {\molmod}'s training data.   
We also tried different numbers of clusters (e.g., 100, 200), and the above conclusions remain the same. 
}

\section{Graph Edit Path Identification for Training Data Generation}
\label{app:graph}

%
Graph edit distance between tree $\treex$ and $\treey$
is defined as the minimum cost to modify $\treex$ into $\treey$ with the following graph edit operations:
\begin{itemize}[noitemsep,topsep=0pt]
	\item Node addition: add a new labeled node into \treex;
	\item Node deletion: delete an existing node from \treex;
	\item Edge addition: add a new edge between a pair of nodes in \treex; and
	\item Edge deletion: delete an existing edge between a pair of nodes in \treex. 
\end{itemize}
Particularly, we did not allow node or edge substitutions as they can be implemented via deletion and addition operations. 
%
%
We identified the optimal graph edit paths using the \mbox{DF-GED} algorithm \cite{abu2015exact} provided 
by a widely-used package \mbox{NetworkX}~\cite{hagberg2008exploring}.
%
%
To identify disconnection sites, 
we denoted the common nodes between $\verticesx$ and $\verticesy$ as matched
nodes \matchSet (i.e., \matchSet = $\verticesx\cap\verticesy$), nodes only in $\verticesx$ as the removal nodes \deleteSet 
(i.e., \deleteSet = $\verticesx\setminus\verticesy$), and the nodes only in $\verticesy$ as the new nodes \addSet
(i.e., \addSet = $\verticesy \setminus \verticesx$), all with respect to \treex. 
Therefore, the disconnection sites  will be the matched nodes in $\treex$ that are also connected with a new node or a 
removal node, that is, $\{\nodeb|(\nodeb\in \matchSet) \land(\neigh(\nodeb)\cap(\deleteSet\cup\addSet)\neq\emptyset)\}$.

\section{Disconnection Site Analysis {in {\plogp} Training Data}}
\label{appendix:disconnection}

\begin{figure}
	\centering
	\begin{subfigure}[b]{.35\linewidth}
		\centering
		\caption{}
		\begingroup
		\makeatletter
		\providecommand\color[2][]{%
			\GenericError{(gnuplot) \space\space\space\@spaces}{%
				Package color not loaded in conjunction with
				terminal option `colourtext'%
			}{See the gnuplot documentation for explanation.%
			}{Either use 'blacktext' in gnuplot or load the package
				color.sty in LaTeX.}%
			\renewcommand\color[2][]{}%
		}%
		\providecommand\includegraphics[2][]{%
			\GenericError{(gnuplot) \space\space\space\@spaces}{%
				Package graphicx or graphics not loaded%
			}{See the gnuplot documentation for explanation.%
			}{The gnuplot epslatex terminal needs graphicx.sty or graphics.sty.}%
			\renewcommand\includegraphics[2][]{}%
		}%
		\providecommand\rotatebox[2]{#2}%
		\@ifundefined{ifGPcolor}{%
			\newif\ifGPcolor
			\GPcolorfalse
		}{}%
		\@ifundefined{ifGPblacktext}{%
			\newif\ifGPblacktext
			\GPblacktexttrue
		}{}%
		\let\gplgaddtomacro\g@addto@macro
		\gdef\gplbacktext{}%
		\gdef\gplfronttext{}%
		\makeatother
		\ifGPblacktext
		\def\colorrgb#1{}%
		\def\colorgray#1{}%
		\else
		\ifGPcolor
		\def\colorrgb#1{\color[rgb]{#1}}%
		\def\colorgray#1{\color[gray]{#1}}%
		\expandafter\def\csname LTw\endcsname{\color{white}}%
		\expandafter\def\csname LTb\endcsname{\color{black}}%
		\expandafter\def\csname LTa\endcsname{\color{black}}%
		\expandafter\def\csname LT0\endcsname{\color[rgb]{1,0,0}}%
		\expandafter\def\csname LT1\endcsname{\color[rgb]{0,1,0}}%
		\expandafter\def\csname LT2\endcsname{\color[rgb]{0,0,1}}%
		\expandafter\def\csname LT3\endcsname{\color[rgb]{1,0,1}}%
		\expandafter\def\csname LT4\endcsname{\color[rgb]{0,1,1}}%
		\expandafter\def\csname LT5\endcsname{\color[rgb]{1,1,0}}%
		\expandafter\def\csname LT6\endcsname{\color[rgb]{0,0,0}}%
		\expandafter\def\csname LT7\endcsname{\color[rgb]{1,0.3,0}}%
		\expandafter\def\csname LT8\endcsname{\color[rgb]{0.5,0.5,0.5}}%
		\else
		\def\colorrgb#1{\color{black}}%
		\def\colorgray#1{\color[gray]{#1}}%
		\expandafter\def\csname LTw\endcsname{\color{white}}%
		\expandafter\def\csname LTb\endcsname{\color{black}}%
		\expandafter\def\csname LTa\endcsname{\color{black}}%
		\expandafter\def\csname LT0\endcsname{\color{black}}%
		\expandafter\def\csname LT1\endcsname{\color{black}}%
		\expandafter\def\csname LT2\endcsname{\color{black}}%
		\expandafter\def\csname LT3\endcsname{\color{black}}%
		\expandafter\def\csname LT4\endcsname{\color{black}}%
		\expandafter\def\csname LT5\endcsname{\color{black}}%
		\expandafter\def\csname LT6\endcsname{\color{black}}%
		\expandafter\def\csname LT7\endcsname{\color{black}}%
		\expandafter\def\csname LT8\endcsname{\color{black}}%
		\fi
		\fi
		\setlength{\unitlength}{0.0500bp}%
		\ifx\gptboxheight\undefined%
		\newlength{\gptboxheight}%
		\newlength{\gptboxwidth}%
		\newsavebox{\gptboxtext}%
		\fi%
		\setlength{\fboxrule}{0.5pt}%
		\setlength{\fboxsep}{1pt}%
		\begin{picture}(1800.00,1260.00)%
		\gplgaddtomacro\gplbacktext{%
			\csname LTb\endcsname
			\put(-27,174){\makebox(0,0)[r]{\strut{}\footnotesize{10}}}%
			\put(-27,478){\makebox(0,0)[r]{\strut{}\footnotesize{30}}}%
			\put(-27,782){\makebox(0,0)[r]{\strut{}\footnotesize{50}}}%
			\put(-27,1086){\makebox(0,0)[r]{\strut{}\footnotesize{70}}}%
			\put(197,-88){\makebox(0,0){\strut{}\footnotesize{1}}}%
			\put(444,-88){\makebox(0,0){\strut{}\footnotesize{2}}}%
			\put(690,-88){\makebox(0,0){\strut{}\footnotesize{3}}}%
			\put(936,-88){\makebox(0,0){\strut{}\footnotesize{4}}}%
			\put(1183,-88){\makebox(0,0){\strut{}\footnotesize{5}}}%
			\put(1429,-88){\makebox(0,0){\strut{}\footnotesize{6}}}%
			\put(1676,-88){\makebox(0,0){\strut{}\footnotesize{7}}}%
		}%
		\gplgaddtomacro\gplfronttext{%
			\csname LTb\endcsname
			\put(-352,630){\rotatebox{-270}{\makebox(0,0){\strut{}\footnotesize{\% molecule pairs}}}}%
			\put(899,-308){\makebox(0,0){\strut{}\footnotesize{\# sites of disconnection}}}%
		}%
		\gplbacktext
		\put(0,0){\includegraphics{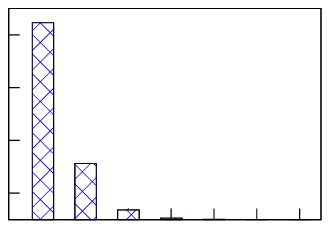}}%
		\gplfronttext
		\end{picture}%
		\endgroup	
		\vspace{20pt}
		\label{fig:bp}
	\end{subfigure}
	\begin{subfigure}[b]{.6\linewidth}
		\centering
		\caption{}
		\begingroup
		\makeatletter
		\providecommand\color[2][]{%
			\GenericError{(gnuplot) \space\space\space\@spaces}{%
				Package color not loaded in conjunction with
				terminal option `colourtext'%
			}{See the gnuplot documentation for explanation.%
			}{Either use 'blacktext' in gnuplot or load the package
				color.sty in LaTeX.}%
			\renewcommand\color[2][]{}%
		}%
		\providecommand\includegraphics[2][]{%
			\GenericError{(gnuplot) \space\space\space\@spaces}{%
				Package graphicx or graphics not loaded%
			}{See the gnuplot documentation for explanation.%
			}{The gnuplot epslatex terminal needs graphicx.sty or graphics.sty.}%
			\renewcommand\includegraphics[2][]{}%
		}%
		\providecommand\rotatebox[2]{#2}%
		\@ifundefined{ifGPcolor}{%
			\newif\ifGPcolor
			\GPcolorfalse
		}{}%
		\@ifundefined{ifGPblacktext}{%
			\newif\ifGPblacktext
			\GPblacktexttrue
		}{}%
		\let\gplgaddtomacro\g@addto@macro
		\gdef\gplbacktext{}%
		\gdef\gplfronttext{}%
		\makeatother
		\ifGPblacktext
		\def\colorrgb#1{}%
		\def\colorgray#1{}%
		\else
		\ifGPcolor
		\def\colorrgb#1{\color[rgb]{#1}}%
		\def\colorgray#1{\color[gray]{#1}}%
		\expandafter\def\csname LTw\endcsname{\color{white}}%
		\expandafter\def\csname LTb\endcsname{\color{black}}%
		\expandafter\def\csname LTa\endcsname{\color{black}}%
		\expandafter\def\csname LT0\endcsname{\color[rgb]{1,0,0}}%
		\expandafter\def\csname LT1\endcsname{\color[rgb]{0,1,0}}%
		\expandafter\def\csname LT2\endcsname{\color[rgb]{0,0,1}}%
		\expandafter\def\csname LT3\endcsname{\color[rgb]{1,0,1}}%
		\expandafter\def\csname LT4\endcsname{\color[rgb]{0,1,1}}%
		\expandafter\def\csname LT5\endcsname{\color[rgb]{1,1,0}}%
		\expandafter\def\csname LT6\endcsname{\color[rgb]{0,0,0}}%
		\expandafter\def\csname LT7\endcsname{\color[rgb]{1,0.3,0}}%
		\expandafter\def\csname LT8\endcsname{\color[rgb]{0.5,0.5,0.5}}%
		\else
		\def\colorrgb#1{\color{black}}%
		\def\colorgray#1{\color[gray]{#1}}%
		\expandafter\def\csname LTw\endcsname{\color{white}}%
		\expandafter\def\csname LTb\endcsname{\color{black}}%
		\expandafter\def\csname LTa\endcsname{\color{black}}%
		\expandafter\def\csname LT0\endcsname{\color{black}}%
		\expandafter\def\csname LT1\endcsname{\color{black}}%
		\expandafter\def\csname LT2\endcsname{\color{black}}%
		\expandafter\def\csname LT3\endcsname{\color{black}}%
		\expandafter\def\csname LT4\endcsname{\color{black}}%
		\expandafter\def\csname LT5\endcsname{\color{black}}%
		\expandafter\def\csname LT6\endcsname{\color{black}}%
		\expandafter\def\csname LT7\endcsname{\color{black}}%
		\expandafter\def\csname LT8\endcsname{\color{black}}%
		\fi
		\fi
		\setlength{\unitlength}{0.0500bp}%
		\ifx\gptboxheight\undefined%
		\newlength{\gptboxheight}%
		\newlength{\gptboxwidth}%
		\newsavebox{\gptboxtext}%
		\fi%
		\setlength{\fboxrule}{0.5pt}%
		\setlength{\fboxsep}{1pt}%
		\begin{picture}(3600.00,1260.00)%
		\gplgaddtomacro\gplbacktext{%
			\csname LTb\endcsname
			\put(-27,174){\makebox(0,0)[r]{\strut{}\footnotesize{5}}}%
			\put(-27,478){\makebox(0,0)[r]{\strut{}\footnotesize{15}}}%
			\put(-27,782){\makebox(0,0)[r]{\strut{}\footnotesize{25}}}%
			\put(-27,1086){\makebox(0,0)[r]{\strut{}\footnotesize{35}}}%
			\put(216,-88){\makebox(0,0){\strut{}\footnotesize{0}}}%
			\put(487,-88){\makebox(0,0){\strut{}\footnotesize{1}}}%
			\put(758,-88){\makebox(0,0){\strut{}\footnotesize{2}}}%
			\put(1028,-88){\makebox(0,0){\strut{}\footnotesize{3}}}%
			\put(1299,-88){\makebox(0,0){\strut{}\footnotesize{4}}}%
			\put(1569,-88){\makebox(0,0){\strut{}\footnotesize{5}}}%
			\put(1840,-88){\makebox(0,0){\strut{}\footnotesize{6}}}%
			\put(2111,-88){\makebox(0,0){\strut{}\footnotesize{7}}}%
			\put(2381,-88){\makebox(0,0){\strut{}\footnotesize{8}}}%
			\put(2652,-88){\makebox(0,0){\strut{}\footnotesize{9}}}%
			\put(2922,-88){\makebox(0,0){\strut{}\footnotesize{10}}}%
			\put(3193,-88){\makebox(0,0){\strut{}\footnotesize{11}}}%
			\put(3464,-88){\makebox(0,0){\strut{}\footnotesize{12}}}%
		}%
		\gplgaddtomacro\gplfronttext{%
			\csname LTb\endcsname
			\put(-352,630){\rotatebox{-270}{\makebox(0,0){\strut{}\footnotesize{\% molecule pairs}}}}%
			\put(1799,-308){\makebox(0,0){\strut{}\footnotesize{\# nodes}}}%
			\csname LTb\endcsname
			\put(2279,1065){\makebox(0,0)[l]{\strut{}\footnotesize{removal nodes}}}%
			\csname LTb\endcsname
			\put(2279,845){\makebox(0,0)[l]{\strut{}\footnotesize{attaching nodes}}}%
		}%
		\gplbacktext
		\put(0,0){\includegraphics{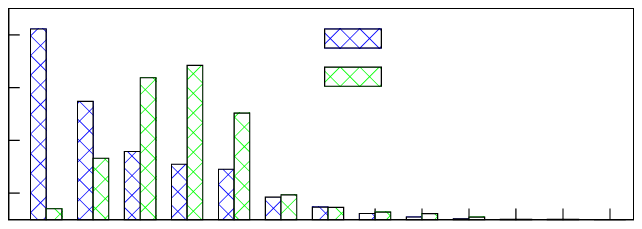}}%
		\gplfronttext
		\end{picture}%
		\endgroup
		\vspace{20pt}
		\label{fig:nodes}
	\end{subfigure}
	\caption{\textbf{Distributions of Modification Operations.} \textbf{a,} Distribution of disconnection site counts. 
	\textbf{b,} Distribution of removal and attaching node counts.}
	\label{fig:dist}
\end{figure}

The key idea of \molmod is to modify one fragment at a time under the similarity and property constraints. 
%
This is based on the assumption that for a pair of molecules with a high similarity (e.g., \mbox{$\text{sim}(\molx,\moly)>0.6$}), 
they are very likely to have only one (small) fragment different, and thus one site of disconnection. 
Fig.~\ref{fig:bp} presents the number of disconnection sites among the 75K pairs of molecules in the 
benchmark dataset provided by Jin \etal~\cite{jin2020hierarchical}, in which each pair has similarity above 0.6.  
%
The distribution in Fig.~\ref{fig:bp} shows that when the molecule similarity is high (e.g., 0.6 in the benchmark dataset), 
most of the molecule pairs (i.e., 74.4\%) 
have only one disconnection site, some of the pairs have two (i.e., 21.1\%), and only 
a few have three. 
This indicates that one-fragment modification at a time is a rational idea and directly applicable to the majority of the optimization 
cases. Even though there could be more disconnection sites, \pipeline and {{\pipelineF}} allow multiple-fragment optimization via 
multiple one-fragment optimizations. 

%
Fig.~\ref{fig:nodes} presents the number of nodes (as in junction tree representations) 
that need to be removed from the disconnection sites 
at \molx (\treex), and the number of nodes that need to be attached at the sites afterwards, 
in Jin's benchmark dataset~\cite{jin2020hierarchical}.  
The figures show that on average, more nodes will be attached to the disconnection sites compared 
to the removal nodes. This indicates that the optimized molecules with better $\plogp$ will become larger, as we have 
observed in our and other's methods. 
\section{Discussion on Parameter Settings {in {\plogp} Optimization}}
\label{appendix:discussion:para_set}

%

Among all the methods that involve random sampling, \jtnn, \hiergtog and our method \molmod sample 20 times 
for each test molecule and thus produce 20 optimized candidates, and identify the best one among these 20 candidates.  
However, \graphaf optimized each test molecule for 200 times and reported the best among those 200 candidates. Thus, 
it is unclear if the overall performance of \graphaf is largely due to the many times of optimization (and thus a larger pool of 
optimized candidates) or the model that does learn how to optimize. 
%
\tspolish \cite{ji2020graph} forced all their compared baseline models including \jtnn and \gcpn to sample only once for each test molecule, 
because \tspolish can only modify one input molecule into one output. 
%
This might not be appropriate either since it artificially underestimated the performance of the baseline models. Instead, it would be fair to 
compare the output from \tspolish with the best output from the baseline methods. 
%


\section{Molecule Similarity Calculation}
\label{appendix:discussion:sim}
%
%

\begin{figure}
	\centering
	\begin{small}
		\fbox{\begin{minipage}{0.7\linewidth}
				\centering
				\scriptsize
				$\text{sim}_{\text{b}}(\molx, \moly)$=0.644,   ~~~~~~~~~~~~~~~~
				$\text{sim}_{\text{c}}(\molx, \moly)$=0.393
				
		\end{minipage}}
	\end{small}
	\\
	%
	\begin{subfigure}{.3\linewidth}
		\centering
		\caption{}
		\includegraphics[width=.8\linewidth]{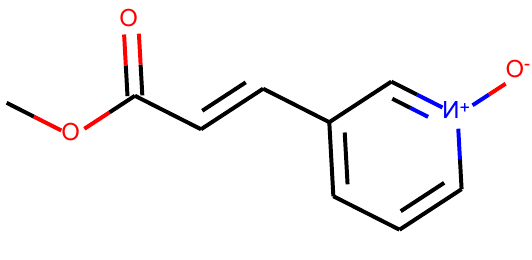}
		\label{fig:original}
	\end{subfigure}
    ~~~~~~~~
	\begin{subfigure}{.4\linewidth}
		\centering
		\vspace{10pt}
		\caption{}
		\vspace{-5pt}
		\includegraphics[width=.8\linewidth]{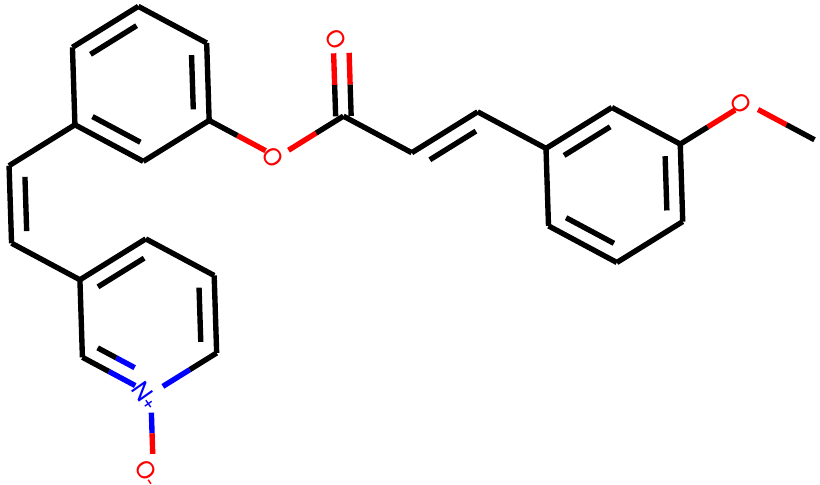}
		\label{fig:step1}
	\end{subfigure}%
	\caption{\textbf{Compound Similarity Comparison.} \textbf{a,} $M_x$. \textbf{b,} $M_y$.}
	\label{fig:sim}
\end{figure}

All baselines except \jtvae in Table~\ref{tbl:overall} in the main manuscript
use binary Morgan fingerprints with radius of 2 and 2,048 bits to represent 
the presence or absence of particular, pre-defined substructures in molecules.
Using such binary fingerprints in molecule similarity calculation may overestimate molecule similarities.  
An example is presented in Fig.~\ref{fig:sim}, where the Tanimoto similarity
from binary Morgan fingerprints ($\text{sim}_b$) of the two molecules is 0.644, 
but they look sufficiently different, with the similarity from Morgan fingerprints 
of substructure counts ($\text{sim}_c$) only 0.393. 
%
According to our experimental results (e.g., Fig.~\ref{fig:example_logp}c and Fig.~\ref{fig:example_logp}d in the main manuscript)
and fragment analysis later in Section~\ref{appendix:opt}, 
we observed that aromatic rings could contribute to large \plogp values and thus would be attached to the optimized molecules.
Using binary Morgan fingerprints in calculating molecule similarities in this case would easily lead to the solution that many 
aromatic rings will be attached for molecule optimization, while still satisfying the similarity constraint due to the similarity overestimation, 
but result in very large molecules that are less drug-like~\cite{Ghose1999}. 
%
%
{Therefore, to prevent the model from generating such molecules, 
we could either consider Morgan fingerprints of substructure counts, or limit the size of optimized molecules.
In {\molmod}, we used the same binary Morgan fingerprints as in the baselines 
and added an additional constraint to limit the size of optimized molecules to be at most 38
(38 is the maximum number of atoms in the molecules in the ZINC dataset). }
%
%
%

\begin{table*}
	\vspace{-25pt}    
	\begin{center}
		\caption{{{\pipeline} Performance on Optimizing {\plogp}}}
		\label{tbl:pipeline2}
		
		\begin{small}
			\begin{threeparttable}
				\scriptsize{
					\begin{tabular}{
							@{\hspace{0pt}}l@{\hspace{1pt}}
							@{\hspace{1pt}}r@{\hspace{1pt}}
							@{\hspace{1pt}}r@{\hspace{2pt}}
							@{\hspace{2pt}}r@{\hspace{2pt}}
							@{\hspace{2pt}}r@{\hspace{2pt}}
							@{\hspace{2pt}}r@{\hspace{3pt}}
							@{\hspace{3pt}}c@{\hspace{2pt}}
							@{\hspace{2pt}}c@{\hspace{2pt}}
							@{\hspace{2pt}}c@{\hspace{3pt}}
							@{\hspace{4pt}}c@{\hspace{2pt}}
							@{\hspace{2pt}}c@{\hspace{4pt}}
							%
							%
							%
							@{\hspace{2pt}}c@{\hspace{1pt}}
							@{\hspace{2pt}}c@{\hspace{1pt}}
							@{\hspace{2pt}}c@{\hspace{2pt}}
							@{\hspace{1pt}}c@{\hspace{1pt}}
							@{\hspace{2pt}}c@{\hspace{1pt}}
							@{\hspace{2pt}}c@{\hspace{1pt}}
							@{\hspace{1pt}}c@{\hspace{2pt}}
							@{\hspace{2pt}}c@{\hspace{0pt}}
						}
						\toprule
						\multirow{2}{*}{$\delta$} & \multirow{2}{*}{$t$} & \multirow{2}{*}{\#in\%}  &  \multirow{2}{*}{\#p\%} & \multirow{2}{*}{\#n\%} & \multirow{2}{*}{\#z\%} 
						& \multicolumn{3}{c}{property improvement} & \multicolumn{2}{c}{molecule sim}   
						&
						\multicolumn{3}{c}{avgsim w. Trn} & &\multicolumn{3}{c}{\mbox{top$\text{-}10$} sim w. Trn}  &sim$_t$  \\	
						\cmidrule(lr){7-9} \cmidrule(lr){10-11}	 \cmidrule(lr){12-14} \cmidrule(lr){16-18}	
						& & & & & & p$_t$$\pm$std       & n$_t$$\pm$std       & p$\pm$std   & sim$_t$$\pm$std          & sim$\pm$std  
						& all  & Trn$_x$ &  Trn$_y$ &  & all & Trn$_x$  &Trn$_y$ &  \mbox{$\!\moly\!$} \\	
						\midrule
						\multirow{5}{*}{0.0}
						& 1  & 100.00 & 99.62 & 0.00 & 0.38 & 5.12$\pm$1.73 & 0.00$\pm$0.00 & 5.10$\pm$1.76 & 0.43$\pm$0.13 & 0.44$\pm$0.13 & 0.119 & 0.112 & 0.125 & & 0.369 & 0.333 & 0.355 & 0.146\\ 
						& 2  & 99.62 & 84.88 & 0.62 & 14.12 & 1.94$\pm$1.26 & -1.55$\pm$1.12 & 6.74$\pm$2.14 & 0.63$\pm$0.19 & 0.26$\pm$0.15 & 0.116 & 0.107 & 0.125 & & 0.365 & 0.319 & 0.358 & 0.173\\ 
						& 3  & 84.88 & 55.00 & 1.75 & 28.12 & 1.07$\pm$0.98 & -0.52$\pm$0.59 & 7.33$\pm$2.23 & 0.76$\pm$0.21 & 0.22$\pm$0.15 & 0.114 & 0.103 & 0.124 & & 0.365 & 0.312 & 0.360 & 0.194\\ 
						& 4  & 55.75 & 32.12 & 0.38 & 23.25 & 0.62$\pm$0.76 & -0.60$\pm$0.78 & 7.53$\pm$2.28 & 0.84$\pm$0.18 & 0.22$\pm$0.15 & 0.111 & 0.099 & 0.122 & & 0.361 & 0.306 & 0.358 & 0.213\\ 
						& 5  & 32.50 & 16.50 & 0.38 & 15.62 & 0.49$\pm$0.61 & -1.68$\pm$1.98 & 7.61$\pm$2.30 & 0.86$\pm$0.17 & 0.21$\pm$0.15 & 0.107 & 0.095 & 0.119 & & 0.356 & 0.299 & 0.353 & 0.225\\ 
						\midrule
						\multirow{5}{*}{0.2}
						& 1  & 100.00 & 99.62 & 0.00 & 0.38 & 4.92$\pm$1.58 & 0.00$\pm$0.00 & 4.91$\pm$1.60 & 0.46$\pm$0.13 & 0.46$\pm$0.12 & 0.117 & 0.111 & 0.123 & & 0.364 & 0.334 & 0.346 & 0.138\\ 
						& 2  & 99.62 & 68.75 & 2.38 & 28.50 & 1.54$\pm$0.97 & -0.78$\pm$0.84 & 5.97$\pm$1.73 & 0.71$\pm$0.19 & 0.36$\pm$0.11 & 0.117 & 0.109 & 0.125 & & 0.351 & 0.317 & 0.340 & 0.154 \\ 
						& 3  & 70.12 & 28.62 & 3.25 & 38.25 & 0.75$\pm$0.71 & -0.67$\pm$0.87 & 6.18$\pm$1.76 & 0.86$\pm$0.15 & 0.34$\pm$0.11 & 0.117 & 0.108 & 0.125 & & 0.348 & 0.312 & 0.340 & 0.166\\
						& 4  & 29.88 & 8.38 & 1.25 & 20.25 & 0.53$\pm$0.48 & -0.28$\pm$0.28 & 6.22$\pm$1.77 & 0.92$\pm$0.13 & 0.34$\pm$0.11 & 0.115 & 0.106 & 0.124 & & 0.345 & 0.307 & 0.338 & 0.175\\
						& 5  & 9.25 & 1.75 & 0.25 & 7.25 & 0.32$\pm$0.22 & -0.17$\pm$0.14 & 6.23$\pm$1.77 & 0.96$\pm$0.09 & 0.34$\pm$0.11 & 0.113 & 0.103 & 0.122 & & 0.344 & 0.308 & 0.338 & 0.177\\
						
						\midrule
						\multirow{5}{*}{0.4}
						&1	 & 100.00 & 99.12 & 0.00 & 0.88 & 4.50$\pm$1.32 & 0.00$\pm$0.00 & 4.47$\pm$1.38 & 0.53$\pm$0.11 & 0.53$\pm$0.11  
						& 0.114 & 0.110 & 0.118 & & 0.366 & 0.338 & 0.344 & 0.131 \\
						&2  & 99.12 & 34.25 & 3.00 & 61.88 & 1.36$\pm$0.95 & -0.65$\pm$0.56 & 4.93$\pm$1.49 & 0.75$\pm$0.22 & 0.49$\pm$0.09 
						& 0.114 & 0.110 & 0.119 & & 0.360 & 0.331 & 0.341 & 0.134\\
						&3  & 37.62 & 8.62 & 0.75 & 28.25 & 0.65$\pm$0.64 & -0.77$\pm$0.67 & 4.99$\pm$1.52 & 0.91$\pm$0.14 & 0.48$\pm$0.09 
						& 0.115 & 0.110 & 0.120 & & 0.360 & 0.329 & 0.343 & 0.146\\
						&4  & 9.25 & 1.88 & 0.12 & 7.25 & 0.57$\pm$0.54 & -0.29$\pm$0.00 & 5.00$\pm$1.53 & 0.93$\pm$0.12 & 0.48$\pm$0.09 
						& 0.114 & 0.109 & 0.119 & & 0.358 & 0.328 & 0.341 & 0.149\\
						&5 & 2.25 & 0.38 & 0.00 & 1.88 & 0.37$\pm$0.06 & 0.00$\pm$0.00 & 5.00$\pm$1.53 & 0.97$\pm$0.06 & 0.48$\pm$0.09  
						& 0.112 & 0.107 & 0.117 & & 0.355 & 0.328 & 0.335 & 0.142\\
						\midrule
						\multirow{5}{*}{0.6}
						& 1 & 100.00 & 84.25 & 0.12 & 15.62 & 3.01$\pm$1.16 & -0.31$\pm$0.00 & 2.54$\pm$1.53 & 0.60$\pm$0.23 & 0.66$\pm$0.05 & 0.109 & 0.107 & 0.111 & & 0.383 & 0.360 & 0.351 & 0.117 \\
						& 2 & 84.25 & 14.62 & 1.88 & 67.75 & 1.16$\pm$0.92 & -0.69$\pm$0.76 & 2.71$\pm$1.68 & 0.83$\pm$0.17 & 0.66$\pm$0.05 & 0.113 & 0.111 & 0.116 & & 0.382 & 0.357 & 0.355 & 0.128 \\
						& 3 & 16.00 & 1.00 & 0.25 & 14.75 & 0.84$\pm$0.67 & -0.03$\pm$0.00 & 2.72$\pm$1.68 & 0.94$\pm$0.11 & 0.65$\pm$0.05 & 0.113 & 0.109 & 0.117 & & 0.369 & 0.342 & 0.346 & 0.136 \\
						& 4 & 1.25 & 0.12 & 0.00 & 1.12 & 0.38$\pm$0.00 & 0.00$\pm$0.00 & 2.72$\pm$1.68 & 0.92$\pm$0.14 & 0.65$\pm$0.05 & 0.117 & 0.112 & 0.121 & & 0.351 & 0.322 & 0.334 & 0.147 \\
						& 5 & 0.12 & 0.00 & 0.00 & 0.12 & 0.00$\pm$0.00 & 0.00$\pm$0.00 & 2.72$\pm$1.68 & 0.00$\pm$0.00 & 0.65$\pm$0.05 & 0.111 & 0.107 & 0.115 & & 0.271 & 0.255 & 0.262 & 0.000 \\
						\bottomrule
					\end{tabular}%
					%
					%
					%
					%
					%
					%
					\begin{tablenotes}[normal,flushleft]
						\begin{scriptsize}
							
							\item 
							Columns represent: ``$t$": the iteration; 
							`\#in\%": the number of input molecules  in each iteration in percentage over 
							all the testing molecules; 
							``\#p"/``\#n"/``\#z\%": the percentage of molecules optimized 
							with better/worse/same properties; 
							``p$_t$"/``n$_t$": property improvement/decline in the $t$-th iteration;
							``p": the overall property improvement up to the $t$-th iteration; 
							``sim$_t$"/``sim": the similarities between the molecules before and after optimization in/up to the $t$-th iteration; 
							%
							%
							``avgsim w. Trn"/``top-10 sim w. Trn": the average similarities with all/top-10 most similar training molecules; 
							``all"/``Trn$_x$"/`Trn$_y$": the comparison molecules identified from 
							all/poor-property/good-property training molecules; 
							\mbox{``sim$_t$ \moly\!\!\!"}: the average pairwise similarities among optimized molecules. 
							\par
						\end{scriptsize}
					\end{tablenotes}
				}
			\end{threeparttable}
		\end{small}
		
	\end{center}
\end{table*}

\section{Comparison Fairness among Existing Methods}
\label{appendix:results:overall:fair}
%

\subsection{Discussion on \plogp Test Sets}

%
Note that in Table~\ref{tbl:overall} in the main manuscript, the results of \graphaf and \moflow are lower than those reported in the respective paper.
This is because in their papers, they used a different test set rather than the benchmark test set, and   
their test molecules were much easier to optimize, which would lead to an unfair comparison with other baseline methods.
In our experiments, we tested \graphaf and \moflow on the benchmark test set and reported the results in Table~\ref{tbl:overall} in the main 
manuscript.
%

	%
	The benchmark test set consists of 800 molecules that have the lowest \plogp values in ZINC test set 
	(ZINC test set was split by G{\'{o}}mez-Bombarelli \etal~\cite{gomez2018automatic}). 
	The \plogp values of these molecules are in range [-11.02, -0.56], 
	with an average value -2.75$\pm$1.52. 
	The test set used in \graphaf and \moflow consists of 800 molecules that have the lowest \plogp values from the 
	\emph{entire} ZINC dataset, not only from the ZINC \emph{test} set.
	The \plogp values of these molecules are in range [-62.52, 2.42],  
	with an average score -12.00$\pm$5.89.  
	That is, the \graphaf and \moflow's test molecules have much worse properties compared to those in the benchmark test set, 
	and they are much easier to optimize with larger property improvement. 
	Due to the different test sets, the results reported in \graphaf and \moflow are not comparable to those reported in other 
	baseline methods.

	%
	For \graphaf, we fine-tuned their pre-trained model with a set of 10K molecules that do not overlap with the benchmark test set. 
	%
	We tested \moflow using the trained model provided by its authors (note that {\moflow} training learns molecule latent 
	representations, not how to optimize molecules; 
	molecule optimization is conducted during testing via gradient-ascent search in the latent representation space). 
	The results show that unfortunately, 
	\graphaf and \moflow do not outperform \gcpn, \jtnn, \hiergtog and our method \molmod on the benchmark 
	test data. 
%


\subsection{Discussion on Reinforcement Learning Settings}

\gcpn, \moldqn and \graphaf are reinforcement learning-based methods. 
	They all used the 800 test molecules  (benchmark test set for \gcpn and \moldqn, 
	and a different non-benchmark test set for \graphaf) to either directly train a model 
	or fine-tune a pre-trained model to optimize the test molecules.
	Therefore, these models are specific to their test set, and would suffer from overfitting issues and 
	not generalize to new molecules. They may also have the issue of not really learning but essentially memorizing 
	optimization actions/paths. 
	Previous studies have analyzed this so-called environment overfitting problem~\cite{whiteson2011,zhang2018}
	in reinforcement learning.
	They concluded that using a test set with overlapping samples in the training set (i.e., non-isolated test set)
	can lead to artificially high performance by simply memorizing the sequences of actions, 
	and suggested that reinforcement learning should use an isolated test set that is completely disjoint with training set.
	The issues with non-isolated test sets remain true for \gcpn, \moldqn and \graphaf.

	In our experiments on \graphaf, instead of using the test set to fine-tune the pre-trained \graphaf model,
	we sampled 10K molecules from ZINC dataset that 
	do not overlap with the benchmark test molecules and also have the property range similar with the test molecules (i.e., [-11.0, -0.5]).
	We then fine-tuned the pre-trained \graphaf model provided by the authors over these 10K molecules 
	and used the fine-tuned model to optimize each test molecule. 
	The results in Table~\ref{tbl:overall} in the main manuscript 
	demonstrate that unfortunately, \graphaf model does not outperform other baseline methods
	over the isolated benchmark test molecules.

\section{Additional Experimental Results on \plogp Optimization}
\label{appendix:results:pipleline}

\subsection{Overall Pipeline Performance}
\label{appendix:results:pipleline:overall}

Table~\ref{tbl:pipeline2} presents the \pipeline performance in each of its iterations
under \mbox{$\delta$=$0.0,0.2,0.4$} and $0.6$, 
where the output of \pipeline at $t$-th iteration for input molecule $\mol_x$ is denoted as $\mol_x^{(t)}$.
%
Without similarity constraints (i.e., $\delta$=0.0),
Table~\ref{tbl:pipeline2} shows that 84.88\% of the molecules can go through three iterations in \pipeline, and 
fewer molecules go through further iterations. 
The property improvement (i.e., ``p$_t\pm$std") gets slower in later iterations as it becomes more difficult to optimize a good molecule.
This is also indicated by the increasing molecule similarities in later iterations (i.e., large sim$_t\pm$std values over iterations).
Still, after each iteration, the overall property improvement out of the pipeline 
(i.e., ``p$\pm$std") increases (e.g., 7.61$\pm$2.30 
after iteration 5), while the overall 
molecule similarity (i.e., ``sim$\pm$std") decreases over iterations.

With similarity constraint \mbox{$\delta$=0.2},  
Table~\ref{tbl:pipeline2} shows trends very similar as to those with \mbox{$\delta$=0.0}.
In addition, at \mbox{$\delta$=0.2}, most molecules can be optimized within first three iterations and do not
go through further iterations (e.g., at \mbox{$t$=4}, \mbox{``\#in\%"=29.88\%} and \mbox{``\#p\%"=8.38\%}).
This could be due to that the decoded molecules in later iterations do not satisfy the similarity constraint.
Accordingly, the output optimized molecules may not necessarily be of the best properties, and  
there are a few output molecules with even declined properties (``\#n\%"). 
With even higher similarity threshold \mbox{$\delta$=0.4 or 0.6}, Table~\ref{tbl:pipeline2} shows that even fewer molecules (\#in\%)
can be further optimized and property improvement from each iteration is also smaller. 
%

\subsection{Molecule Similarity Comparison}
\label{appendix:results:pipeline:sim}
%
Table~\ref{tbl:pipeline2} also presents the similarities between the optimized molecules and training molecules 
(i.e., ``avgsim w. Trn" and \mbox{``top-10 sim w. Trn"}), and the average pairwise similarities among optimized molecules 
(i.e., ``sim$_t$ $\mol_y$"). 
Table~\ref{tbl:pipeline2} shows that as \pipeline optimizes a molecule more,
the average pairwise similarities between the optimized and all the training molecules (``avgsim w. Trn") almost remain 
the same with small values.
This could be due to the effect that there are many training modules (104,708) and the 
all-pairwise similarities are smoothed out.
However, the average similarities between the optimized molecules and their top-10 most 
similar training molecules (\mbox{``top-10 sim w. Trn"}) decrease. 
This indicates that \molmod generates new molecules that are 
in general different from training molecules. 
In addition, the optimized molecules out of each iteration have low similarities around 0.14-0.18 
(\mbox{``sim$_t$ $\mol_y$"}), indicating their diversity. The optimized molecules become slightly more similar to one another. 
This could indicate that the optimized molecules also share certain similarities due to their good \plogp
properties (e.g., aromatic proportion~\cite{Ritchie2009}). 
%

\subsection{{Optimized Molecule Size Analysis}}
\label{appendix:pipeline:addtional}

%
%

\begin{table*}
	\centering
	\caption{{Optimized Molecule Size}}
	\label{tbl:atomsize}
	
	\begin{threeparttable}
		\scriptsize{
			\begin{tabular}{
					@{\hspace{5pt}}l@{\hspace{5pt}}
					@{\hspace{3pt}}r@{\hspace{2pt}}
					@{\hspace{2pt}}r@{\hspace{3pt}}
					@{\hspace{0pt}}c@{\hspace{0pt}}
					@{\hspace{3pt}}r@{\hspace{2pt}}
					@{\hspace{2pt}}r@{\hspace{3pt}}
					@{\hspace{0pt}}c@{\hspace{0pt}}
					@{\hspace{3pt}}r@{\hspace{2pt}}
					@{\hspace{2pt}}c@{\hspace{3pt}}
					@{\hspace{0pt}}c@{\hspace{0pt}}
					@{\hspace{3pt}}c@{\hspace{2pt}}
					@{\hspace{2pt}}c@{\hspace{3pt}}
				}
				\toprule
				\multirow{2}{*}{iter} & \multicolumn{2}{c}{$\delta$=0.0} &&  \multicolumn{2}{c}{$\delta$=0.2} && \multicolumn{2}{c}{$\delta$=0.4} && \multicolumn{2}{c}{$\delta$=0.6} \\
				\cmidrule(lr){2-3} \cmidrule(lr){5-6} \cmidrule(lr){8-9} \cmidrule(lr){11-12}
				&  \#$\atom_x$ & \#$\atom_y$ & & \#$\atom_x$ & \#$\atom_y$ & & \#$\atom_x$ & \#$\atom_y$ & & \#$\atom_x$ & \#$\atom_y$ \\
				\midrule
				1 & 20.51 & 32.70 && 20.51 & 32.64 && 20.51 & 32.13 && 20.51 & 25.25 \\
				2 & 32.82 & 34.45 && 32.71 & 35.09 && 32.26 & 33.60 && 29.08 & 29.78 \\
				3 & 34.26 & 35.28 && 35.16 & 35.83 && 34.63 & 35.02 && 33.03 & 33.30 \\
				4 & 34.91 & 35.83 && 36.00 & 36.23 && 35.05 & 35.43 && 36.20 & 36.30 \\
				5 & 35.51 & 36.20 && 36.41 & 36.50 && 35.78 & 35.94 && 38.00 & 38.00 \\
				\bottomrule
			\end{tabular}%
		}
	\end{threeparttable}
	
\end{table*}

%
%
%
Table~\ref{tbl:atomsize}
presents the average size of the optimized molecules in each iteration with \mbox{$\delta$}=$0.0, 0.2, 0.4$ and $0.6$.
{%
Without any similarity constraints (i.e., \mbox{$\delta$=$0.0$}), 
the average size of the optimized molecules keeps increasing as large as 36.20 
after 5 iterations of optimization (note that the optimized molecules are always constrained to have fewer than 38 atoms).}
{%
In addition, the number of added atoms  (i.e., \mbox{$\text{\#}\atom_y - \text{\#}\atom_x$})
becomes smaller in later iterations (e.g., for $\delta$=$0.0$, 12.19, 1.63, 1.02, 0.92, 0.69
from iteration 1 to 5, respectively).
This indicates that in later iterations, {\molmod} identifies fewer fragments 
that should be removed from the input molecules.
In the meantime, the constraint of optimized molecule size ensures that the newly added fragments in later iterations should not 
increase the size of optimized molecules substantially. 
This also explains the less property improvement (p$_t$$\pm$std) 
with higher similarity constraint (e.g., $\delta$=0.6) in Table~{\ref{tbl:pipeline2}}.
With similarity constraint $\delta$=$0.2, 0.4$ and $0.6$, the size of optimized molecules exhibits 
trends similar to those with $\delta$=$0.0$.}
%

\begin{figure*}
	\centering
	%
	%
	\begin{subfigure}{.43\linewidth}
		\centering
		\vspace{10pt}
		\caption{}
		\begingroup
		\makeatletter
		\providecommand\color[2][]{%
			\GenericError{(gnuplot) \space\space\space\@spaces}{%
				Package color not loaded in conjunction with
				terminal option `colourtext'%
			}{See the gnuplot documentation for explanation.%
			}{Either use 'blacktext' in gnuplot or load the package
				color.sty in LaTeX.}%
			\renewcommand\color[2][]{}%
		}%
		\providecommand\includegraphics[2][]{%
			\GenericError{(gnuplot) \space\space\space\@spaces}{%
				Package graphicx or graphics not loaded%
			}{See the gnuplot documentation for explanation.%
			}{The gnuplot epslatex terminal needs graphicx.sty or graphics.sty.}%
			\renewcommand\includegraphics[2][]{}%
		}%
		\providecommand\rotatebox[2]{#2}%
		\@ifundefined{ifGPcolor}{%
			\newif\ifGPcolor
			\GPcolorfalse
		}{}%
		\@ifundefined{ifGPblacktext}{%
			\newif\ifGPblacktext
			\GPblacktexttrue
		}{}%
		\let\gplgaddtomacro\g@addto@macro
		\gdef\gplbacktext{}%
		\gdef\gplfronttext{}%
		\makeatother
		\ifGPblacktext
		\def\colorrgb#1{}%
		\def\colorgray#1{}%
		\else
		\ifGPcolor
		\def\colorrgb#1{\color[rgb]{#1}}%
		\def\colorgray#1{\color[gray]{#1}}%
		\expandafter\def\csname LTw\endcsname{\color{white}}%
		\expandafter\def\csname LTb\endcsname{\color{black}}%
		\expandafter\def\csname LTa\endcsname{\color{black}}%
		\expandafter\def\csname LT0\endcsname{\color[rgb]{1,0,0}}%
		\expandafter\def\csname LT1\endcsname{\color[rgb]{0,1,0}}%
		\expandafter\def\csname LT2\endcsname{\color[rgb]{0,0,1}}%
		\expandafter\def\csname LT3\endcsname{\color[rgb]{1,0,1}}%
		\expandafter\def\csname LT4\endcsname{\color[rgb]{0,1,1}}%
		\expandafter\def\csname LT5\endcsname{\color[rgb]{1,1,0}}%
		\expandafter\def\csname LT6\endcsname{\color[rgb]{0,0,0}}%
		\expandafter\def\csname LT7\endcsname{\color[rgb]{1,0.3,0}}%
		\expandafter\def\csname LT8\endcsname{\color[rgb]{0.5,0.5,0.5}}%
		\else
		\def\colorrgb#1{\color{black}}%
		\def\colorgray#1{\color[gray]{#1}}%
		\expandafter\def\csname LTw\endcsname{\color{white}}%
		\expandafter\def\csname LTb\endcsname{\color{black}}%
		\expandafter\def\csname LTa\endcsname{\color{black}}%
		\expandafter\def\csname LT0\endcsname{\color{black}}%
		\expandafter\def\csname LT1\endcsname{\color{black}}%
		\expandafter\def\csname LT2\endcsname{\color{black}}%
		\expandafter\def\csname LT3\endcsname{\color{black}}%
		\expandafter\def\csname LT4\endcsname{\color{black}}%
		\expandafter\def\csname LT5\endcsname{\color{black}}%
		\expandafter\def\csname LT6\endcsname{\color{black}}%
		\expandafter\def\csname LT7\endcsname{\color{black}}%
		\expandafter\def\csname LT8\endcsname{\color{black}}%
		\fi
		\fi
		\setlength{\unitlength}{0.0500bp}%
		\ifx\gptboxheight\undefined%
		\newlength{\gptboxheight}%
		\newlength{\gptboxwidth}%
		\newsavebox{\gptboxtext}%
		\fi%
		\setlength{\fboxrule}{0.5pt}%
		\setlength{\fboxsep}{1pt}%
		\begin{picture}(3600.00,1512.00)%
		\gplgaddtomacro\gplbacktext{%
			\csname LTb\endcsname
			\put(-27,22){\makebox(0,0)[r]{\strut{}\footnotesize{0}}}%
			\put(-27,316){\makebox(0,0)[r]{\strut{}\footnotesize{10}}}%
			\put(-27,609){\makebox(0,0)[r]{\strut{}\footnotesize{20}}}%
			\put(-27,903){\makebox(0,0)[r]{\strut{}\footnotesize{30}}}%
			\put(-27,1196){\makebox(0,0)[r]{\strut{}\footnotesize{40}}}%
			\put(142,-88){\makebox(0,0){\strut{}\footnotesize{-10}}}%
			\put(331,-88){\makebox(0,0){\strut{}\footnotesize{-9}}}%
			\put(521,-88){\makebox(0,0){\strut{}\footnotesize{-8}}}%
			\put(710,-88){\makebox(0,0){\strut{}\footnotesize{-7}}}%
			\put(900,-88){\makebox(0,0){\strut{}\footnotesize{-6}}}%
			\put(1089,-88){\makebox(0,0){\strut{}\footnotesize{-5}}}%
			\put(1279,-88){\makebox(0,0){\strut{}\footnotesize{-4}}}%
			\put(1468,-88){\makebox(0,0){\strut{}\footnotesize{-3}}}%
			\put(1657,-88){\makebox(0,0){\strut{}\footnotesize{-2}}}%
			\put(1847,-88){\makebox(0,0){\strut{}\footnotesize{-1}}}%
			\put(2036,-88){\makebox(0,0){\strut{}\footnotesize{0}}}%
			\put(2226,-88){\makebox(0,0){\strut{}\footnotesize{1}}}%
			\put(2415,-88){\makebox(0,0){\strut{}\footnotesize{2}}}%
			\put(2605,-88){\makebox(0,0){\strut{}\footnotesize{3}}}%
			\put(2794,-88){\makebox(0,0){\strut{}\footnotesize{4}}}%
			\put(2983,-88){\makebox(0,0){\strut{}\footnotesize{5}}}%
			\put(3173,-88){\makebox(0,0){\strut{}\footnotesize{6}}}%
			\put(3362,-88){\makebox(0,0){\strut{}\footnotesize{7}}}%
			\put(3552,-88){\makebox(0,0){\strut{}\footnotesize{8}}}%
		}%
		\gplgaddtomacro\gplfronttext{%
			\csname LTb\endcsname
			\put(-352,756){\rotatebox{-270}{\makebox(0,0){\strut{}\scriptsize{\% molecules}}}}%
			\put(1799,-308){\makebox(0,0){\strut{}\scriptsize{\plogp}}}%
			\csname LTb\endcsname
			\put(3467,1317){\makebox(0,0)[r]{\strut{}~\molx}}%
			\csname LTb\endcsname
			\put(3467,1097){\makebox(0,0)[r]{\strut{}~\moly}}%
		}%
		\gplbacktext
		\put(0,0){\includegraphics{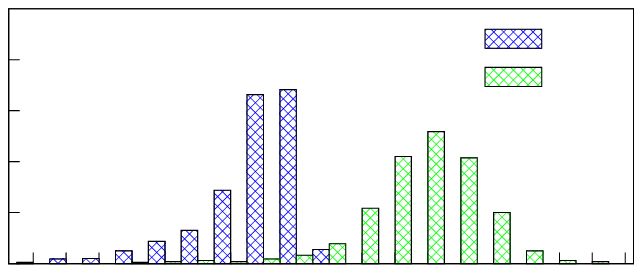}}%
		\gplfronttext
		\end{picture}%
		\endgroup
		\vspace{20pt}
		\label{fig:norm_plogp}
	\end{subfigure}
	~~
	\begin{subfigure}{.43\linewidth}
		\centering
		\vspace{10pt}
		\caption{}
		\begingroup
		\makeatletter
		\providecommand\color[2][]{%
			\GenericError{(gnuplot) \space\space\space\@spaces}{%
				Package color not loaded in conjunction with
				terminal option `colourtext'%
			}{See the gnuplot documentation for explanation.%
			}{Either use 'blacktext' in gnuplot or load the package
				color.sty in LaTeX.}%
			\renewcommand\color[2][]{}%
		}%
		\providecommand\includegraphics[2][]{%
			\GenericError{(gnuplot) \space\space\space\@spaces}{%
				Package graphicx or graphics not loaded%
			}{See the gnuplot documentation for explanation.%
			}{The gnuplot epslatex terminal needs graphicx.sty or graphics.sty.}%
			\renewcommand\includegraphics[2][]{}%
		}%
		\providecommand\rotatebox[2]{#2}%
		\@ifundefined{ifGPcolor}{%
			\newif\ifGPcolor
			\GPcolorfalse
		}{}%
		\@ifundefined{ifGPblacktext}{%
			\newif\ifGPblacktext
			\GPblacktexttrue
		}{}%
		\let\gplgaddtomacro\g@addto@macro
		\gdef\gplbacktext{}%
		\gdef\gplfronttext{}%
		\makeatother
		\ifGPblacktext
		\def\colorrgb#1{}%
		\def\colorgray#1{}%
		\else
		\ifGPcolor
		\def\colorrgb#1{\color[rgb]{#1}}%
		\def\colorgray#1{\color[gray]{#1}}%
		\expandafter\def\csname LTw\endcsname{\color{white}}%
		\expandafter\def\csname LTb\endcsname{\color{black}}%
		\expandafter\def\csname LTa\endcsname{\color{black}}%
		\expandafter\def\csname LT0\endcsname{\color[rgb]{1,0,0}}%
		\expandafter\def\csname LT1\endcsname{\color[rgb]{0,1,0}}%
		\expandafter\def\csname LT2\endcsname{\color[rgb]{0,0,1}}%
		\expandafter\def\csname LT3\endcsname{\color[rgb]{1,0,1}}%
		\expandafter\def\csname LT4\endcsname{\color[rgb]{0,1,1}}%
		\expandafter\def\csname LT5\endcsname{\color[rgb]{1,1,0}}%
		\expandafter\def\csname LT6\endcsname{\color[rgb]{0,0,0}}%
		\expandafter\def\csname LT7\endcsname{\color[rgb]{1,0.3,0}}%
		\expandafter\def\csname LT8\endcsname{\color[rgb]{0.5,0.5,0.5}}%
		\else
		\def\colorrgb#1{\color{black}}%
		\def\colorgray#1{\color[gray]{#1}}%
		\expandafter\def\csname LTw\endcsname{\color{white}}%
		\expandafter\def\csname LTb\endcsname{\color{black}}%
		\expandafter\def\csname LTa\endcsname{\color{black}}%
		\expandafter\def\csname LT0\endcsname{\color{black}}%
		\expandafter\def\csname LT1\endcsname{\color{black}}%
		\expandafter\def\csname LT2\endcsname{\color{black}}%
		\expandafter\def\csname LT3\endcsname{\color{black}}%
		\expandafter\def\csname LT4\endcsname{\color{black}}%
		\expandafter\def\csname LT5\endcsname{\color{black}}%
		\expandafter\def\csname LT6\endcsname{\color{black}}%
		\expandafter\def\csname LT7\endcsname{\color{black}}%
		\expandafter\def\csname LT8\endcsname{\color{black}}%
		\fi
		\fi
		\setlength{\unitlength}{0.0500bp}%
		\ifx\gptboxheight\undefined%
		\newlength{\gptboxheight}%
		\newlength{\gptboxwidth}%
		\newsavebox{\gptboxtext}%
		\fi%
		\setlength{\fboxrule}{0.5pt}%
		\setlength{\fboxsep}{1pt}%
		\begin{picture}(3600.00,1512.00)%
		\gplgaddtomacro\gplbacktext{%
			\csname LTb\endcsname
			\put(-27,22){\makebox(0,0)[r]{\strut{}\footnotesize{0}}}%
			\put(-27,316){\makebox(0,0)[r]{\strut{}\footnotesize{10}}}%
			\put(-27,609){\makebox(0,0)[r]{\strut{}\footnotesize{20}}}%
			\put(-27,903){\makebox(0,0)[r]{\strut{}\footnotesize{30}}}%
			\put(-27,1196){\makebox(0,0)[r]{\strut{}\footnotesize{40}}}%
			\put(142,-88){\makebox(0,0){\strut{}\footnotesize{-10}}}%
			\put(331,-88){\makebox(0,0){\strut{}\footnotesize{-9}}}%
			\put(521,-88){\makebox(0,0){\strut{}\footnotesize{-8}}}%
			\put(710,-88){\makebox(0,0){\strut{}\footnotesize{-7}}}%
			\put(900,-88){\makebox(0,0){\strut{}\footnotesize{-6}}}%
			\put(1089,-88){\makebox(0,0){\strut{}\footnotesize{-5}}}%
			\put(1279,-88){\makebox(0,0){\strut{}\footnotesize{-4}}}%
			\put(1468,-88){\makebox(0,0){\strut{}\footnotesize{-3}}}%
			\put(1657,-88){\makebox(0,0){\strut{}\footnotesize{-2}}}%
			\put(1847,-88){\makebox(0,0){\strut{}\footnotesize{-1}}}%
			\put(2036,-88){\makebox(0,0){\strut{}\footnotesize{0}}}%
			\put(2226,-88){\makebox(0,0){\strut{}\footnotesize{1}}}%
			\put(2415,-88){\makebox(0,0){\strut{}\footnotesize{2}}}%
			\put(2605,-88){\makebox(0,0){\strut{}\footnotesize{3}}}%
			\put(2794,-88){\makebox(0,0){\strut{}\footnotesize{4}}}%
			\put(2983,-88){\makebox(0,0){\strut{}\footnotesize{5}}}%
			\put(3173,-88){\makebox(0,0){\strut{}\footnotesize{6}}}%
			\put(3362,-88){\makebox(0,0){\strut{}\footnotesize{7}}}%
			\put(3552,-88){\makebox(0,0){\strut{}\footnotesize{8}}}%
		}%
		\gplgaddtomacro\gplfronttext{%
			\csname LTb\endcsname
			\put(-352,756){\rotatebox{-270}{\makebox(0,0){\strut{}\scriptsize{\% molecules}}}}%
			\put(1799,-308){\makebox(0,0){\strut{}\scriptsize{\logp}}}%
			\csname LTb\endcsname
			\put(3467,1317){\makebox(0,0)[r]{\strut{}~\molx}}%
			\csname LTb\endcsname
			\put(3467,1097){\makebox(0,0)[r]{\strut{}~\moly}}%
		}%
		\gplbacktext
		\put(0,0){\includegraphics{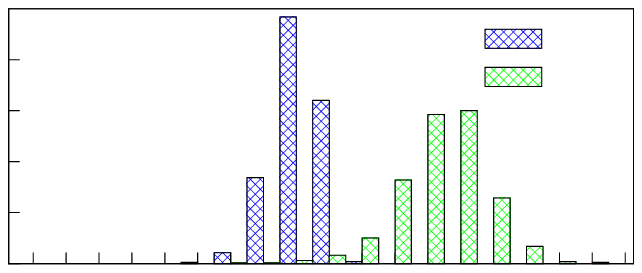}}%
		\gplfronttext
		\end{picture}%
		\endgroup
		\vspace{20pt}
		\label{fig:norm_logp}
	\end{subfigure}%
	\\
	\begin{subfigure}{.43\linewidth}
		\centering
		\vspace{-10pt}
		\caption{}
		%
		\begingroup
		\makeatletter
		\providecommand\color[2][]{%
			\GenericError{(gnuplot) \space\space\space\@spaces}{%
				Package color not loaded in conjunction with
				terminal option `colourtext'%
			}{See the gnuplot documentation for explanation.%
			}{Either use 'blacktext' in gnuplot or load the package
				color.sty in LaTeX.}%
			\renewcommand\color[2][]{}%
		}%
		\providecommand\includegraphics[2][]{%
			\GenericError{(gnuplot) \space\space\space\@spaces}{%
				Package graphicx or graphics not loaded%
			}{See the gnuplot documentation for explanation.%
			}{The gnuplot epslatex terminal needs graphicx.sty or graphics.sty.}%
			\renewcommand\includegraphics[2][]{}%
		}%
		\providecommand\rotatebox[2]{#2}%
		\@ifundefined{ifGPcolor}{%
			\newif\ifGPcolor
			\GPcolorfalse
		}{}%
		\@ifundefined{ifGPblacktext}{%
			\newif\ifGPblacktext
			\GPblacktexttrue
		}{}%
		\let\gplgaddtomacro\g@addto@macro
		\gdef\gplbacktext{}%
		\gdef\gplfronttext{}%
		\makeatother
		\ifGPblacktext
		\def\colorrgb#1{}%
		\def\colorgray#1{}%
		\else
		\ifGPcolor
		\def\colorrgb#1{\color[rgb]{#1}}%
		\def\colorgray#1{\color[gray]{#1}}%
		\expandafter\def\csname LTw\endcsname{\color{white}}%
		\expandafter\def\csname LTb\endcsname{\color{black}}%
		\expandafter\def\csname LTa\endcsname{\color{black}}%
		\expandafter\def\csname LT0\endcsname{\color[rgb]{1,0,0}}%
		\expandafter\def\csname LT1\endcsname{\color[rgb]{0,1,0}}%
		\expandafter\def\csname LT2\endcsname{\color[rgb]{0,0,1}}%
		\expandafter\def\csname LT3\endcsname{\color[rgb]{1,0,1}}%
		\expandafter\def\csname LT4\endcsname{\color[rgb]{0,1,1}}%
		\expandafter\def\csname LT5\endcsname{\color[rgb]{1,1,0}}%
		\expandafter\def\csname LT6\endcsname{\color[rgb]{0,0,0}}%
		\expandafter\def\csname LT7\endcsname{\color[rgb]{1,0.3,0}}%
		\expandafter\def\csname LT8\endcsname{\color[rgb]{0.5,0.5,0.5}}%
		\else
		\def\colorrgb#1{\color{black}}%
		\def\colorgray#1{\color[gray]{#1}}%
		\expandafter\def\csname LTw\endcsname{\color{white}}%
		\expandafter\def\csname LTb\endcsname{\color{black}}%
		\expandafter\def\csname LTa\endcsname{\color{black}}%
		\expandafter\def\csname LT0\endcsname{\color{black}}%
		\expandafter\def\csname LT1\endcsname{\color{black}}%
		\expandafter\def\csname LT2\endcsname{\color{black}}%
		\expandafter\def\csname LT3\endcsname{\color{black}}%
		\expandafter\def\csname LT4\endcsname{\color{black}}%
		\expandafter\def\csname LT5\endcsname{\color{black}}%
		\expandafter\def\csname LT6\endcsname{\color{black}}%
		\expandafter\def\csname LT7\endcsname{\color{black}}%
		\expandafter\def\csname LT8\endcsname{\color{black}}%
		\fi
		\fi
		\setlength{\unitlength}{0.0500bp}%
		\ifx\gptboxheight\undefined%
		\newlength{\gptboxheight}%
		\newlength{\gptboxwidth}%
		\newsavebox{\gptboxtext}%
		\fi%
		\setlength{\fboxrule}{0.5pt}%
		\setlength{\fboxsep}{1pt}%
		\begin{picture}(3600.00,1512.00)%
		\gplgaddtomacro\gplbacktext{%
			\csname LTb\endcsname
			\put(-27,22){\makebox(0,0)[r]{\strut{}\footnotesize{0}}}%
			\put(-27,316){\makebox(0,0)[r]{\strut{}\footnotesize{10}}}%
			\put(-27,609){\makebox(0,0)[r]{\strut{}\footnotesize{20}}}%
			\put(-27,903){\makebox(0,0)[r]{\strut{}\footnotesize{30}}}%
			\put(-27,1196){\makebox(0,0)[r]{\strut{}\footnotesize{40}}}%
			\put(142,-88){\makebox(0,0){\strut{}\footnotesize{-10}}}%
			\put(331,-88){\makebox(0,0){\strut{}\footnotesize{-9}}}%
			\put(521,-88){\makebox(0,0){\strut{}\footnotesize{-8}}}%
			\put(710,-88){\makebox(0,0){\strut{}\footnotesize{-7}}}%
			\put(900,-88){\makebox(0,0){\strut{}\footnotesize{-6}}}%
			\put(1089,-88){\makebox(0,0){\strut{}\footnotesize{-5}}}%
			\put(1279,-88){\makebox(0,0){\strut{}\footnotesize{-4}}}%
			\put(1468,-88){\makebox(0,0){\strut{}\footnotesize{-3}}}%
			\put(1657,-88){\makebox(0,0){\strut{}\footnotesize{-2}}}%
			\put(1847,-88){\makebox(0,0){\strut{}\footnotesize{-1}}}%
			\put(2036,-88){\makebox(0,0){\strut{}\footnotesize{0}}}%
			\put(2226,-88){\makebox(0,0){\strut{}\footnotesize{1}}}%
			\put(2415,-88){\makebox(0,0){\strut{}\footnotesize{2}}}%
			\put(2605,-88){\makebox(0,0){\strut{}\footnotesize{3}}}%
			\put(2794,-88){\makebox(0,0){\strut{}\footnotesize{4}}}%
			\put(2983,-88){\makebox(0,0){\strut{}\footnotesize{5}}}%
			\put(3173,-88){\makebox(0,0){\strut{}\footnotesize{6}}}%
			\put(3362,-88){\makebox(0,0){\strut{}\footnotesize{7}}}%
			\put(3552,-88){\makebox(0,0){\strut{}\footnotesize{8}}}%
		}%
		\gplgaddtomacro\gplfronttext{%
			\csname LTb\endcsname
			\put(-352,756){\rotatebox{-270}{\makebox(0,0){\strut{}\scriptsize{\% molecules}}}}%
			\put(1799,-308){\makebox(0,0){\strut{}\scriptsize{normalized SA}}}%
			\csname LTb\endcsname
			\put(3467,1317){\makebox(0,0)[r]{\strut{}~\molx}}%
			\csname LTb\endcsname
			\put(3467,1097){\makebox(0,0)[r]{\strut{}~\moly}}%
		}%
		\gplbacktext
		\put(0,0){\includegraphics{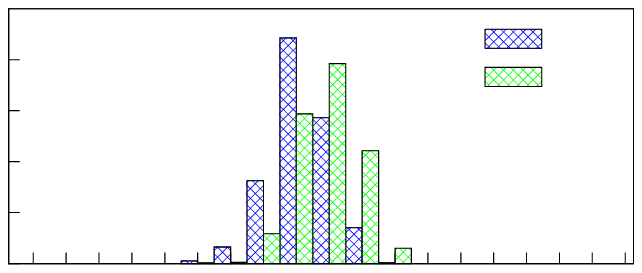}}%
		\gplfronttext
		\end{picture}%
		\endgroup
		\label{fig:norm_SA}
	\end{subfigure}%
	~~
	\begin{subfigure}{.43\linewidth}
		\centering
		\vspace{10pt}
		\caption{}
		\begingroup
		\makeatletter
		\providecommand\color[2][]{%
			\GenericError{(gnuplot) \space\space\space\@spaces}{%
				Package color not loaded in conjunction with
				terminal option `colourtext'%
			}{See the gnuplot documentation for explanation.%
			}{Either use 'blacktext' in gnuplot or load the package
				color.sty in LaTeX.}%
			\renewcommand\color[2][]{}%
		}%
		\providecommand\includegraphics[2][]{%
			\GenericError{(gnuplot) \space\space\space\@spaces}{%
				Package graphicx or graphics not loaded%
			}{See the gnuplot documentation for explanation.%
			}{The gnuplot epslatex terminal needs graphicx.sty or graphics.sty.}%
			\renewcommand\includegraphics[2][]{}%
		}%
		\providecommand\rotatebox[2]{#2}%
		\@ifundefined{ifGPcolor}{%
			\newif\ifGPcolor
			\GPcolorfalse
		}{}%
		\@ifundefined{ifGPblacktext}{%
			\newif\ifGPblacktext
			\GPblacktexttrue
		}{}%
		\let\gplgaddtomacro\g@addto@macro
		\gdef\gplbacktext{}%
		\gdef\gplfronttext{}%
		\makeatother
		\ifGPblacktext
		\def\colorrgb#1{}%
		\def\colorgray#1{}%
		\else
		\ifGPcolor
		\def\colorrgb#1{\color[rgb]{#1}}%
		\def\colorgray#1{\color[gray]{#1}}%
		\expandafter\def\csname LTw\endcsname{\color{white}}%
		\expandafter\def\csname LTb\endcsname{\color{black}}%
		\expandafter\def\csname LTa\endcsname{\color{black}}%
		\expandafter\def\csname LT0\endcsname{\color[rgb]{1,0,0}}%
		\expandafter\def\csname LT1\endcsname{\color[rgb]{0,1,0}}%
		\expandafter\def\csname LT2\endcsname{\color[rgb]{0,0,1}}%
		\expandafter\def\csname LT3\endcsname{\color[rgb]{1,0,1}}%
		\expandafter\def\csname LT4\endcsname{\color[rgb]{0,1,1}}%
		\expandafter\def\csname LT5\endcsname{\color[rgb]{1,1,0}}%
		\expandafter\def\csname LT6\endcsname{\color[rgb]{0,0,0}}%
		\expandafter\def\csname LT7\endcsname{\color[rgb]{1,0.3,0}}%
		\expandafter\def\csname LT8\endcsname{\color[rgb]{0.5,0.5,0.5}}%
		\else
		\def\colorrgb#1{\color{black}}%
		\def\colorgray#1{\color[gray]{#1}}%
		\expandafter\def\csname LTw\endcsname{\color{white}}%
		\expandafter\def\csname LTb\endcsname{\color{black}}%
		\expandafter\def\csname LTa\endcsname{\color{black}}%
		\expandafter\def\csname LT0\endcsname{\color{black}}%
		\expandafter\def\csname LT1\endcsname{\color{black}}%
		\expandafter\def\csname LT2\endcsname{\color{black}}%
		\expandafter\def\csname LT3\endcsname{\color{black}}%
		\expandafter\def\csname LT4\endcsname{\color{black}}%
		\expandafter\def\csname LT5\endcsname{\color{black}}%
		\expandafter\def\csname LT6\endcsname{\color{black}}%
		\expandafter\def\csname LT7\endcsname{\color{black}}%
		\expandafter\def\csname LT8\endcsname{\color{black}}%
		\fi
		\fi
		\setlength{\unitlength}{0.0500bp}%
		\ifx\gptboxheight\undefined%
		\newlength{\gptboxheight}%
		\newlength{\gptboxwidth}%
		\newsavebox{\gptboxtext}%
		\fi%
		\setlength{\fboxrule}{0.5pt}%
		\setlength{\fboxsep}{1pt}%
		\begin{picture}(3600.00,1512.00)%
		\gplgaddtomacro\gplbacktext{%
			\csname LTb\endcsname
			\put(-27,22){\makebox(0,0)[r]{\strut{}\footnotesize{0}}}%
			\put(-27,316){\makebox(0,0)[r]{\strut{}\footnotesize{10}}}%
			\put(-27,609){\makebox(0,0)[r]{\strut{}\footnotesize{20}}}%
			\put(-27,903){\makebox(0,0)[r]{\strut{}\footnotesize{30}}}%
			\put(-27,1196){\makebox(0,0)[r]{\strut{}\footnotesize{40}}}%
			\put(142,-88){\makebox(0,0){\strut{}\footnotesize{-10}}}%
			\put(331,-88){\makebox(0,0){\strut{}\footnotesize{-9}}}%
			\put(521,-88){\makebox(0,0){\strut{}\footnotesize{-8}}}%
			\put(710,-88){\makebox(0,0){\strut{}\footnotesize{-7}}}%
			\put(900,-88){\makebox(0,0){\strut{}\footnotesize{-6}}}%
			\put(1089,-88){\makebox(0,0){\strut{}\footnotesize{-5}}}%
			\put(1279,-88){\makebox(0,0){\strut{}\footnotesize{-4}}}%
			\put(1468,-88){\makebox(0,0){\strut{}\footnotesize{-3}}}%
			\put(1657,-88){\makebox(0,0){\strut{}\footnotesize{-2}}}%
			\put(1847,-88){\makebox(0,0){\strut{}\footnotesize{-1}}}%
			\put(2036,-88){\makebox(0,0){\strut{}\footnotesize{0}}}%
			\put(2226,-88){\makebox(0,0){\strut{}\footnotesize{1}}}%
			\put(2415,-88){\makebox(0,0){\strut{}\footnotesize{2}}}%
			\put(2605,-88){\makebox(0,0){\strut{}\footnotesize{3}}}%
			\put(2794,-88){\makebox(0,0){\strut{}\footnotesize{4}}}%
			\put(2983,-88){\makebox(0,0){\strut{}\footnotesize{5}}}%
			\put(3173,-88){\makebox(0,0){\strut{}\footnotesize{6}}}%
			\put(3362,-88){\makebox(0,0){\strut{}\footnotesize{7}}}%
			\put(3552,-88){\makebox(0,0){\strut{}\footnotesize{8}}}%
		}%
		\gplgaddtomacro\gplfronttext{%
			\csname LTb\endcsname
			\put(-352,756){\rotatebox{-270}{\makebox(0,0){\strut{}\footnotesize{\% molecules}}}}%
			\put(1799,-308){\makebox(0,0){\strut{}\footnotesize{normalized SA + cycle score}}}%
			\csname LTb\endcsname
			\put(3467,1317){\makebox(0,0)[r]{\strut{}~\molx}}%
			\csname LTb\endcsname
			\put(3467,1097){\makebox(0,0)[r]{\strut{}~\moly}}%
		}%
		\gplbacktext
		\put(0,0){\includegraphics{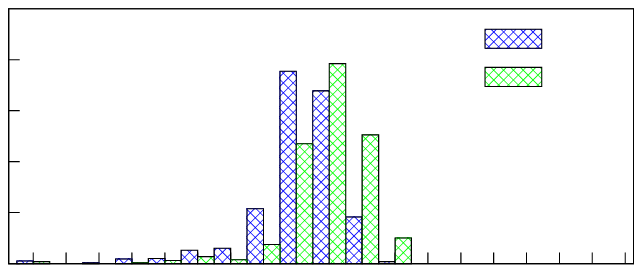}}%
		\gplfronttext
		\end{picture}%
		\endgroup
		\vspace{20pt}
		\label{fig:norm_SA_circle}
	\end{subfigure}%
	\vspace{-5pt}
	\caption{\textbf{Test Molecule Property Distributions before and after {\pipeline} Optimization}.
		\textbf{a,} \plogp distribution.
		\textbf{b,} \logp distribution
		\textbf{c,} Normalized synthetic accessibility distribution.
		\textbf{d,} Synthetic accessibility + cycle score distribution.}
	\label{fig:score_distribution}
	\vspace{5pt}
\end{figure*}

\subsection{{Improvement on Different Components of \plogp}}
\label{appendix:pipeline:sa}

{Recall that {\plogp} has three components: octanol-water partition coefficients ({\logp}), 
synthetic accessibility (SA; measured by normalized SA scores~{\cite{ertl2009estimation}}) 
and ring size (measured as cycle scores = \mbox{$-(\min(0, \max(\text{ring size}) - 6)$}), 
that is, \mbox{{\plogp} = {\logp} + SA score + cycle score}. 
Fig.~{\ref{fig:score_distribution}} presents the distributions of {\plogp} values (Fig.~{\ref{fig:norm_plogp}}), {\logp} values
(Fig.~{\ref{fig:norm_logp}}), SA scores (Fig.~{\ref{fig:norm_SA}}) and 
the combined values of SA and circle scores (Fig.~{\ref{fig:norm_SA_circle}}) among the test molecules 
before and after {\pipeline} optimization ($\delta$=0.4).}   

%

%

\begin{figure*}
	\centering
	\begin{subfigure}{\linewidth}
		\centering
		\begingroup
		\makeatletter
		\providecommand\color[2][]{%
			\GenericError{(gnuplot) \space\space\space\@spaces}{%
				Package color not loaded in conjunction with
				terminal option `colourtext'%
			}{See the gnuplot documentation for explanation.%
			}{Either use 'blacktext' in gnuplot or load the package
				color.sty in LaTeX.}%
			\renewcommand\color[2][]{}%
		}%
		\providecommand\includegraphics[2][]{%
			\GenericError{(gnuplot) \space\space\space\@spaces}{%
				Package graphicx or graphics not loaded%
			}{See the gnuplot documentation for explanation.%
			}{The gnuplot epslatex terminal needs graphicx.sty or graphics.sty.}%
			\renewcommand\includegraphics[2][]{}%
		}%
		\providecommand\rotatebox[2]{#2}%
		\@ifundefined{ifGPcolor}{%
			\newif\ifGPcolor
			\GPcolortrue
		}{}%
		\@ifundefined{ifGPblacktext}{%
			\newif\ifGPblacktext
			\GPblacktexttrue
		}{}%
		\let\gplgaddtomacro\g@addto@macro
		\gdef\gplbacktext{}%
		\gdef\gplfronttext{}%
		\makeatother
		\ifGPblacktext
		\def\colorrgb#1{}%
		\def\colorgray#1{}%
		\else
		\ifGPcolor
		\def\colorrgb#1{\color[rgb]{#1}}%
		\def\colorgray#1{\color[gray]{#1}}%
		\expandafter\def\csname LTw\endcsname{\color{white}}%
		\expandafter\def\csname LTb\endcsname{\color{black}}%
		\expandafter\def\csname LTa\endcsname{\color{black}}%
		\expandafter\def\csname LT0\endcsname{\color[rgb]{1,0,0}}%
		\expandafter\def\csname LT1\endcsname{\color[rgb]{0,1,0}}%
		\expandafter\def\csname LT2\endcsname{\color[rgb]{0,0,1}}%
		\expandafter\def\csname LT3\endcsname{\color[rgb]{1,0,1}}%
		\expandafter\def\csname LT4\endcsname{\color[rgb]{0,1,1}}%
		\expandafter\def\csname LT5\endcsname{\color[rgb]{1,1,0}}%
		\expandafter\def\csname LT6\endcsname{\color[rgb]{0,0,0}}%
		\expandafter\def\csname LT7\endcsname{\color[rgb]{1,0.3,0}}%
		\expandafter\def\csname LT8\endcsname{\color[rgb]{0.5,0.5,0.5}}%
		\else
		\def\colorrgb#1{\color{black}}%
		\def\colorgray#1{\color[gray]{#1}}%
		\expandafter\def\csname LTw\endcsname{\color{white}}%
		\expandafter\def\csname LTb\endcsname{\color{black}}%
		\expandafter\def\csname LTa\endcsname{\color{black}}%
		\expandafter\def\csname LT0\endcsname{\color{black}}%
		\expandafter\def\csname LT1\endcsname{\color{black}}%
		\expandafter\def\csname LT2\endcsname{\color{black}}%
		\expandafter\def\csname LT3\endcsname{\color{black}}%
		\expandafter\def\csname LT4\endcsname{\color{black}}%
		\expandafter\def\csname LT5\endcsname{\color{black}}%
		\expandafter\def\csname LT6\endcsname{\color{black}}%
		\expandafter\def\csname LT7\endcsname{\color{black}}%
		\expandafter\def\csname LT8\endcsname{\color{black}}%
		\fi
		\fi
		\setlength{\unitlength}{0.0500bp}%
		\ifx\gptboxheight\undefined%
		\newlength{\gptboxheight}%
		\newlength{\gptboxwidth}%
		\newsavebox{\gptboxtext}%
		\fi%
		\setlength{\fboxrule}{0.5pt}%
		\setlength{\fboxsep}{1pt}%
		\begin{picture}(7200.00,720.00)%
		\gplgaddtomacro\gplbacktext{%
		}%
		\gplgaddtomacro\gplfronttext{%
			\csname LTb\endcsname
			\put(3008,547){\makebox(0,0)[r]{\strut{}\scriptsize{test/optimized molecules}}}%
			\csname LTb\endcsname
			\put(5579,547){\makebox(0,0)[r]{\strut{}\scriptsize{ZINC molecues}}}%
		}%
		\gplbacktext
		\put(0,0){\includegraphics{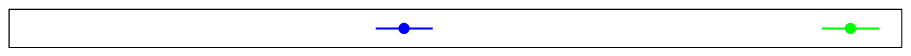}}%
		\gplfronttext
		\end{picture}%
		\endgroup
		
		\vspace{-10pt}
	\end{subfigure}%
	\\
	%
	%
	\begin{subfigure}{.4\linewidth}
		\centering
		\caption{}
		\begingroup
		\makeatletter
		\providecommand\color[2][]{%
			\GenericError{(gnuplot) \space\space\space\@spaces}{%
				Package color not loaded in conjunction with
				terminal option `colourtext'%
			}{See the gnuplot documentation for explanation.%
			}{Either use 'blacktext' in gnuplot or load the package
				color.sty in LaTeX.}%
			\renewcommand\color[2][]{}%
		}%
		\providecommand\includegraphics[2][]{%
			\GenericError{(gnuplot) \space\space\space\@spaces}{%
				Package graphicx or graphics not loaded%
			}{See the gnuplot documentation for explanation.%
			}{The gnuplot epslatex terminal needs graphicx.sty or graphics.sty.}%
			\renewcommand\includegraphics[2][]{}%
		}%
		\providecommand\rotatebox[2]{#2}%
		\@ifundefined{ifGPcolor}{%
			\newif\ifGPcolor
			\GPcolorfalse
		}{}%
		\@ifundefined{ifGPblacktext}{%
			\newif\ifGPblacktext
			\GPblacktexttrue
		}{}%
		\let\gplgaddtomacro\g@addto@macro
		\gdef\gplbacktext{}%
		\gdef\gplfronttext{}%
		\makeatother
		\ifGPblacktext
		\def\colorrgb#1{}%
		\def\colorgray#1{}%
		\else
		\ifGPcolor
		\def\colorrgb#1{\color[rgb]{#1}}%
		\def\colorgray#1{\color[gray]{#1}}%
		\expandafter\def\csname LTw\endcsname{\color{white}}%
		\expandafter\def\csname LTb\endcsname{\color{black}}%
		\expandafter\def\csname LTa\endcsname{\color{black}}%
		\expandafter\def\csname LT0\endcsname{\color[rgb]{1,0,0}}%
		\expandafter\def\csname LT1\endcsname{\color[rgb]{0,1,0}}%
		\expandafter\def\csname LT2\endcsname{\color[rgb]{0,0,1}}%
		\expandafter\def\csname LT3\endcsname{\color[rgb]{1,0,1}}%
		\expandafter\def\csname LT4\endcsname{\color[rgb]{0,1,1}}%
		\expandafter\def\csname LT5\endcsname{\color[rgb]{1,1,0}}%
		\expandafter\def\csname LT6\endcsname{\color[rgb]{0,0,0}}%
		\expandafter\def\csname LT7\endcsname{\color[rgb]{1,0.3,0}}%
		\expandafter\def\csname LT8\endcsname{\color[rgb]{0.5,0.5,0.5}}%
		\else
		\def\colorrgb#1{\color{black}}%
		\def\colorgray#1{\color[gray]{#1}}%
		\expandafter\def\csname LTw\endcsname{\color{white}}%
		\expandafter\def\csname LTb\endcsname{\color{black}}%
		\expandafter\def\csname LTa\endcsname{\color{black}}%
		\expandafter\def\csname LT0\endcsname{\color{black}}%
		\expandafter\def\csname LT1\endcsname{\color{black}}%
		\expandafter\def\csname LT2\endcsname{\color{black}}%
		\expandafter\def\csname LT3\endcsname{\color{black}}%
		\expandafter\def\csname LT4\endcsname{\color{black}}%
		\expandafter\def\csname LT5\endcsname{\color{black}}%
		\expandafter\def\csname LT6\endcsname{\color{black}}%
		\expandafter\def\csname LT7\endcsname{\color{black}}%
		\expandafter\def\csname LT8\endcsname{\color{black}}%
		\fi
		\fi
		\setlength{\unitlength}{0.0500bp}%
		\ifx\gptboxheight\undefined%
		\newlength{\gptboxheight}%
		\newlength{\gptboxwidth}%
		\newsavebox{\gptboxtext}%
		\fi%
		\setlength{\fboxrule}{0.5pt}%
		\setlength{\fboxsep}{1pt}%
		\begin{picture}(2880.00,1663.20)%
		\gplgaddtomacro\gplbacktext{%
			\csname LTb\endcsname
			\put(-66,317){\makebox(0,0)[r]{\strut{}\scriptsize{2.0}}}%
			\put(-66,611){\makebox(0,0)[r]{\strut{}\scriptsize{4.0}}}%
			\put(-66,905){\makebox(0,0)[r]{\strut{}\scriptsize{6.0}}}%
			\put(-66,1199){\makebox(0,0)[r]{\strut{}\scriptsize{8.0}}}%
			\put(-66,1493){\makebox(0,0)[r]{\strut{}\scriptsize{10.0}}}%
			\put(1,-88){\makebox(0,0){\strut{}\scriptsize{0}}}%
			\put(510,-88){\makebox(0,0){\strut{}\scriptsize{10}}}%
			\put(1019,-88){\makebox(0,0){\strut{}\scriptsize{20}}}%
			\put(1529,-88){\makebox(0,0){\strut{}\scriptsize{30}}}%
			\put(2038,-88){\makebox(0,0){\strut{}\scriptsize{40}}}%
			\put(2548,-88){\makebox(0,0){\strut{}\scriptsize{50}}}%
		}%
		\gplgaddtomacro\gplfronttext{%
			\csname LTb\endcsname
			\put(-484,831){\rotatebox{-270}{\makebox(0,0){\strut{}\scriptsize{\% in cluster}}}}%
			\put(1439,-264){\makebox(0,0){\strut{}\scriptsize{cluster ID}}}%
		}%
		\gplbacktext
		\put(0,0){\includegraphics{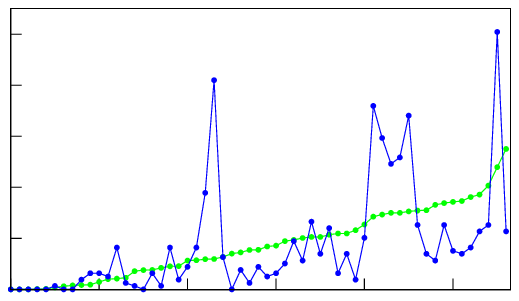}}%
		\gplfronttext
		\end{picture}%
		\endgroup
		\vspace{20pt}
		\label{fig:transform:test}
	\end{subfigure}
	\begin{subfigure}{.4\linewidth}
		\centering
		\caption{}
		\begingroup
		\makeatletter
		\providecommand\color[2][]{%
			\GenericError{(gnuplot) \space\space\space\@spaces}{%
				Package color not loaded in conjunction with
				terminal option `colourtext'%
			}{See the gnuplot documentation for explanation.%
			}{Either use 'blacktext' in gnuplot or load the package
				color.sty in LaTeX.}%
			\renewcommand\color[2][]{}%
		}%
		\providecommand\includegraphics[2][]{%
			\GenericError{(gnuplot) \space\space\space\@spaces}{%
				Package graphicx or graphics not loaded%
			}{See the gnuplot documentation for explanation.%
			}{The gnuplot epslatex terminal needs graphicx.sty or graphics.sty.}%
			\renewcommand\includegraphics[2][]{}%
		}%
		\providecommand\rotatebox[2]{#2}%
		\@ifundefined{ifGPcolor}{%
			\newif\ifGPcolor
			\GPcolorfalse
		}{}%
		\@ifundefined{ifGPblacktext}{%
			\newif\ifGPblacktext
			\GPblacktexttrue
		}{}%
		\let\gplgaddtomacro\g@addto@macro
		\gdef\gplbacktext{}%
		\gdef\gplfronttext{}%
		\makeatother
		\ifGPblacktext
		\def\colorrgb#1{}%
		\def\colorgray#1{}%
		\else
		\ifGPcolor
		\def\colorrgb#1{\color[rgb]{#1}}%
		\def\colorgray#1{\color[gray]{#1}}%
		\expandafter\def\csname LTw\endcsname{\color{white}}%
		\expandafter\def\csname LTb\endcsname{\color{black}}%
		\expandafter\def\csname LTa\endcsname{\color{black}}%
		\expandafter\def\csname LT0\endcsname{\color[rgb]{1,0,0}}%
		\expandafter\def\csname LT1\endcsname{\color[rgb]{0,1,0}}%
		\expandafter\def\csname LT2\endcsname{\color[rgb]{0,0,1}}%
		\expandafter\def\csname LT3\endcsname{\color[rgb]{1,0,1}}%
		\expandafter\def\csname LT4\endcsname{\color[rgb]{0,1,1}}%
		\expandafter\def\csname LT5\endcsname{\color[rgb]{1,1,0}}%
		\expandafter\def\csname LT6\endcsname{\color[rgb]{0,0,0}}%
		\expandafter\def\csname LT7\endcsname{\color[rgb]{1,0.3,0}}%
		\expandafter\def\csname LT8\endcsname{\color[rgb]{0.5,0.5,0.5}}%
		\else
		\def\colorrgb#1{\color{black}}%
		\def\colorgray#1{\color[gray]{#1}}%
		\expandafter\def\csname LTw\endcsname{\color{white}}%
		\expandafter\def\csname LTb\endcsname{\color{black}}%
		\expandafter\def\csname LTa\endcsname{\color{black}}%
		\expandafter\def\csname LT0\endcsname{\color{black}}%
		\expandafter\def\csname LT1\endcsname{\color{black}}%
		\expandafter\def\csname LT2\endcsname{\color{black}}%
		\expandafter\def\csname LT3\endcsname{\color{black}}%
		\expandafter\def\csname LT4\endcsname{\color{black}}%
		\expandafter\def\csname LT5\endcsname{\color{black}}%
		\expandafter\def\csname LT6\endcsname{\color{black}}%
		\expandafter\def\csname LT7\endcsname{\color{black}}%
		\expandafter\def\csname LT8\endcsname{\color{black}}%
		\fi
		\fi
		\setlength{\unitlength}{0.0500bp}%
		\ifx\gptboxheight\undefined%
		\newlength{\gptboxheight}%
		\newlength{\gptboxwidth}%
		\newsavebox{\gptboxtext}%
		\fi%
		\setlength{\fboxrule}{0.5pt}%
		\setlength{\fboxsep}{1pt}%
		\begin{picture}(2880.00,1663.20)%
		\gplgaddtomacro\gplbacktext{%
			\csname LTb\endcsname
			\put(-66,317){\makebox(0,0)[r]{\strut{}\scriptsize{2.0}}}%
			\put(-66,611){\makebox(0,0)[r]{\strut{}\scriptsize{4.0}}}%
			\put(-66,905){\makebox(0,0)[r]{\strut{}\scriptsize{6.0}}}%
			\put(-66,1199){\makebox(0,0)[r]{\strut{}\scriptsize{8.0}}}%
			\put(-66,1493){\makebox(0,0)[r]{\strut{}\scriptsize{10.0}}}%
			\put(1,-88){\makebox(0,0){\strut{}\scriptsize{0}}}%
			\put(510,-88){\makebox(0,0){\strut{}\scriptsize{10}}}%
			\put(1019,-88){\makebox(0,0){\strut{}\scriptsize{20}}}%
			\put(1529,-88){\makebox(0,0){\strut{}\scriptsize{30}}}%
			\put(2038,-88){\makebox(0,0){\strut{}\scriptsize{40}}}%
			\put(2548,-88){\makebox(0,0){\strut{}\scriptsize{50}}}%
		}%
		\gplgaddtomacro\gplfronttext{%
			\csname LTb\endcsname
			\put(-484,831){\rotatebox{-270}{\makebox(0,0){\strut{}\footnotesize{\% in cluster}}}}%
			\put(1439,-264){\makebox(0,0){\strut{}\footnotesize{cluster ID}}}%
		}%
		\gplbacktext
		\put(0,0){\includegraphics{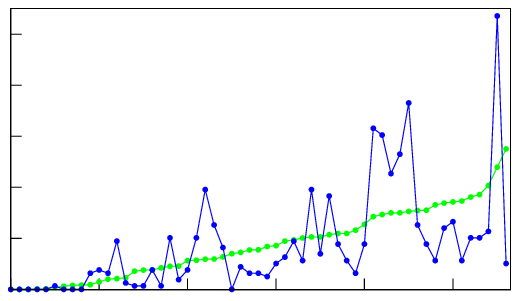}}%
		\gplfronttext
		\end{picture}%
		\endgroup
		\vspace{20pt}
		\label{fig:transform:opt}
	\end{subfigure}%
	%
	\caption{\textbf{Molecule Transformation via {\pipeline} optimization in ZINC Chemical Space.} 
		\textbf{a,} Test molecules representativeness.
	    \textbf{b,} Optimized molecules representativeness.}
	\label{fig:transform}
\end{figure*}
%
\begin{figure*}
	\centering
	\begin{small}
		\fbox{\begin{minipage}{0.6\linewidth}
				\centering
				\scriptsize{
				Retained scaffolds after \pipeline optimization are highlighted in \colorbox{SkyBlue!60}{sky blue}. \\
				Numbers associated with \molx and \moly are the corresponding \plogp values. \par
			}
			\end{minipage}
		}
	\end{small}
	\\
	\begin{minipage}{0.9\linewidth}
	\begin{subfigure}[b]{.32\textwidth}
		\vspace{10pt}
		\caption{}
		\begin{subfigure}[b]{.43\textwidth}
			\centering
			\captionsetup{justification=centering}
			\includegraphics[width=0.8\textwidth]{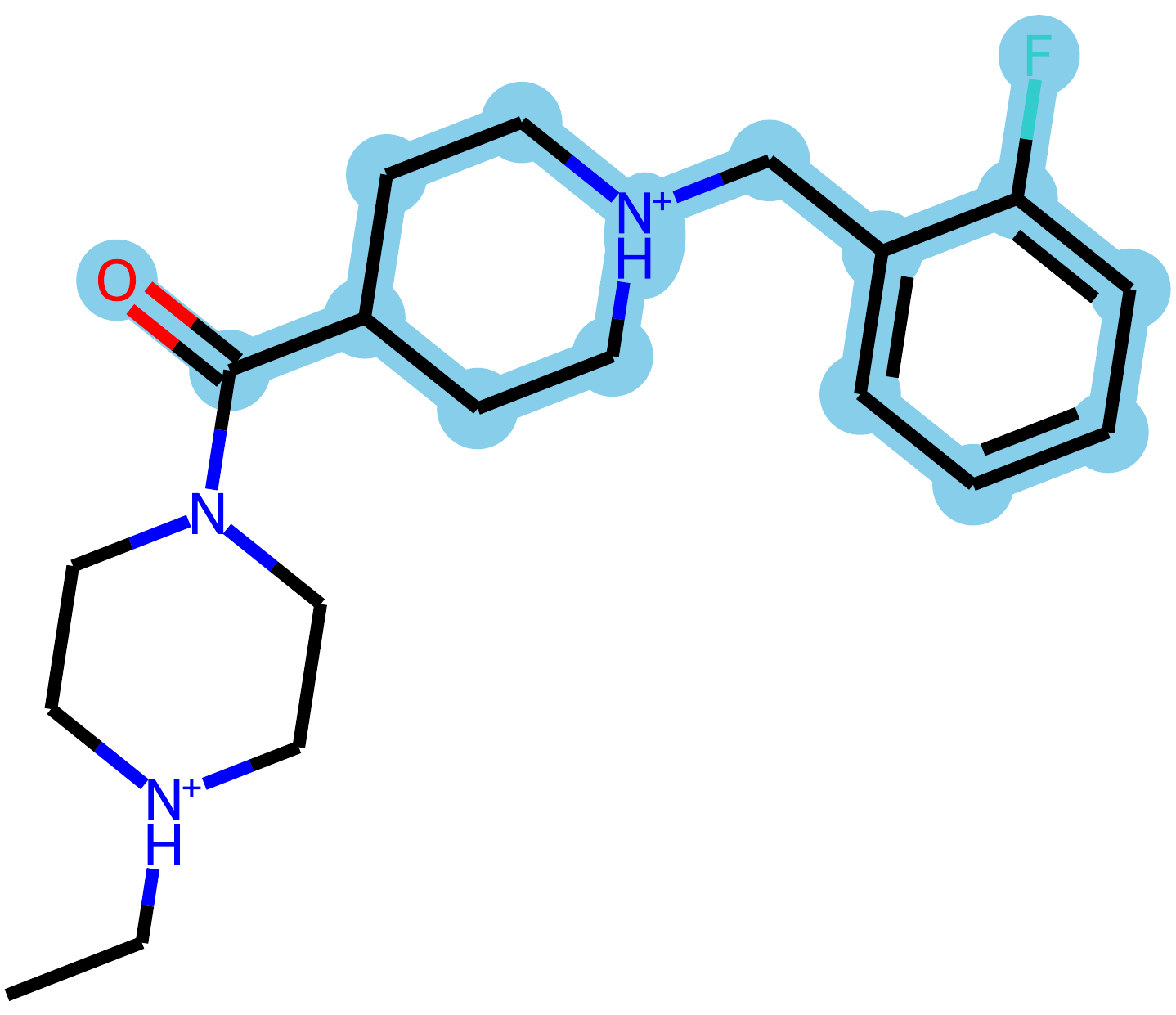}
			\caption*{$\mol_x$ -3.61}
			\label{fig:original}
		\end{subfigure}
		\begin{subfigure}[b]{.53\textwidth}
			\centering
			\captionsetup{justification=centering}
			\includegraphics[width=0.8\textwidth]{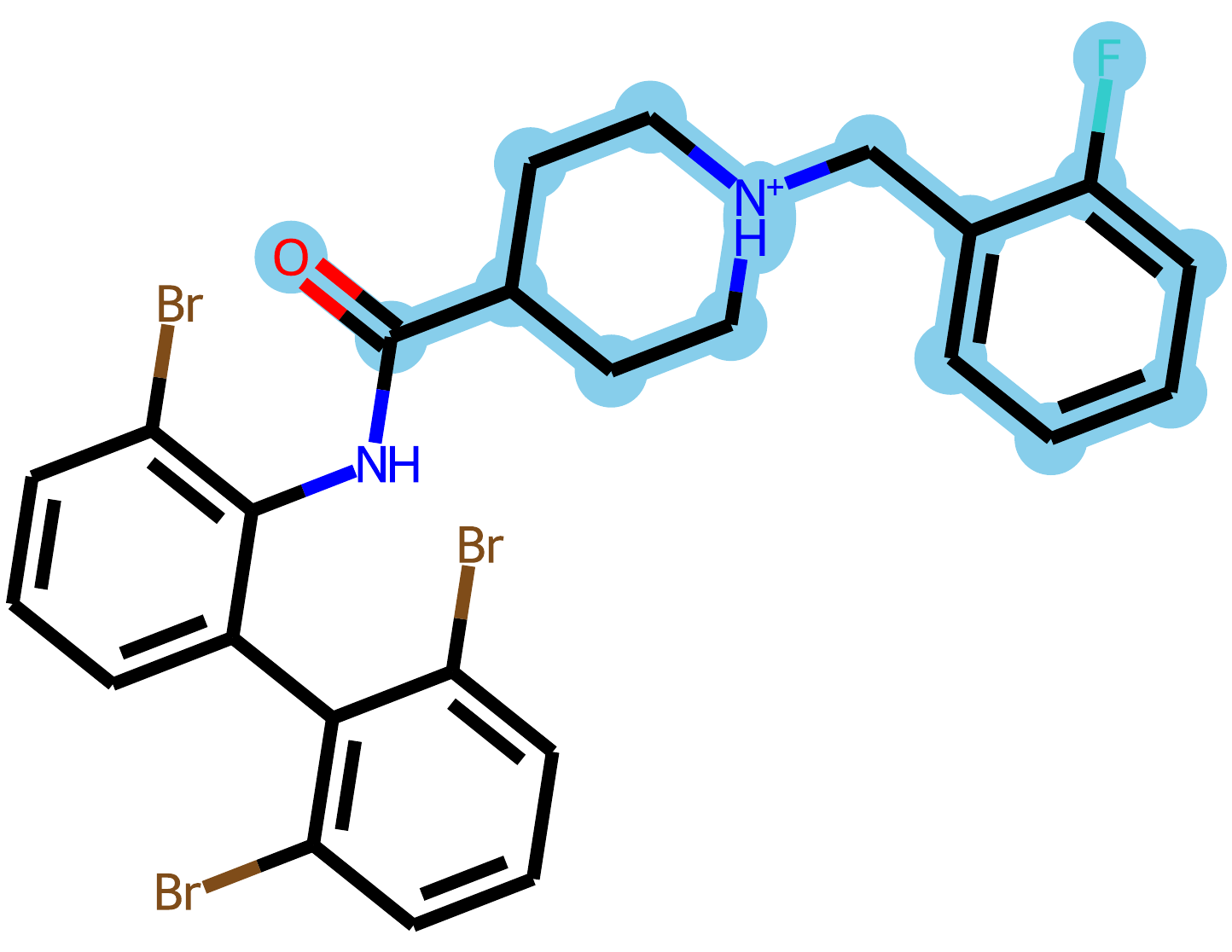}
			\caption*{$\mol_y$ 2.36}
			\label{fig:step1}
		\end{subfigure}%
	\end{subfigure}
	~
	\begin{subfigure}[b]{.32\textwidth}
		\caption{}
		\vspace{8pt}
		\begin{subfigure}[b]{.40\textwidth}
			\centering
			\captionsetup{justification=centering}
			\includegraphics[width=0.9\textwidth]{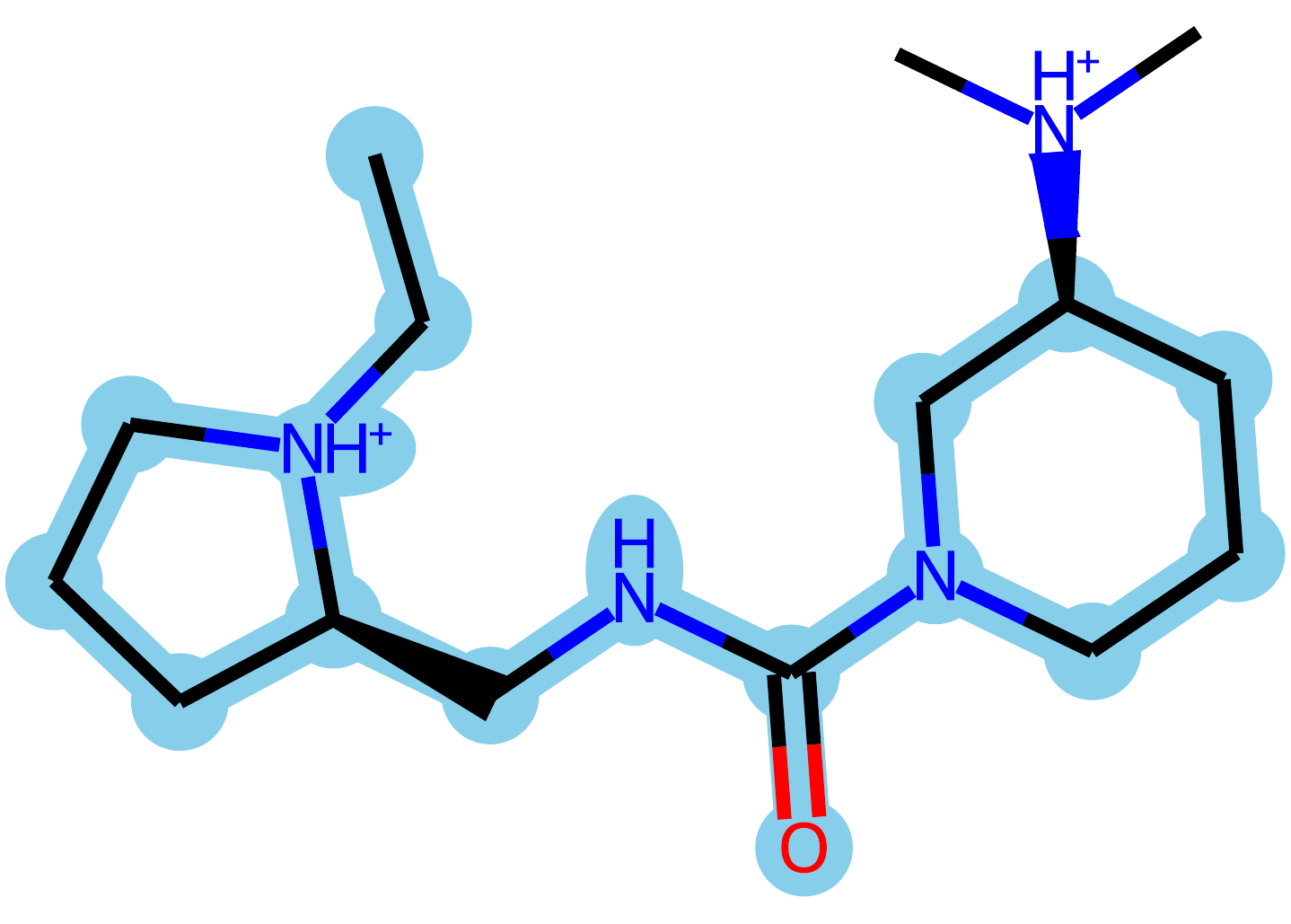}
			\caption*{$\mol_x$ -5.79}
			\label{fig:step2}
		\end{subfigure}
		\begin{subfigure}[b]{.55\textwidth}
			\centering
			\captionsetup{justification=centering}
			\includegraphics[width=0.85\textwidth]{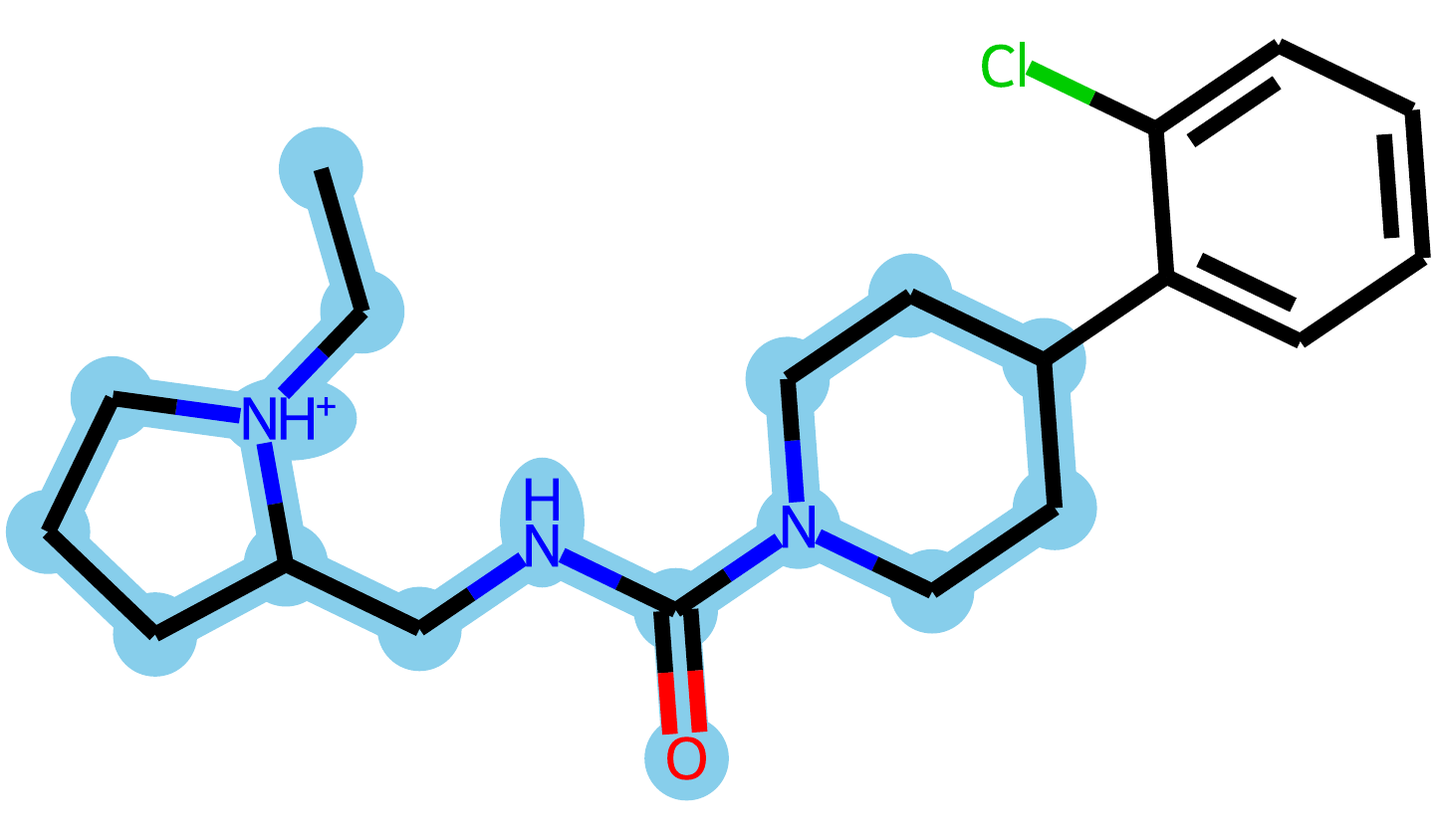}
			\caption*{$\mol_y$ -1.11}
			\label{fig:step2}
		\end{subfigure}
	\end{subfigure}
	~
	\begin{subfigure}[b]{.32\textwidth}
		\caption{}
		\vspace{5pt}
		\begin{subfigure}[b]{.48\textwidth}
			\centering
			\captionsetup{justification=centering}
			\includegraphics[width=0.9\textwidth]{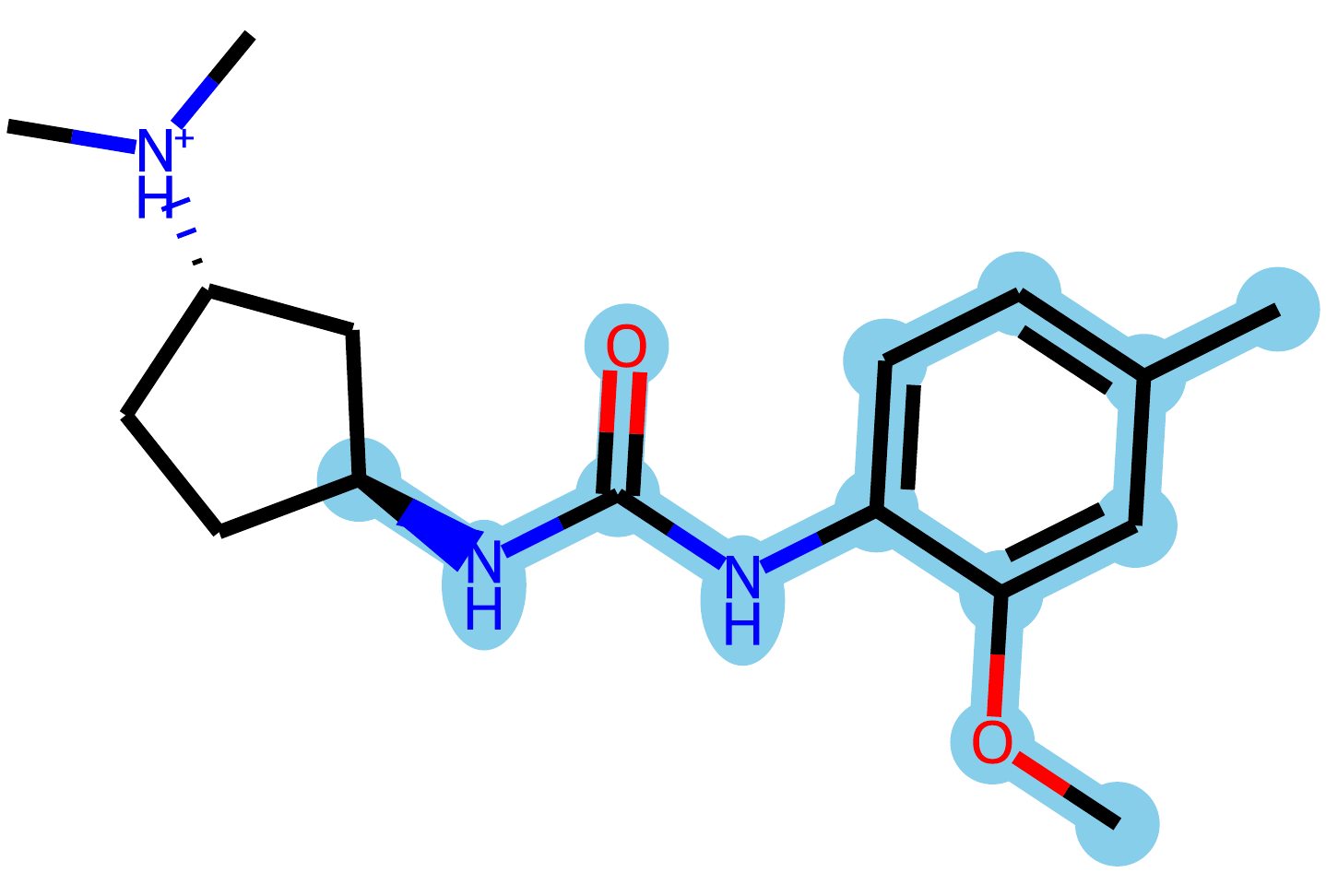}
			\caption*{$\mol_x$ -1.52}
			\label{fig:step3}
		\end{subfigure}
		\begin{subfigure}[b]{.48\textwidth}
			\centering
			\captionsetup{justification=centering}
			\includegraphics[width=0.9\textwidth]{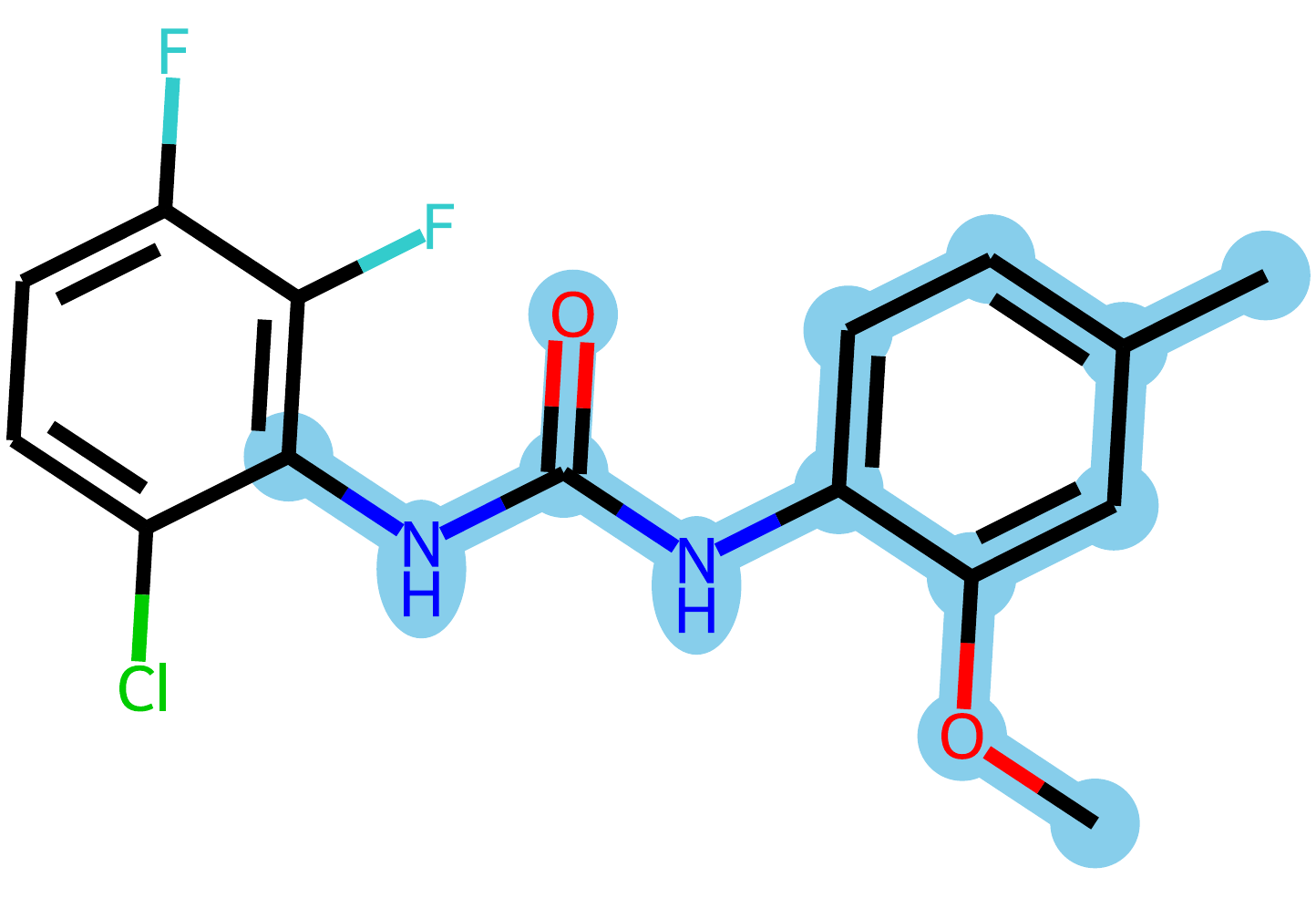}
			\caption*{$\mol_y$ 2.90}
			\label{fig:step4}
		\end{subfigure}
	\end{subfigure}
	\\
	\begin{subfigure}[b]{.32\textwidth}
		\caption{}
		\begin{subfigure}[b]{.44\textwidth}
			\centering
			\vspace{-10pt}
			\captionsetup{justification=centering}
			\includegraphics[width=0.7\textwidth]{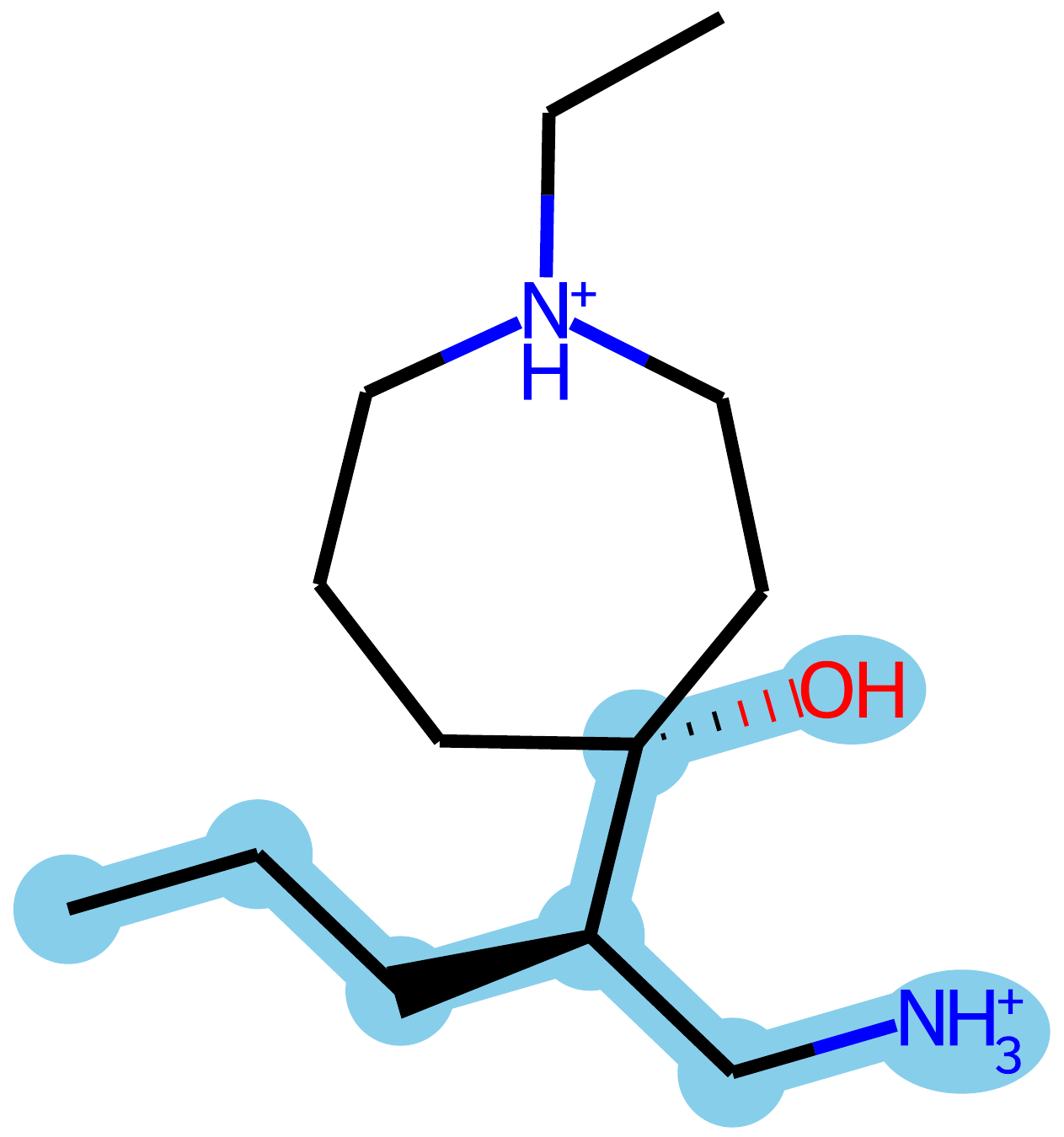}
			\caption*{$\mol_x$ -8.88}
			\label{fig:step3}
		\end{subfigure}
		\begin{subfigure}[b]{.48\textwidth}
			\centering
			\vspace{-2pt}
			\captionsetup{justification=centering}
			\includegraphics[width=0.8\textwidth]{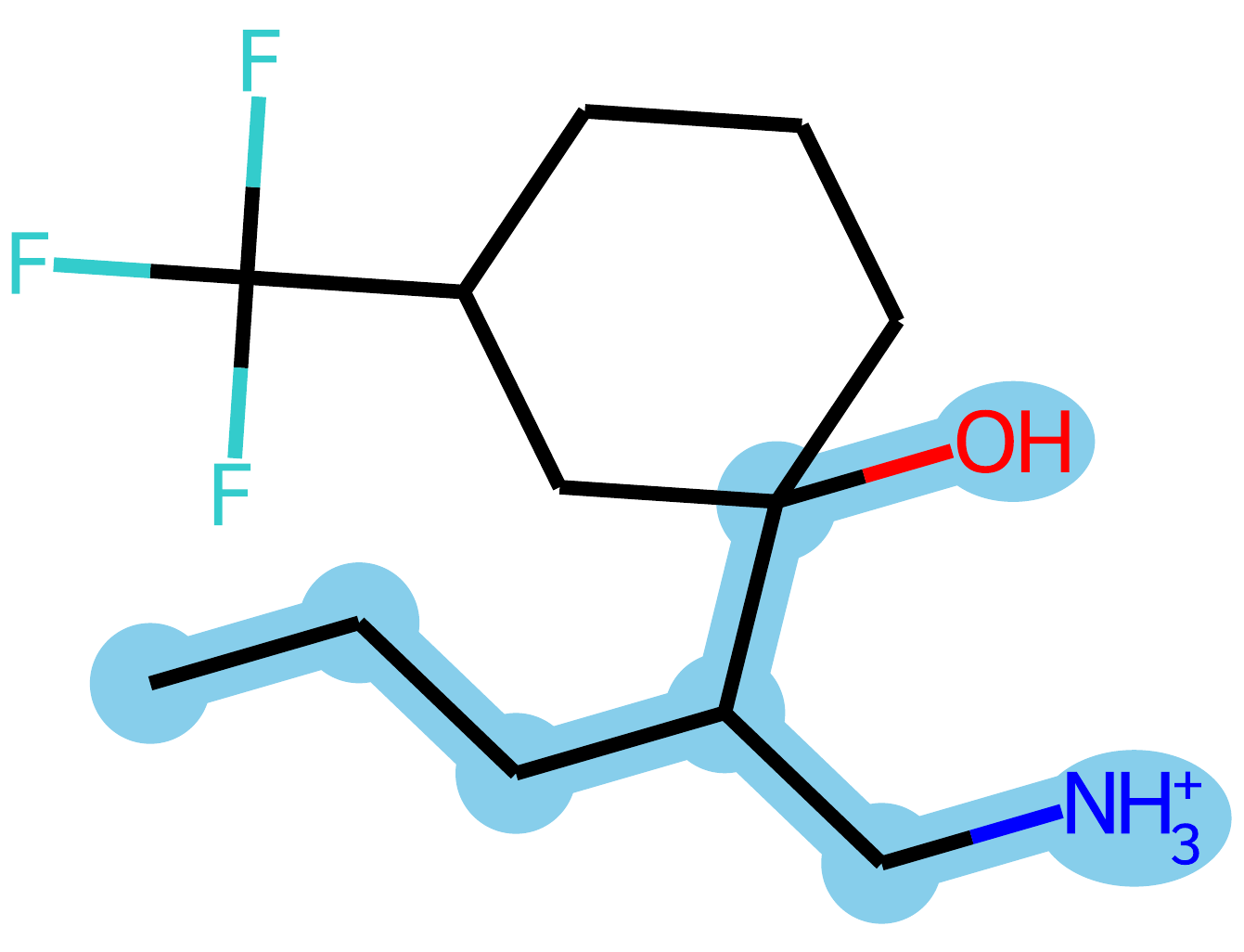}
			\vspace{-5pt}
			\caption*{$\mol_y$ -1.87}
			\label{fig:step4}
		\end{subfigure}
	\end{subfigure}
	~
	\begin{subfigure}[b]{.32\textwidth}
		\caption{}
		\begin{subfigure}[b]{.48\textwidth}
			\centering
			\captionsetup{justification=centering}
			\includegraphics[width=0.9\textwidth]{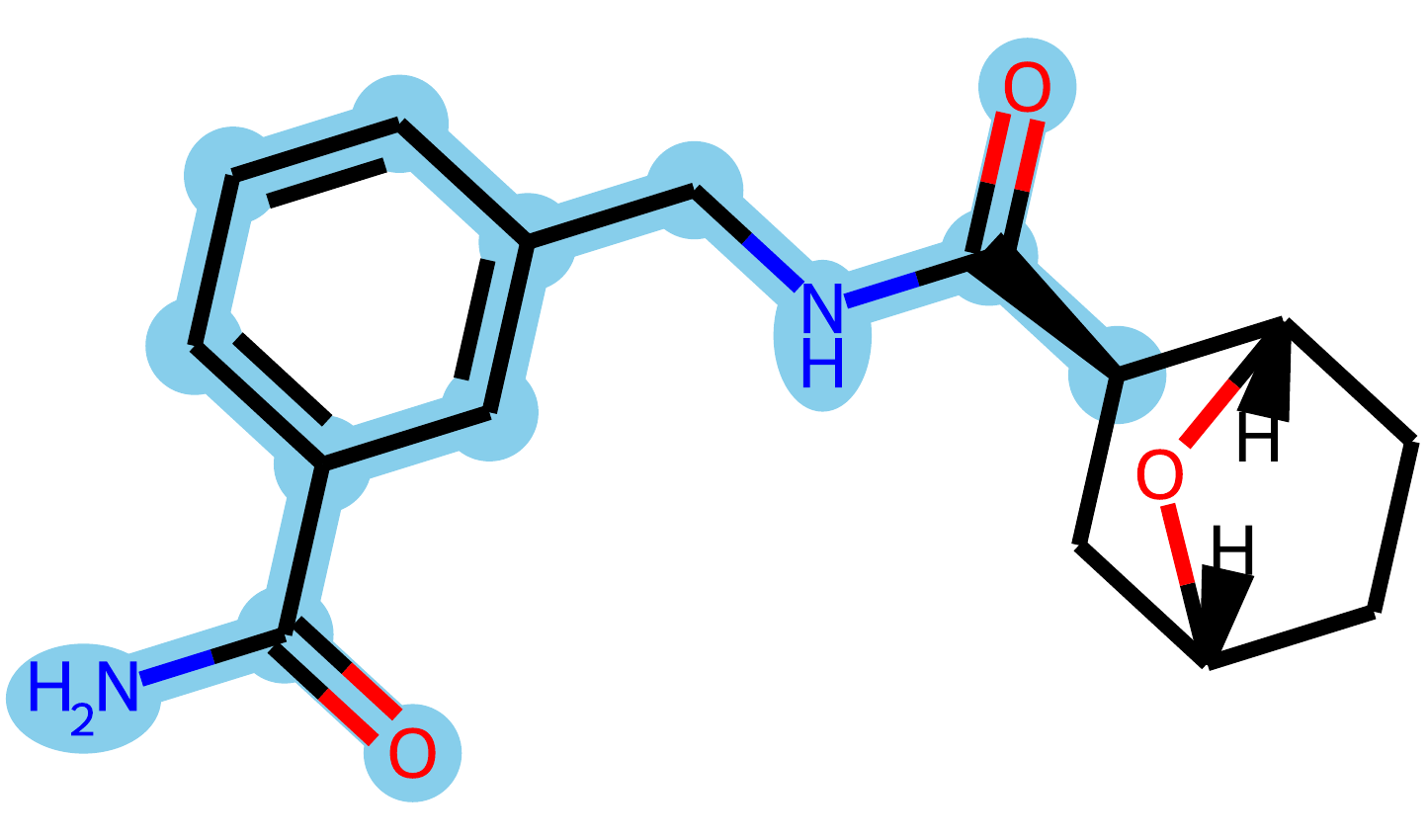}
			\caption*{$\mol_x$  -4.37}
			\label{fig:step3}
		\end{subfigure}
		\begin{subfigure}[b]{.48\textwidth}
			\centering
			\captionsetup{justification=centering}
			\includegraphics[width=0.9\textwidth]{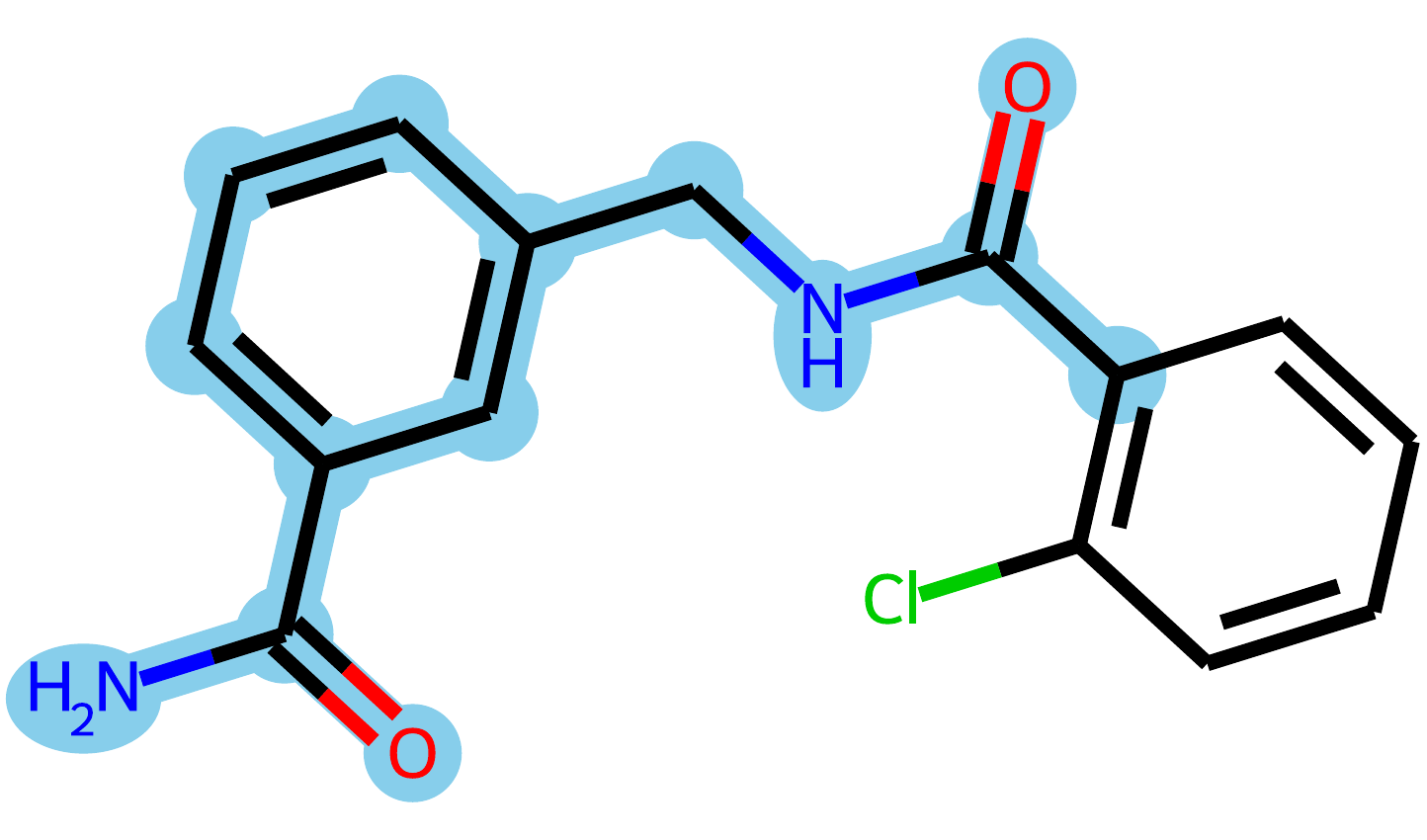}
			\caption*{$\mol_y$ 3.26}
			\label{fig:step4}
		\end{subfigure}
	\end{subfigure}
	~
	\begin{subfigure}[b]{.32\textwidth}
		\caption{}
		\begin{subfigure}[b]{.48\textwidth}
			\centering
			\vspace{2pt}
			\captionsetup{justification=centering}
			\includegraphics[width=0.9\textwidth]{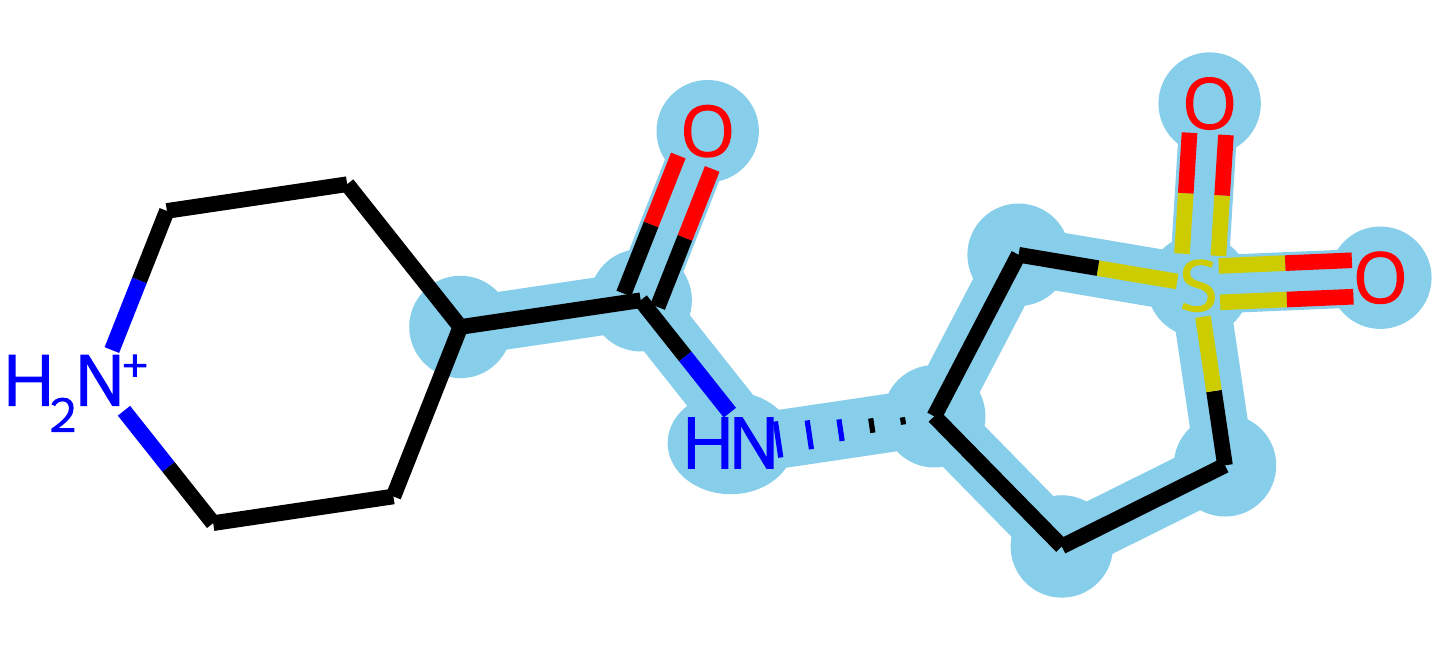}
			\caption*{$\mol_x$  -4.07}
			\label{fig:step3}
		\end{subfigure}
		\begin{subfigure}[b]{.48\textwidth}
			\centering
			\captionsetup{justification=centering}
			\includegraphics[width=0.9\textwidth]{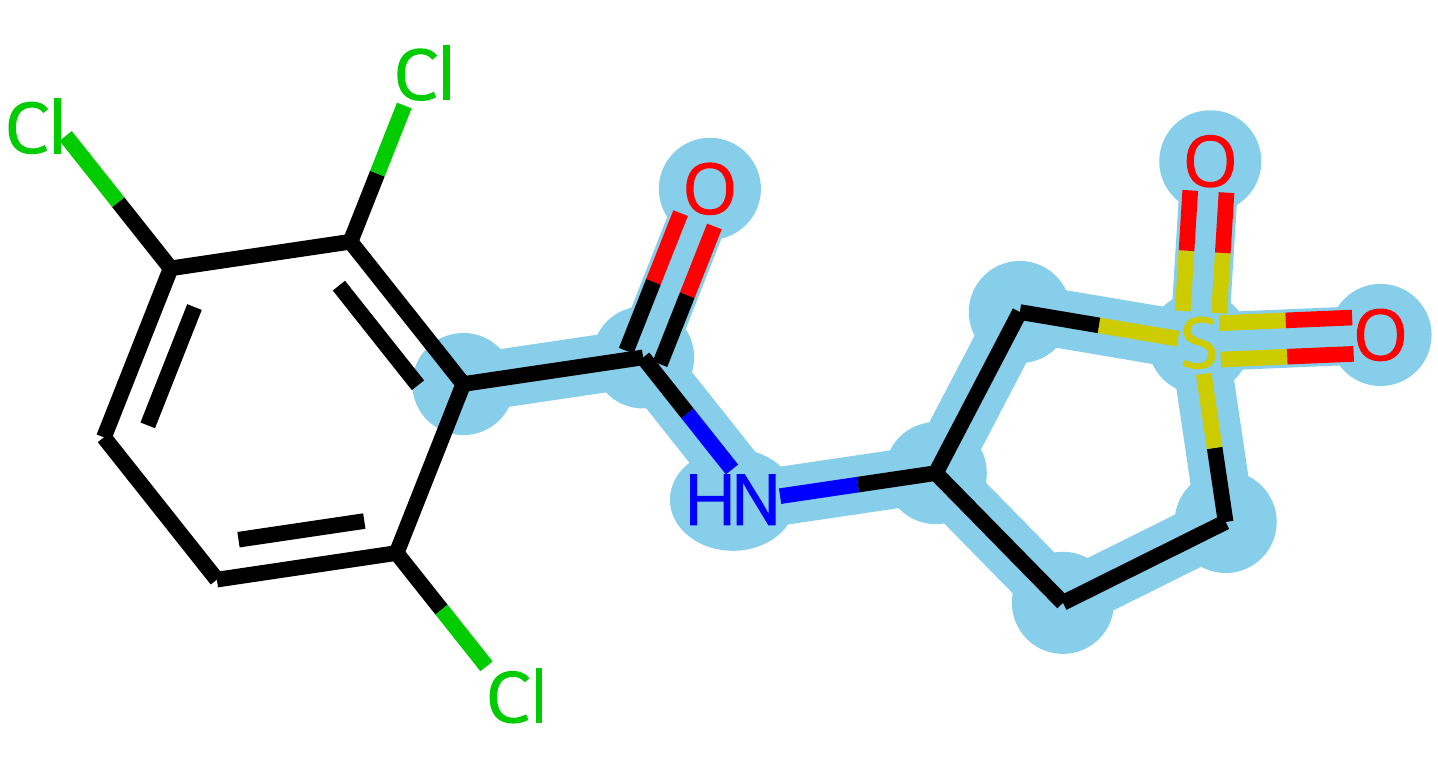}
			\caption*{$\mol_y$ 0.60}
			\label{fig:step4}
		\end{subfigure}
	\end{subfigure}
\end{minipage}
	\caption{\textbf{Examples of Test Molecules and their Optimized Molecules in {\plogp} in Different ZINC Subspaces.} 
		}
	\label{fig:test_optim_cluster}
	\vspace{20pt}
\end{figure*}

{
Fig.~{\ref{fig:norm_plogp}} shows that {\pipeline} is able to significantly improve {\plogp} (mean value changed from -2.75
to 2.25 due to the {\pipeline} optimization). 
Among the improved {\plogp} values, improvement from {\logp} contributes most as Fig.~{\ref{fig:norm_logp}} shows, where
{\logp} mean values changed from -1.40 
to 2.85. 
Even though, SA scores are also improved by {\pipeline} as in Fig.~{\ref{fig:norm_SA}}, where the mean value of SA scores 
is improved from -1.30 to -0.64. 
This indicates that while focusing on learning how to improve {\logp} 
via fragment-based modification, {\pipeline} is also able to learn and incorporate synthetic accessibility into its modification process, 
and generate new molecules that are even better synthesizable. 
Fig.~{\ref{fig:norm_SA_circle}} shows the improvement of SA scores and 
circle scores together. Note that circle scores are small if there are large rings, which are not preferable for synthesis. 
The improvement of SA and circle scores together also indicates that {\pipeline} modifies molecules into 
more synthesizable ones. }

\subsection{{Chemical Transformation via {\pipeline} Optimization}}
\label{appendix:pipeline:shift}

\begin{figure*}
	\vspace{-10pt}
	\centering
	\begin{small}
		\fbox{\begin{minipage}{0.7\linewidth}
				\centering
				\scriptsize{
				Retained scaffolds after optimization are highlighted in \colorbox{SkyBlue!60}{sky blue}. \\
				Same atoms with different formal charges after optimization are highlighted in \colorbox{Orange!70}{orange}. \\
				Numbers associated different methods are the corresponding \plogp values. \par
			}
			\end{minipage}
		}
	\end{small}
	\\
	\vspace{5pt}
	\begin{minipage}{0.9\linewidth}
		\centering
	\begin{subfigure}[b]{\textwidth}
		\caption{}
		\vspace{-25pt}
		\begin{subfigure}[b]{.12\textwidth}
			\centering
			\captionsetup{justification=centering}
			\includegraphics[width=\textwidth]{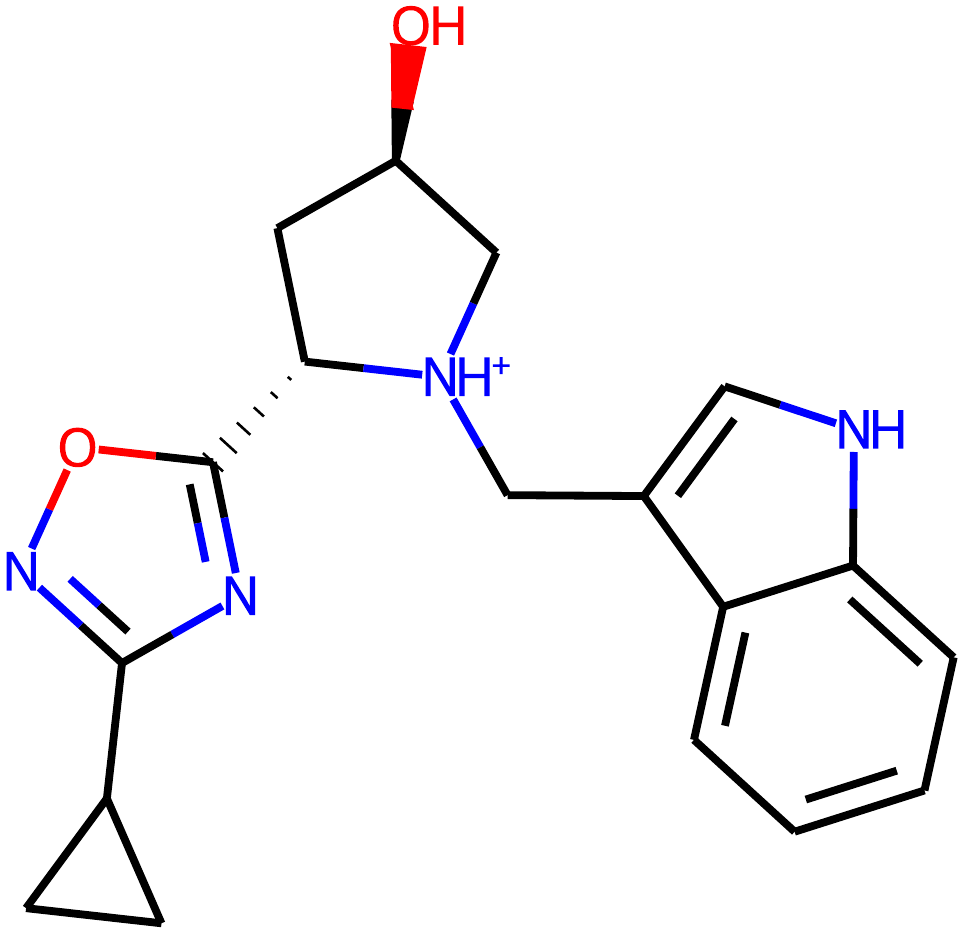}
			\caption*{\molx~-2.36}
			\label{fig:original1}
		\end{subfigure}
		\!\!\!\!\!
		\begin{subfigure}[b]{.14\textwidth}
			\centering
			\vspace{15pt}
			\captionsetup{justification=centering}
			\includegraphics[width=0.9\textwidth]{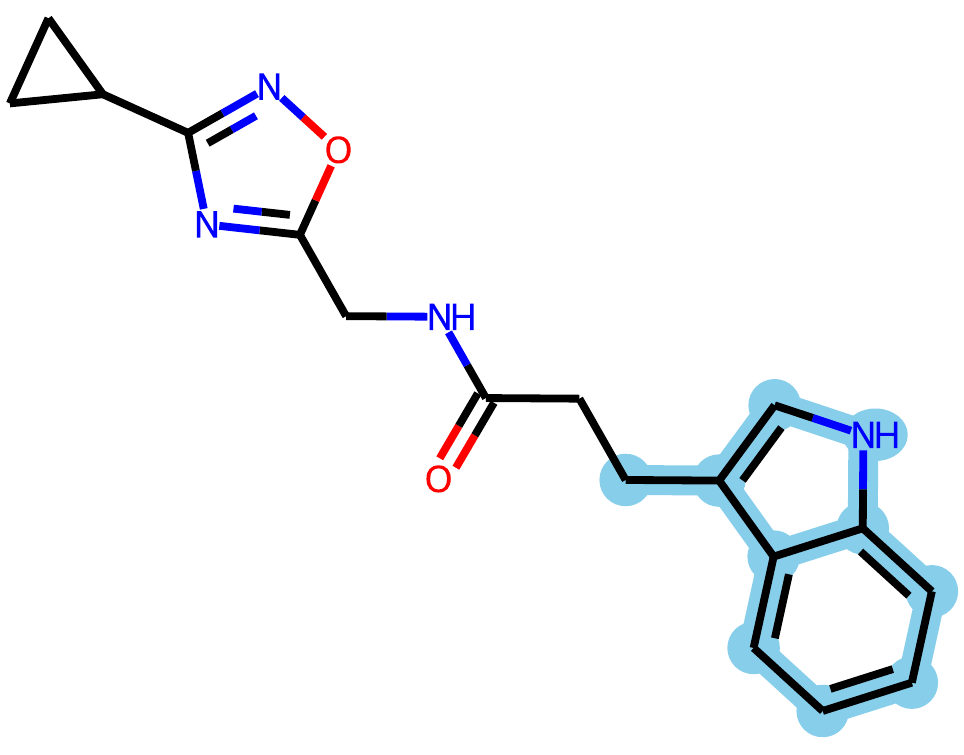}
			\caption*{\jtnn 1.26}
			\label{fig:jtnn1}
		\end{subfigure}%
		~
		\begin{subfigure}[b]{.12\textwidth}
			\centering
			\vspace{15pt}
			\captionsetup{justification=centering}
			\includegraphics[width=\textwidth]{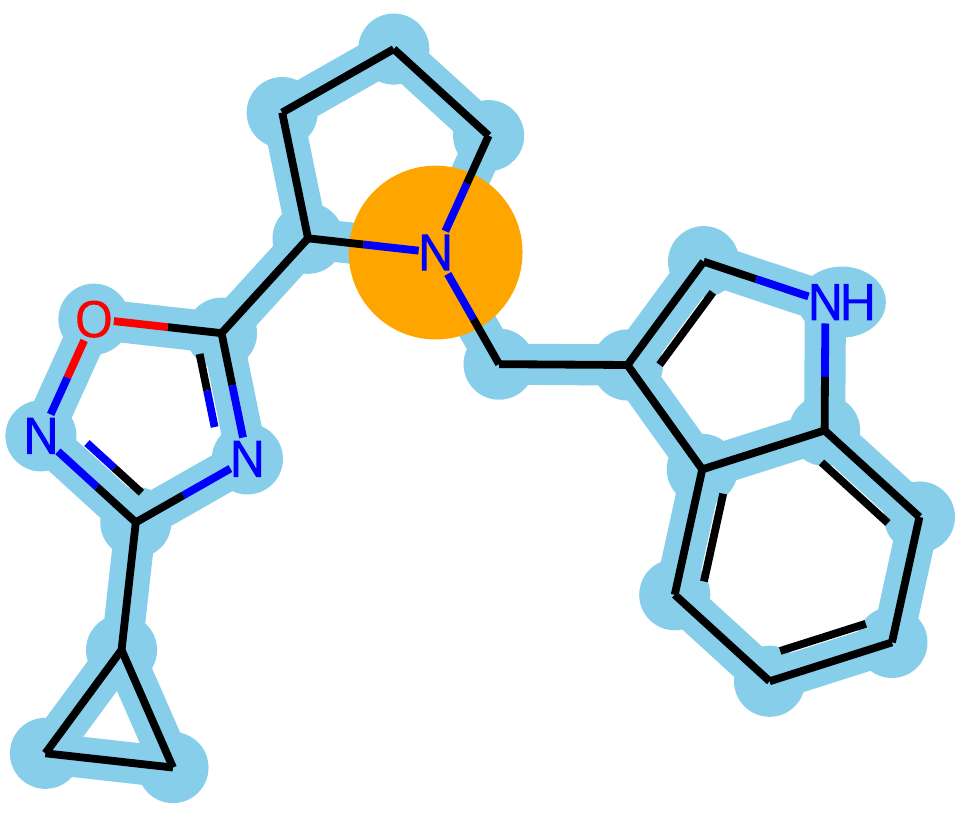}
			\caption*{\hiergtog 1.32}
			\label{fig:hierg2g1}
		\end{subfigure}
		~
		\begin{subfigure}[b]{.12\textwidth}
			\centering
			\vspace{15pt}
			\captionsetup{justification=centering}
			\includegraphics[width=\textwidth]{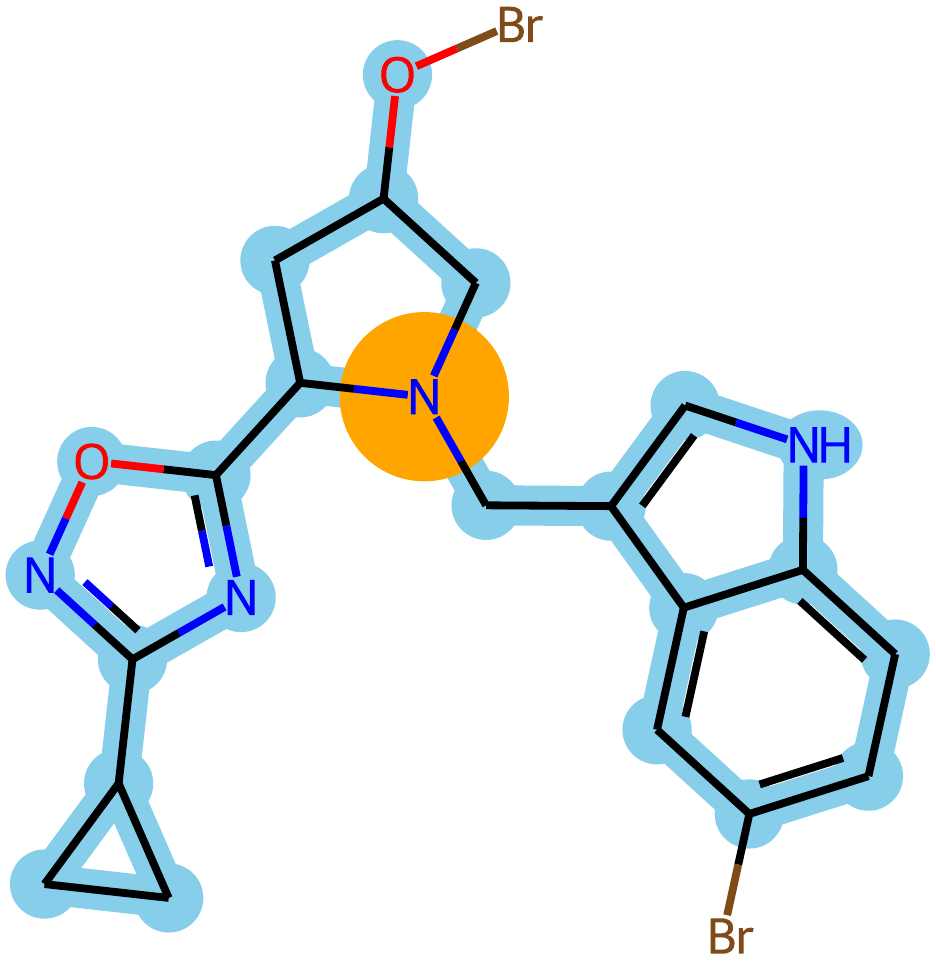}
			\vspace{-10pt}\caption*{\graphaf 0.97}
			\label{fig:graphaf1}
		\end{subfigure}
		~
		\begin{subfigure}[b]{.12\textwidth}
			\centering
			\captionsetup{justification=centering}
			\includegraphics[width=.95\textwidth]{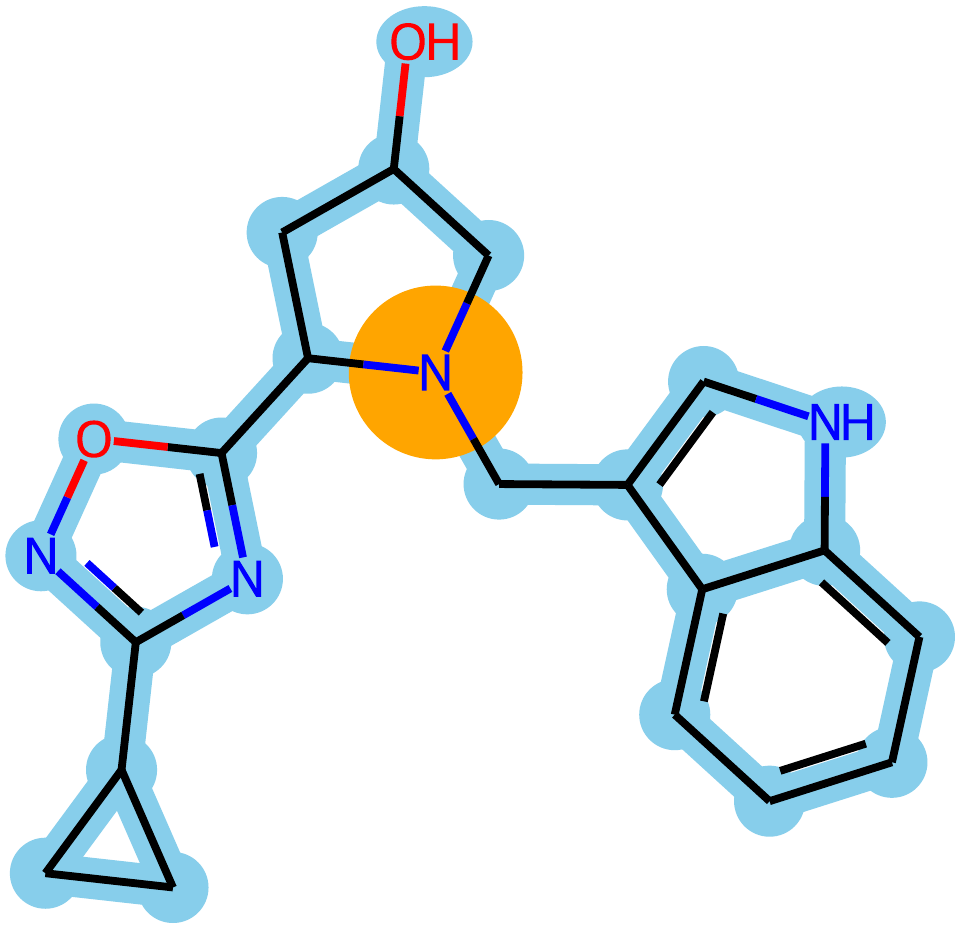}
			\caption*{\moflow 0.07}
			\label{fig:moflow1}
		\end{subfigure}
		~
		\begin{subfigure}[b]{.12\textwidth}
			\centering
			\captionsetup{justification=centering}
			\includegraphics[width=.9\textwidth]{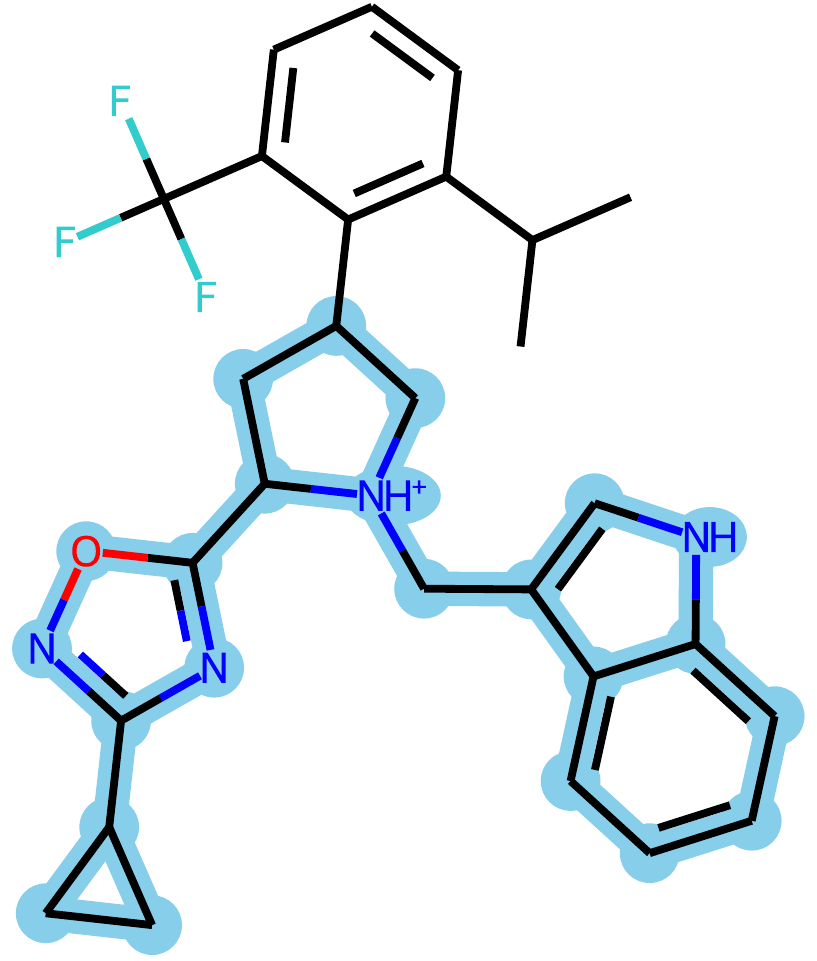}
			\caption*{\pipeline 0.82}
			\label{fig:pipeline1}
		\end{subfigure}
		~
		\begin{subfigure}[b]{.13\textwidth}
			\centering
			\captionsetup{justification=centering}
			\vspace{20pt}
			\includegraphics[width=.9\textwidth]{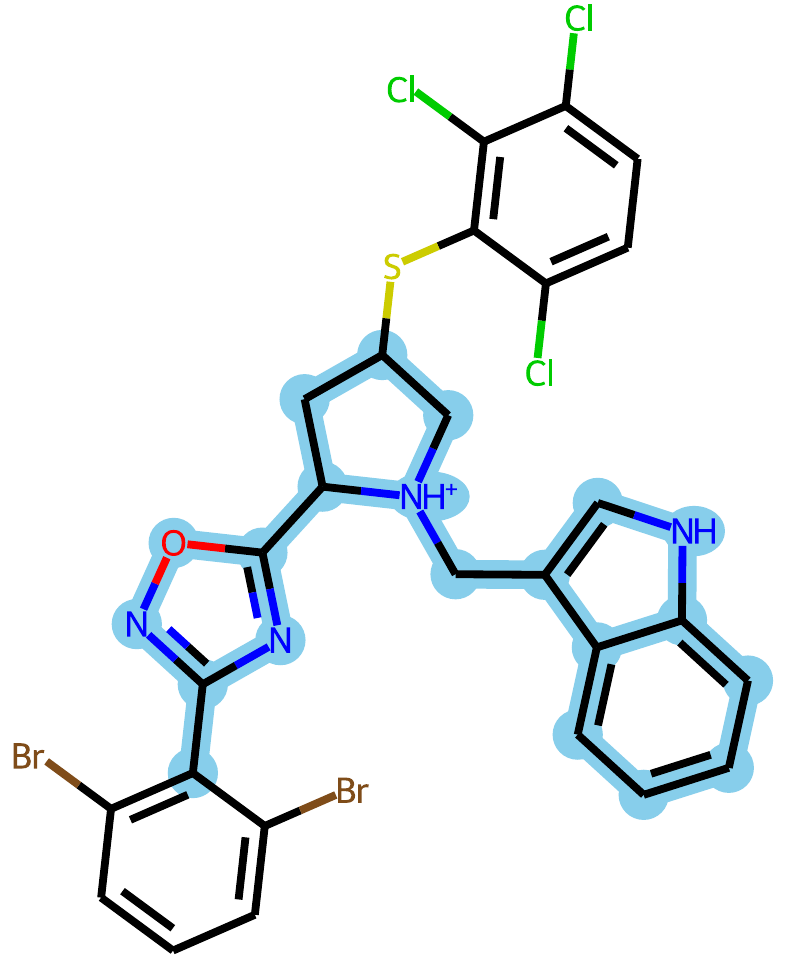}
			\caption*{\pipelineF 2.40}
			\label{fig:pipelineF1}
		\end{subfigure}
		\label{fig:retain_scaffold1}
	\end{subfigure}
	\\
	\begin{subfigure}[b]{\textwidth}
		\vspace{5pt}
		\caption{}
		\vspace{-1pt}
		\begin{subfigure}[b]{.12\textwidth}
			\centering
			\captionsetup{justification=centering}
			\includegraphics[width=\textwidth,angle=90]{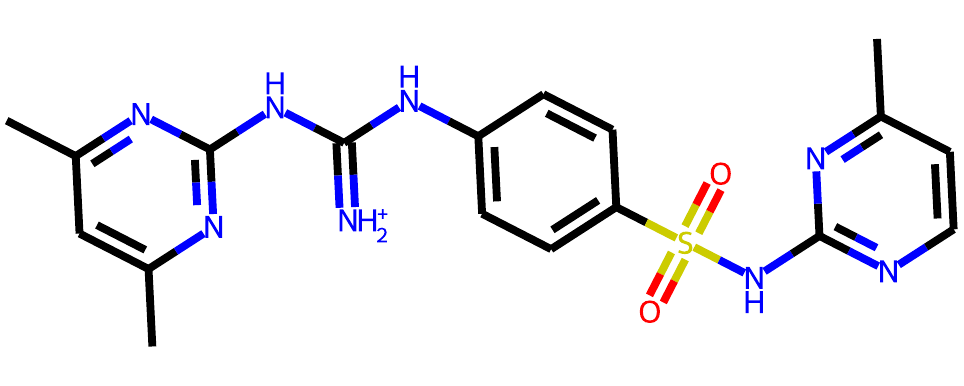}
			\caption*{\molx~-0.85}
			\label{fig:original2}
		\end{subfigure}
		\!\!\!\!\!
		\begin{subfigure}[b]{.16\textwidth}
			\centering
			\captionsetup{justification=centering}
			\includegraphics[width=\textwidth,angle=90,scale=0.8]{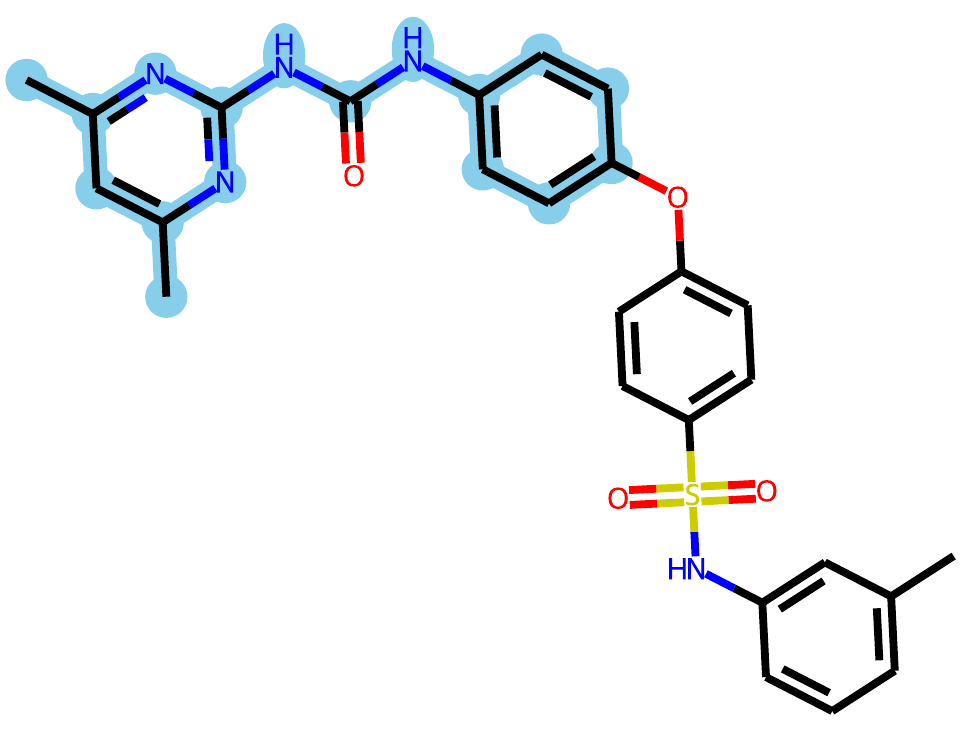}
			\caption*{\jtnn 3.37}
			\label{fig:jtnn2}
		\end{subfigure}%
		\begin{subfigure}[b]{.12\textwidth}
			\centering
			\captionsetup{justification=centering}
			\includegraphics[width=\textwidth, angle=90]{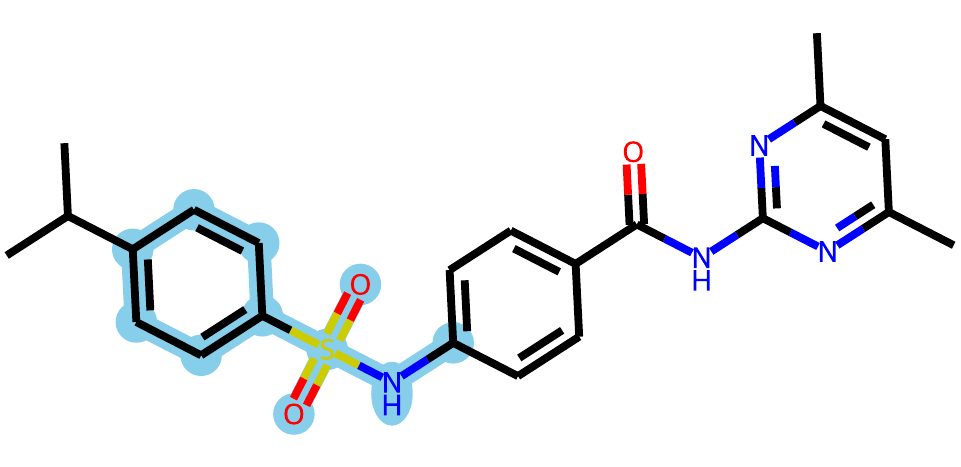}
			\caption*{\hiergtog 2.61}
			\label{fig:hierg2g2}
		\end{subfigure}
		\begin{subfigure}[b]{.12\textwidth}
			\centering
			\captionsetup{justification=centering}
			\includegraphics[width=\textwidth,angle=90]{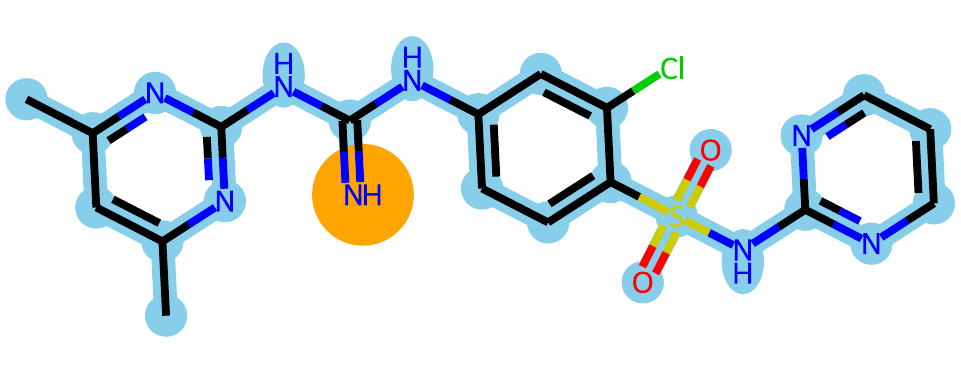}
			\caption*{\graphaf 0.77}
			\label{fig:graphaf2}
		\end{subfigure}
		\begin{subfigure}[b]{.12\textwidth}
			\centering
			\captionsetup{justification=centering}
			\includegraphics[width=\textwidth,angle=90]{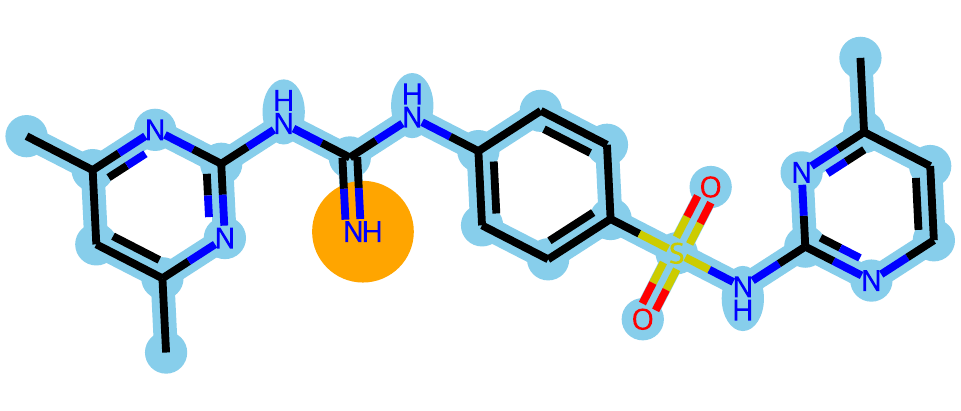}
			\caption*{\moflow 0.63}
			\label{fig:moflow2}
		\end{subfigure}
		\begin{subfigure}[b]{.14\textwidth}
			\centering
			\captionsetup{justification=centering}
			\includegraphics[width=\textwidth,angle=90]{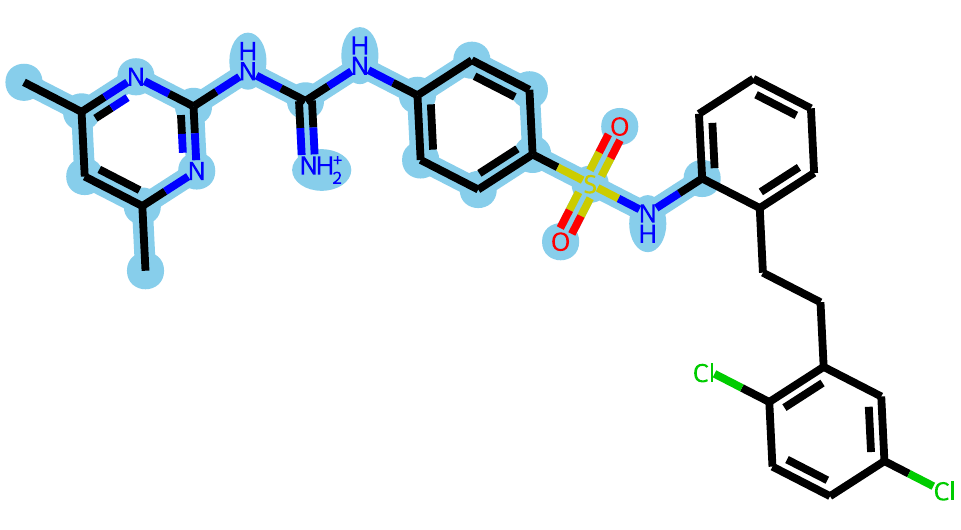}
			\caption*{\pipeline 1.93}
			\label{fig:pipeline2}
		\end{subfigure}
		\begin{subfigure}[b]{.16\textwidth}
			\centering
			\captionsetup{justification=centering}
			\includegraphics[width=\textwidth,angle=90,scale=0.8]{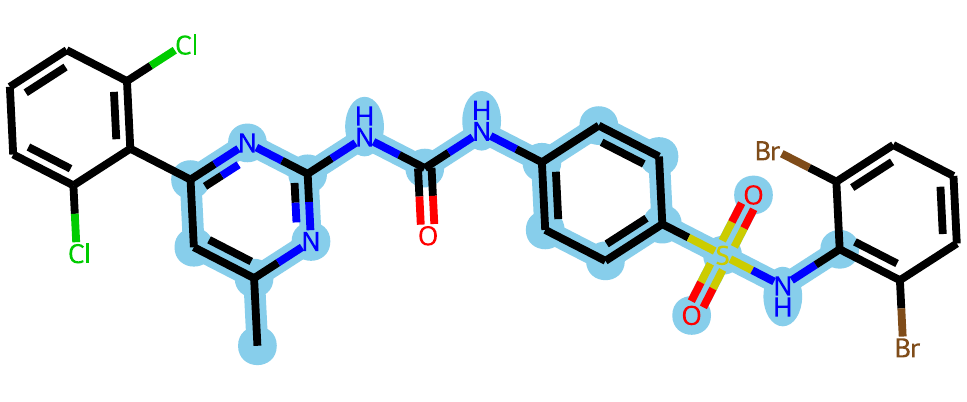}
			\caption*{\pipelineF 4.37}
			\label{fig:pipelineF2}
		\end{subfigure}
		\label{fig:retain_scaffold2}
	\end{subfigure}
\end{minipage}
	\caption{\textbf{Examples of Optimized Molecules in {\plogp} by Different Methods}
			 }
	\label{fig:retain_scaffolds}
\end{figure*}

{An important aspect of chemical optimization is to search for and explore a novel chemical subspace where
molecules have better properties~{\cite{lipinski2004}}. 
To analyze whether the optimized molecules given by {\pipeline} reside in a novel chemical subspace compared to 
the subspace of those before optimization, we conducted the following analysis. 
We clustered the following five groups of molecules all together: 
(1) the molecules in training pairs that have bad properties, denoted as $\{\molx\}_{\text{trn}}$; 
(2) the molecules in training pairs that have good properties, denoted as $\{\moly\}_{\text{trn}}$;   
(3) the molecules in the benchmark test dataset, denoted as $\{\molx\}_{\text{tst}}$ ($|\{\molx\}_{\text{tst}}|=800$); 
(4) the optimized test molecules by {\pipeline} with similarity constraint $\delta$=0.4, denoted as $\{\moly\}_{\text{tst}}$; and 
(5) the rest, all the ZINC dataset molecules that are not in the above four groups. 
We did the clustering in a same way as in Section~{\ref{appendix:data:coverage}} using CLUTO. 
The clustering results represent how the ZINC chemical space looks, and 
how molecules are distributed in the ZINC chemical space. }

{%
Fig.~{\ref{fig:transform}} presents the results from 56 clusters that are sorted in the same way 
as in Section~{\ref{appendix:data:coverage}}. 
Fig.~{\ref{fig:transform:test}} and Fig.~{\ref{fig:transform:opt}} show that the test molecules and their optimized ones 
have roughly similar distributions over ZINC clusters (e.g., more distributed in cluster 40 - 50). 
This indicates that {\pipeline} optimization is not biased at  a certain chemical subspace. 
On the other hand, the distributions still have 
some notable difference, for example, before optimization, about 8\% of test molecules were in cluster 23; after optimization, 
only 4\% of the optimized molecules are in cluster 23. 
Actually, among the 800 test molecules, 24.38\% of the molecules changed to different clusters after {\pipeline} optimization. 
Compared to test molecules, among the training data, 
only about 9.20\% of $\{\molx\}_{\text{trn}}$ have their paired $\{\moly\}_{\text{trn}}$ in a different cluster. 
%
%
%
Fig.~{\ref{fig:test_optim_cluster}} shows the examples of test molecules for which their optimized molecules changed to different 
clusters.
The analysis results indicate that {\pipeline} is able to learn from training data (training chemical subspace), and also 
explore and transform into novel chemical subspaces. 
Note that {\pipeline} always constrains that the optimized molecules are at least $\delta$-similar to those before optimization 
(in this analysis, $\delta$=0.4). Under this constraint, the fact that {\pipeline} is able to find novel chemical subspaces for about 24\% 
of test molecules indicates its strong exploration capability over the entire chemical space for molecule optimization. 
}


\subsection{{Comparison on Retaining Molecule Scaffolds}}
\label{appendix:pipeline:scaffolds}
{
Fig.~{\ref{fig:retain_scaffolds}} presents two examples of molecules optimized by the four baseline methods, {\pipeline} and {\pipelineF}, 
with similarity constraint $\delta$=0.4. 
In these two examples, {\jtnn} tends to insert a new structure within the middle of the molecules. 
{\hiergtog} could have either very minor (in Fig.~{\ref{fig:retain_scaffold1}}) or very dramatic (Fig.~{\ref{fig:retain_scaffold2}}) change
to the major molecular scaffolds. 
Both {\graphaf} and {\moflow} tend to have very small modifications and therefore {\plogp} improvement from these two methods 
is also small. 
{\pipeline} and {\pipelineF} tend to change the periphery of the molecules but can guarantee to retain the major scaffolds 
of the molecules, with substantial {\plogp} improvement.}
\subsection{Model Complexity Comparison}
\label{appendix:additional:complexity}

\begin{table*}
	\caption{Model Complexity for {\plogp} Optimization}
	\label{tbl:mol_comp}
	
	\centering
	\begin{threeparttable}
		\scriptsize{
		\begin{tabular}{
				@{\hspace{10pt}}l@{\hspace{10pt}}
				@{\hspace{10pt}}r@{\hspace{10pt}}
				@{\hspace{10pt}}r@{\hspace{10pt}}        
			}
			\toprule
			model & number of parameters \\
			\midrule
			\jtvae    & 4.53M \\
			\gcpn     & 6.00M \\
			\jtnn ($\delta$=0.4)     & 3.74M \\
			\jtnn ($\delta$=0.6) & 3.15M\\
			\hiergtog & 6.80M \\
			\graphaf  & 1.86M   \\
			\moflow   & 130.35M \\
			\molmod ($\delta$=0.0)  & {0.56M} \\
			\molmod ($\delta$=0.2) & {0.54M} \\
			\molmod ($\delta$=0.4) & {1.89M} \\
			\molmod ($\delta$=0.6) & {0.51M} \\
			\bottomrule
		\end{tabular}
	}
	\end{threeparttable}
	\vspace{-5pt}
\end{table*}

Table~\ref{tbl:mol_comp} presents the number of parameters of baselines and our models.
As shown in this table, the optimal \molmod has 0.56M
parameters with \mbox{$\delta\!=\!0.0$}, 0.54M
parameters with \mbox{$\delta\!=\!0.2$}, 1.89M parameters with \mbox{$\delta\!=\!0.4$} 
and 0.51M parameters with \mbox{$\delta\!=\!0.6$}, 
which are far less than those in the best baselines. 
For example, \jtnn has 3.74M parameters with \mbox{$\delta\!=\!0.4$} and  3.15M with \mbox{$\delta\!=\!0.6$}, and \hiergtog has 6.8M parameters, 
that is, \molmod uses at least 40\% fewer parameters and 26\% less training data 
but outperforms or achieves very comparable results as these state-of-the-art baselines. 
%

\subsection{Parameters for Reproducibility {of  {\plogp} Optimization}}
\label{appendix:logp:param}

\begin{table*}
	\centering
	\caption{{Hyper-Parameter Space for {\plogp} Optimization}}
	\label{tbl:hyper}
	
	\begin{threeparttable}
		\scriptsize{
			\begin{tabular}{
					@{\hspace{2pt}}l@{\hspace{5pt}} 
					@{\hspace{2pt}}l@{\hspace{2pt}}         
				}
				\toprule
				Hyper-parameters &  Space\\
				\midrule
				hidden layer dimension         & \{32, 64, 128, 256\} \\
				atom/node embedding dimension &  \{32, 64, 128, 256\} \\
				$\latent^{\add}$/$\latent^{\delete}$ dimension        & \{8, 16, 32\} \\
				\# iterations of \GMPN  & \{3, 4, 5, 6\} \\
				\# iterations of \TMPN  & \{2, 3, 4\} \\
				\# sampling             & 20          \\
				\bottomrule
			\end{tabular}
			%
			%
		}
	\end{threeparttable}
	
\end{table*}

We implemented our models using \mbox{Python-3.6.9}, \mbox{Pytorch-1.3.1}, \mbox{RDKit-2019.03.4} and \mbox{NetworkX-2.3}.
We trained the models using a Tesla P100 GPU and a CPU with 16GB memory on Red Hat Enterprise 7.7.
%
%
We tuned the hyper-parameters of our models with the grid-search algorithm in the parameter space presented in Table~\ref{tbl:hyper}.
We determined the optimal hyper-parameters according to their corresponding \plogp property improvement over the 
validation molecules. 
To optimize a molecule, we randomly sampled $K$=20 latent vectors in each \molmod iteration.


%
For $\delta$=$0.0$, the optimal dimension of all the hidden layers is {128}, and the 
dimension of latent embedding \latent is 16 (i.e., 8 for $\latent^{-}$ and $\latent^{+}$, respectively), in \molmod.
The optimal iterations of graph message passing \GMPN and tree message passing \TMPN {are 6 and 4, respectively.} 
For $\delta$=$0.2$, the optimal dimension of all the hidden layers is 128, and the 
dimension of latent embedding \latent is {32}, and the optimal iterations of \GMPN and \TMPN are 
{5} and {4}, respectively.
{
	For $\delta$=$0.4$, the optimal dimension of all the hidden layers is 256, and the 
	dimension of latent embedding {\latent} is 32, and the optimal iterations of {\GMPN} and {\TMPN} are 6 and 3, respectively.
	For $\delta$=$0.6$, the optimal dimension of all the hidden layers is 128, the 
	dimension of latent embedding {\latent} is 32, and the optimal iterations of {\GMPN} and {\TMPN} are 4 and 3, respectively.}	
%
%


%
We optimized the models with learning rate 0.001 and batch size 32. 
During the training period, we did not use regularization and dropout 
and used default random number seeds in Pytorch to sample the noise variables employed 
in the reparameterization trick of VAE.
The best performance was typically achieved within {7} epochs of training. 
%
We set the KL regularization weight $\beta$ in the loss function (Equation~\ref{eqn:opt_prob} in the main manuscript) 
as 0.1 in the first epoch, and increased its value by 0.05 every 500 batches until 0.5.

\section{Fragment and Molecule Size Analysis {in Training Data for {\plogp} Optimization}}
\label{appendix:opt}


Among the training molecules for \plogp optimization, the top-5 most popular fragments that have been removed from \molx are: 
O[C:1] (6.43\%), 
N\#[C:1] (4.56\%), 
[O-][C:1] (3.23\%), 
[NH3+][C:1] (2.44\%), 
N[C:1]=O (2.17\%); 
the top-5 most popular fragments to be attached into \moly are:  
CCSc1cccc[c:1]1 (13.92\%), 
Clc1ccc([C:1])cc1 (12.12\%), 
Clc1c[c:1]ccc1 (6.24\%), 
Clc1cccc[c:1]1 (4.67\%), 
c1ccc2sc([C:1])nc2c1 (4.00\%). 
These removal and attaching fragments are visualized in Fig.~\ref{fig:example_logp}a
and Fig.~\ref{fig:example_logp}b in the main manuscript. 

Overall, the removal fragments in training data are on average of 2.85 atoms and the new attached fragments 
are of 7.55 atoms. 
In addition, 39.48\% \molx molecules do not have fragments removed and only have new attached fragments, 
while only 1.78\% \molx molecules do not have new fragments attached and only have fragments removed. 
%
%
This shows that in training data, the optimization of \plogp is typically done via removing small fragments and then 
%
attaching larger fragments. This is also reflected in Table~\ref{tbl:atomsize} (``\#$\atom_x$", ``\#$\atom_y$")
that out of each \molmod iteration, the optimized molecules become larger. 
We observed the similar trend from \jtvae and \jtnn that their optimized molecules are also larger than those 
before optimization. 
%
In the benchmark data, larger molecules 
typically have better \plogp values (e.g., the correlation between molecule size and \plogp values is 0.42). 

%
%

%
%
%
%
%



\section{{Experimental Results on {\drd} and {\qed} Optimization}}
\label{appendix:drd2_qed}
%

{In addition to improving {\plogp}, another two popular benchmarking tasks for molecule optimization 
include improving molecule binding affinities against the dopamine D2 receptor (\drd), 
and improving the drug-likeness estimated by quantitative measures ({\qed})~{\cite{bickerton2012}}.
Specifically, given a molecule that doesn't bind well to the {\drd} receptor (e.g., with low binding affinities), 
the objective of optimizing {\drd} property is to modify the molecule into another one that will better bind to {\drd}. 
%
%
How well the molecules bind to {\drd} is assessed by a support vector machine classifier developed 
by Olivecrona {\etal}~{\cite{olivecrona2017}}, which predicts a {\drd} score to measure the binding.
In the {\qed} task, given a molecule that is not much drug-like, the objective of optimizing 
{\qed} property is to modify this molecule into a more ``drug-like" molecule.
The drug-likeness of molecules is quantified by comparing them with approved drugs on eight widely used molecular properties
such as the number of aromatic rings and molecular polar surface area~{\cite{bickerton2012}}.
Note that for these two tasks, {\molmod} does not restrict the size of the optimized molecules. }


%
\subsection{{Training Data Generation for {\drd} and {\qed} Optimization}}
\label{appendix:drd2_qed_data}
%

\begin{table*}
	\centering
	\caption{{Data Statistics for {\qed} and {\drd} Optimization}}
	\label{tbl:drd_qed_stats}
	
	\begin{threeparttable}
		\scriptsize{
			\begin{tabular}{
					@{\hspace{2pt}}p{0.4\linewidth}@{\hspace{5pt}}
					@{\hspace{2pt}}r@{\hspace{2pt}}
					@{\hspace{2pt}}r@{\hspace{2pt}}       
				}
				\toprule
				description & \drd & \qed \\
				\midrule
				\texttt{\#}training molecules        & 59,696 & 83,161 \\
				\texttt{\#}training (\molx, \moly) pairs  & 125,469 & 92,045 \\
				\texttt{\#}validation molecules           & 500     & 360\\
				\texttt{\#}test molecules              & 1,000   & 800\\
				\midrule
				average similarity of training (\molx, \moly) pairs & {0.6803}  & {0.6578}\\
				average pairwise similarity between training and test molecules & 0.1366 & 0.1227\\
				\midrule
				average training molecule size & 28.76 & 23.22\\
				average training $\{\molx\}$ size & 28.81  & 24.38\\
				average training $\{\moly\}$ size & 28.66  & 22.06 \\
				average test molecule size     & 24.57  & 22.81\\
				\midrule
				average $\{\molx\}$ score                  & 0.1827   & 0.7500 \\
				average $\{\moly\}$ score                  & 0.5366   & 0.8768\\
				average testing molecule score             & 0.0067   & 0.7528\\
				average score improvement in training \mbox{(\molx, \moly)} pairs   & {0.4013} & {0.1355}\\
				\bottomrule
			\end{tabular}
		}
	\end{threeparttable}
	
\end{table*}

{%
We used the {\chembl} dataset~{\cite{gaulton2016}} processed by Olivecrona {\etal}~{\cite{olivecrona2017}}. 
This processed {\chembl} dataset has 1,179,477 molecules in total, and 
each molecule is restricted to have 10 to 50 heavy atoms and only contains atoms 
in \{H, B, C, N, O, F, Si, P, S, Cl, Br, I\}.
We constructed our training data from this processed {\chembl} dataset as follows. 
We first identified 7,743,098 pairs of molecules with similarities $\text{sim}(\molx,\moly)\geq 0.6$ from this 
{\chembl} dataset.
We selected this high similarity threshold because {\molmod} requires training pairs 
different in only one fragment at one disconnection site, and thus should be very similar.
Among these similar molecule pairs, we selected the pairs that satisfy the following property constraints, respectively: 
for the {\drd} optimization task, the {\drd} score difference of the two molecules in a pair 
should be at least 0.2, that is, } 
\begin{equation}
\label{eqn:drd:training}
\drd(\mol_y) - \drd(\mol_x)\geq 0.2; 
\end{equation}
{for the {\qed} optimization task, the {\qed} score difference of the two molecules in a pair should be at least 0.1, 
that is, }
\begin{equation}
\label{eqn:qed:training}
\qed(\mol_y) - \qed(\mol_x)\geq 0.1.
\end{equation}
{Among the molecule pairs that satisfied the above property constraints, respectively, we identified the pairs in which the 
two molecules are different only at one disconnection site. 
Through all these processes, 
we finally identified 125,469 training pairs for the {\drd} task and 92,045 training pairs for the {\qed} task.
The test set for the {\drd} and {\qed} tasks is the benchmark datasets provided by Jin {\etal}~{\cite{jin2019learning}} 
and contains 1,000 molecules and 800 molecules, respectively.
Table~{\ref{tbl:drd_qed_stats}} presents the data statistics for the {\drd} and {\qed} tasks.}
%

%

\begin{figure*}
	\centering
	\begin{small}
		\fbox{\begin{minipage}{0.4\linewidth}
				\centering
				\scriptsize{
				Disconnection sites are highlighted in \colorbox{yellow}{yellow}.
			}
			\end{minipage}
		}
	\end{small}
	\begin{subfigure}{0.9\linewidth}
		\vspace{5pt}
		\caption{}
		\begin{subfigure}[b]{.19\linewidth}
			\centering
			\captionsetup{justification=centering}
			\includegraphics[width=0.6\linewidth]{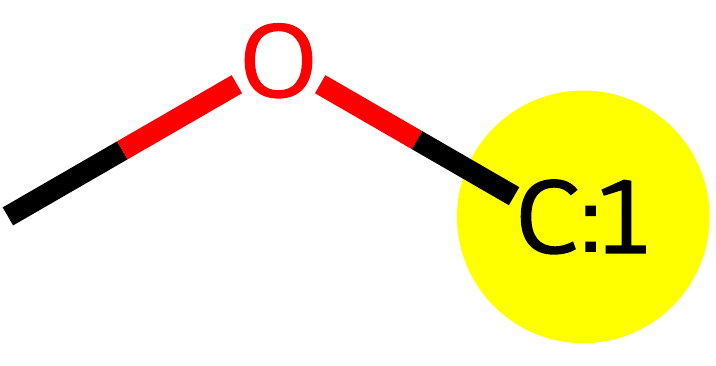}
			\caption*{CO[C:1]}
			\label{fig:drd_del0}
		\end{subfigure}
		%
		\begin{subfigure}[b]{.19\linewidth}
			\centering
			\captionsetup{justification=centering}
			\includegraphics[width=0.5\linewidth]{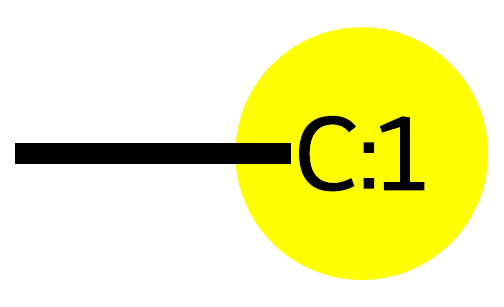}
			\caption*{C[C:1]}
			\label{fig:drd_del1}
		\end{subfigure}%
		%
		\begin{subfigure}[b]{.19\linewidth}
			\centering
			\captionsetup{justification=centering}
			\includegraphics[width=0.4\linewidth]{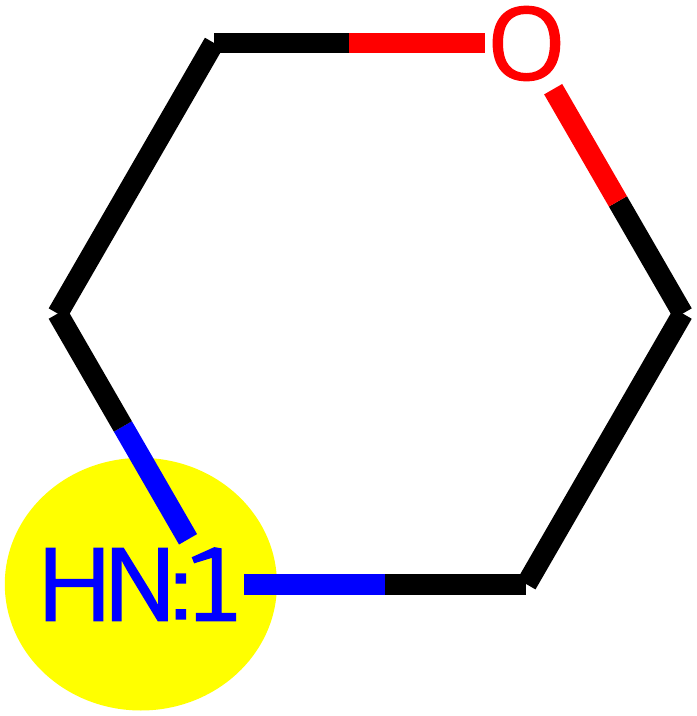}
			\vspace{-3pt}\caption*{C1C[N:1]CCO1}
			\label{fig:drd_del2}
		\end{subfigure}
		%
		\begin{subfigure}[b]{.19\linewidth}
			\centering
			\captionsetup{justification=centering}
			\includegraphics[width=0.45\linewidth]{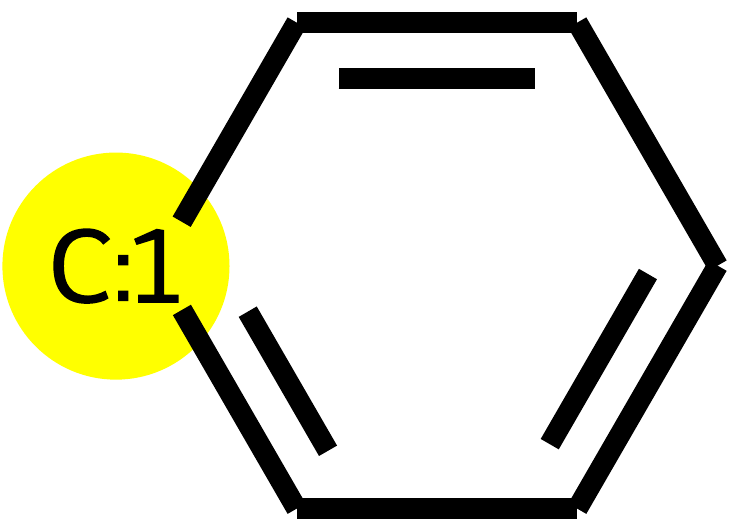}
			\vspace{2pt}\caption*{c1cc[c:1]cc1}
			\label{fig:drd_del3}
		\end{subfigure}
		%
		\begin{subfigure}[b]{.19\linewidth}
			\centering
			\captionsetup{justification=centering}
			\includegraphics[width=0.4\linewidth]{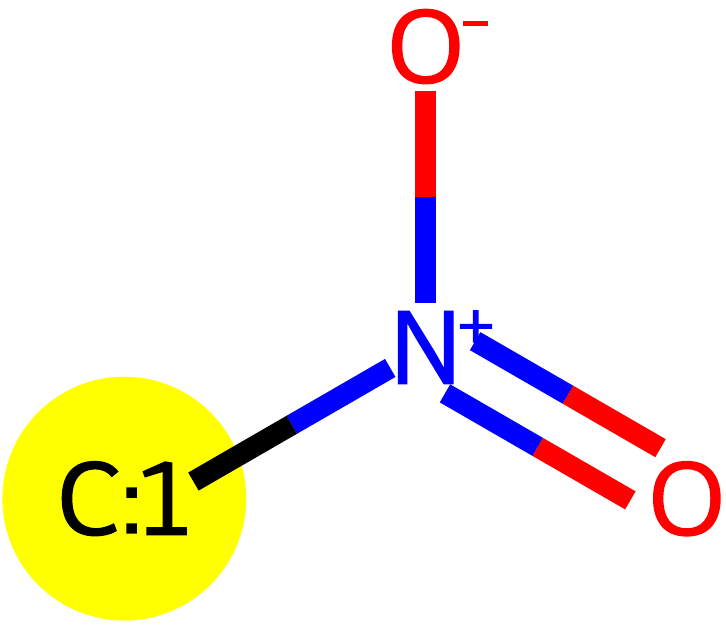}
			\caption*{O=[N+]([O-])[C:1]}
			\label{fig:drd_del4}
		\end{subfigure}
		\label{fig:removal_frag_drd}
	\end{subfigure}
	\begin{subfigure}{0.9\linewidth}
		\vspace{5pt}
		\caption{}
		\begin{subfigure}[b]{.19\linewidth}
			\centering
			\captionsetup{justification=centering}
			\includegraphics[width=0.5\linewidth]{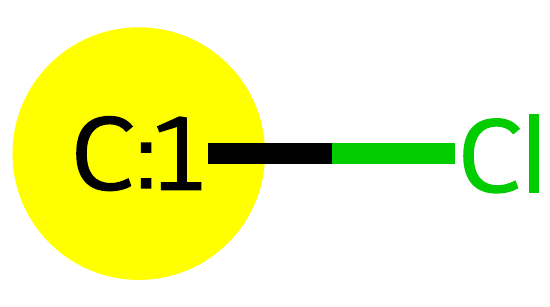}
			\caption*{Cl[C:1]}
			\label{fig:drd_add0}
		\end{subfigure}
		%
		\begin{subfigure}[b]{.19\linewidth}
			\centering
			\captionsetup{justification=centering}
			\includegraphics[width=0.5\linewidth]{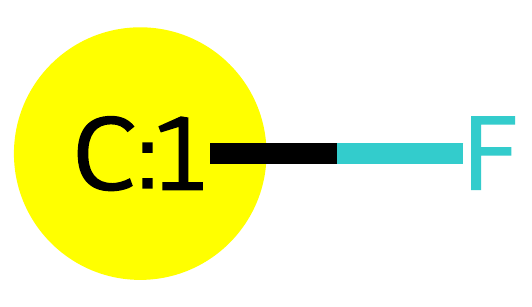}
			\caption*{F[C:1]}
			\label{fig:drd_add1}
		\end{subfigure}%
		%
		\begin{subfigure}[b]{.19\linewidth}
			\centering
			\captionsetup{justification=centering}
			\includegraphics[width=0.5\linewidth]{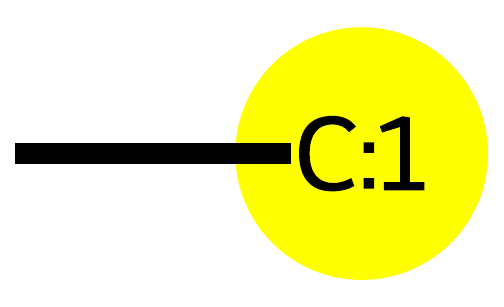}
			\caption*{C[C:1]}
			\label{fig:drd_add2}
		\end{subfigure}
		%
		\begin{subfigure}[b]{.19\linewidth}
			\centering
			\captionsetup{justification=centering}
			\includegraphics[width=0.45\linewidth]{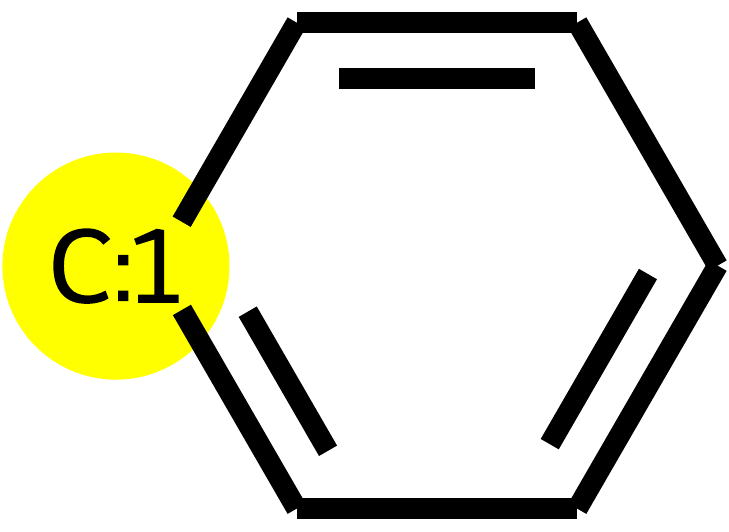}
			\caption*{c1cc[c:1]cc1 }
			\label{fig:drd_add3}
		\end{subfigure}
		%
		\begin{subfigure}[b]{.19\linewidth}
			\centering
			\captionsetup{justification=centering}
			\includegraphics[width=0.5\linewidth]{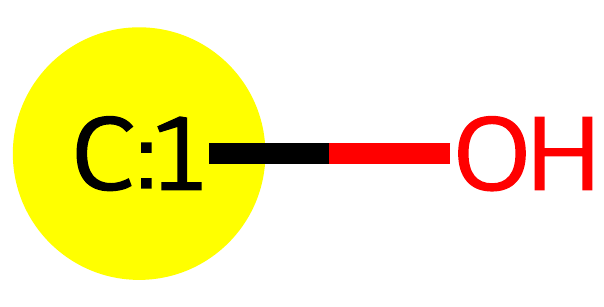}
			\caption*{O[C:1]}
			\label{fig:drd_add4}
		\end{subfigure}
		\label{fig:removal_frag_drd2}
		\vspace{15pt}
	\end{subfigure}
	\vspace{-10pt}
	\caption{\textbf{Popular Fragments in {\drd} Training Data.}
		     \textbf{a,} Visualization of popular removal fragments in \drd training molecules.
			 \textbf{b,} Visualization of popular attaching fragments in \drd training molecules.}
	\label{fig:drd_fragments}
\end{figure*}
\begin{figure*}
	\centering
	\begin{subfigure}{0.9\linewidth}
		\vspace{10pt}
		\caption{}
		\vspace{-10pt}
		\begin{subfigure}[b]{.19\linewidth}
			\centering
			\captionsetup{justification=centering}
			\includegraphics[width=0.43\linewidth]{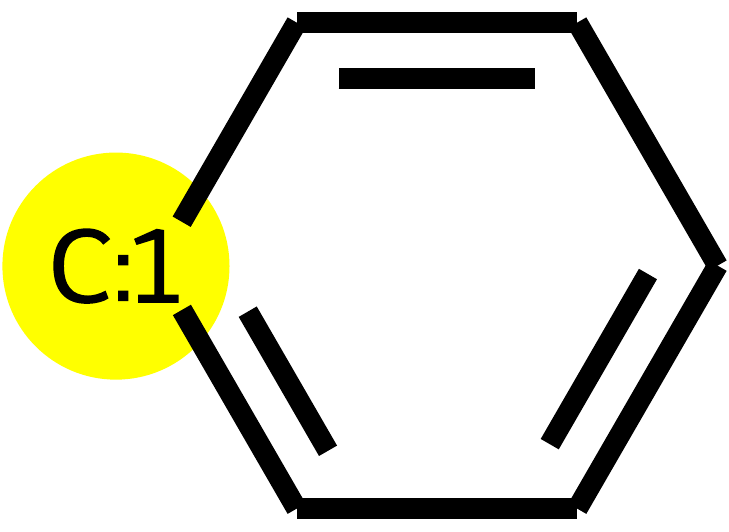}
			\vspace{5pt}\caption*{c1cc[c:1]cc1}
			\label{fig:qed_del0ls}
		\end{subfigure}
		%
		\begin{subfigure}[b]{.19\linewidth}
			\centering
			\captionsetup{justification=centering}
			\includegraphics[width=0.6\linewidth]{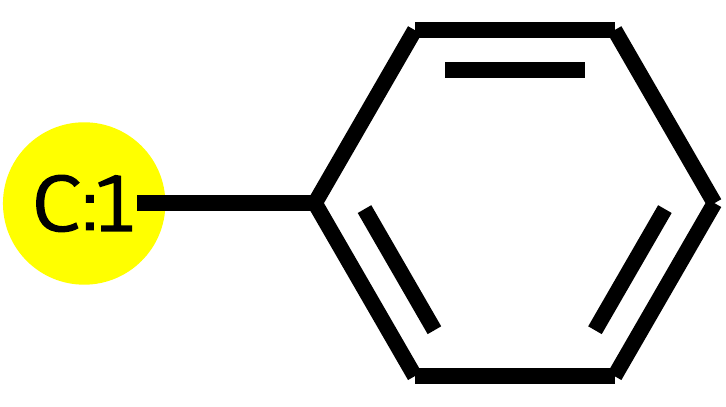}
			\vspace{5pt}\caption*{c1ccc([C:1])cc1}
			\label{fig:qed_del1}
		\end{subfigure}%
		%
		\begin{subfigure}[b]{.19\linewidth}
			\centering
			\captionsetup{justification=centering}
			\includegraphics[width=0.5\linewidth]{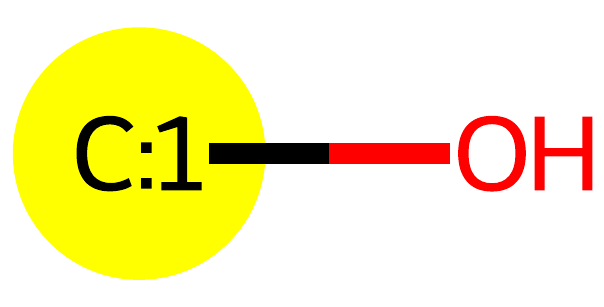}
			\vspace{5pt}\caption*{O[C:1]}
			\label{fig:qed_del2}
		\end{subfigure}
		%
		\begin{subfigure}[b]{.19\linewidth}
			\centering
			\captionsetup{justification=centering}
			\includegraphics[width=0.45\linewidth]{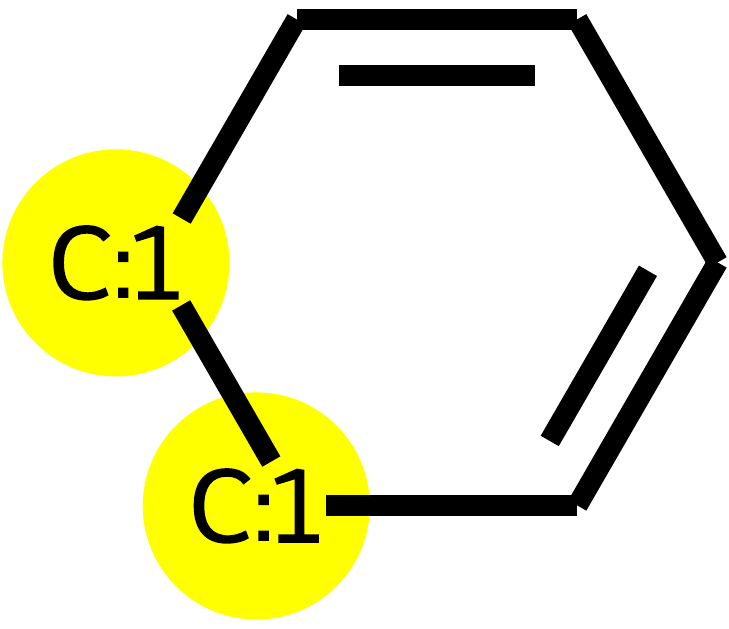}
			\caption*{C1=C[C:1][C:1]C=C1}
			\label{fig:qed_del3}
		\end{subfigure}
		%
		\begin{subfigure}[b]{.19\linewidth}
			\centering
			\captionsetup{justification=centering}
			\includegraphics[width=0.5\linewidth]{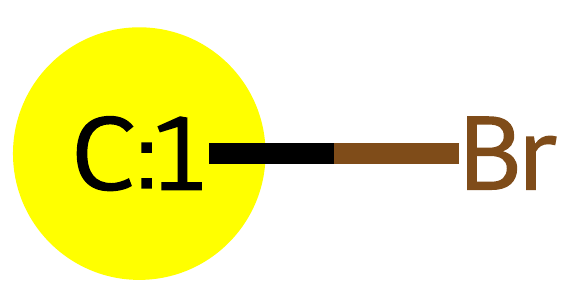}
			\vspace{5pt}\caption*{Br[C:1]}
			\label{fig:qed_del4}
		\end{subfigure}
		\label{fig:removal_frag_qed}
	\end{subfigure}
	\begin{subfigure}{0.9\linewidth}
		\vspace{10pt}
		\caption{}
		\vspace{-10pt}
		\begin{subfigure}[b]{.19\linewidth}
			\centering
			\captionsetup{justification=centering}
			\includegraphics[width=0.5\linewidth]{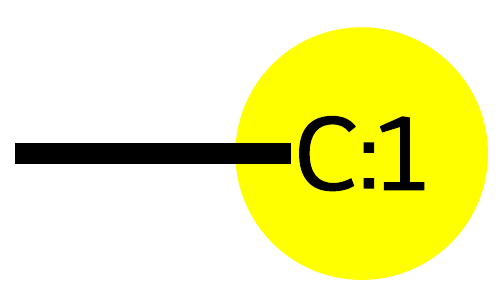}
			\caption*{C[C:1]}
			\label{fig:qed_add0}
		\end{subfigure}
		%
		\begin{subfigure}[b]{.19\linewidth}
			\centering
			\captionsetup{justification=centering}
			\includegraphics[width=0.45\linewidth]{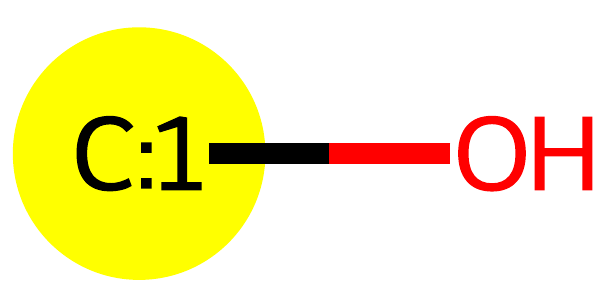}
			\caption*{O[C:1]}
			\label{fig:qed_add1}
		\end{subfigure}%
		%
		\begin{subfigure}[b]{.19\linewidth}
			\centering
			\captionsetup{justification=centering}
			\includegraphics[width=0.5\linewidth]{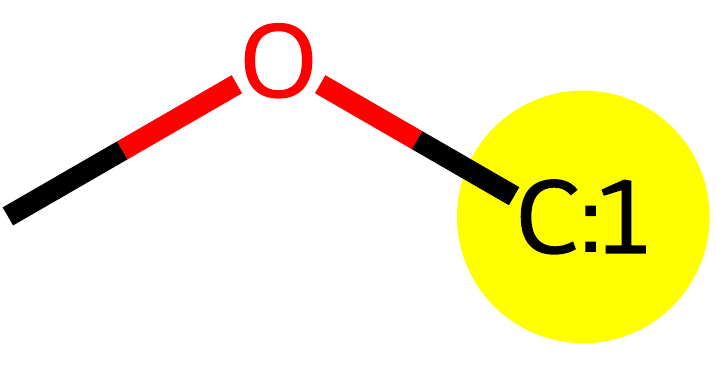}
			\caption*{CO[C:1]}
			\label{fig:qed_add2}
		\end{subfigure}
		%
		\begin{subfigure}[b]{.19\linewidth}
			\centering
			\captionsetup{justification=centering}
			\includegraphics[width=0.5\linewidth]{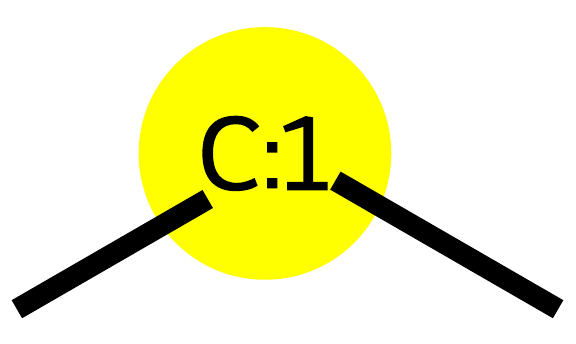}
			\vspace{4pt}\caption*{C[C:1]C}
			\label{fig:qed_add3}
		\end{subfigure}
		%
		\begin{subfigure}[b]{.19\linewidth}
			\centering
			\captionsetup{justification=centering}
			\includegraphics[width=0.45\linewidth]{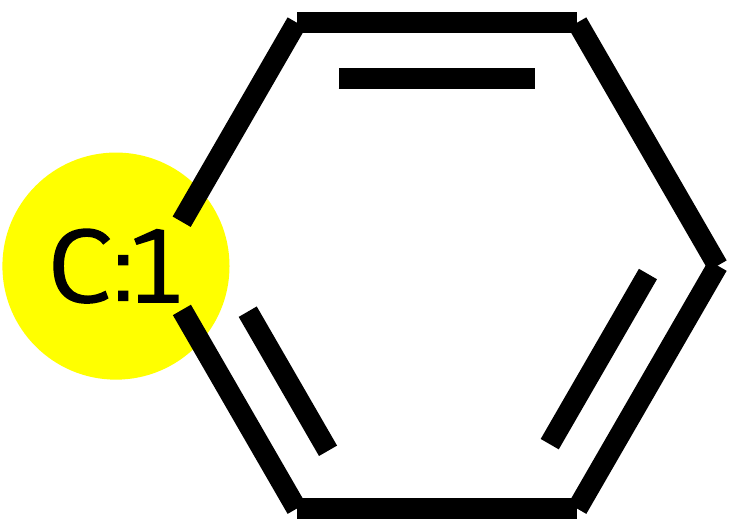}
			\caption*{c1cc[c:1]cc1}
			\label{fig:qed_add4}
		\end{subfigure}
		\label{fig:attach_frag_qed}
		\vspace{15pt}
	\end{subfigure}
	\vspace{-10pt}
	\caption{\textbf{Popular Fragments in {\qed} Training Data.}
			\textbf{a,} Visualization of popular removal fragments in \qed training molecules.
			\textbf{b,} Visualization of popular attaching fragments in \qed training molecules.}
	\label{fig:qed_fragments}
\end{figure*}

%
\subsubsection{{Fragment and Molecule Size Analysis for {\drd} and {\qed} Optimization}}
\label{appendix:drd2_qed_data_analysis}

{Among the training molecules for the {\drd} optimization task, the top-5 most popular fragments that have been removed from {\molx} are:
CO[C3:1] (2.90\%),
C[C3:1] (2.29\%),
C1C[N:1]CCO1 (1.67\%),
c1cc[cH:1]cc1 (1.63\%),
O=[N+]([O-])[C:1] (1.61\%);
the top-5 most popular fragments to be attached into {\moly} are:
Cl[C:1] (4.21\%)
F[C:1] (4.13\%)
C[C:1] (1.95\%)
c1cc[c:1]cc1 (1.89\%)
O[C:1] (1.59\%).
Overall, the attaching fragments are on average of 6.46 atoms and the removed fragments are of 6.51 atoms.
In addition, 7.08\% {\molx} molecules do not have new fragments attached and only have fragments removed, while
9.20\% {\molx} do not have fragments removed and only have fragments attached.
Unlike in {\plogp} training data, molecules in {\drd} training data do not have significant difference in size when they have 
different {\drd} properties. 
Fig.~{\ref{fig:drd_fragments}} presents the popular removal and attaching fragments among {\drd} training molecules. 
}

{
Among the training molecules for the {\qed} optimization task, the top-5 most popular fragments that have been removed from {\molx} are:
c1cc[c:1]cc1 (4.04\%),
c1ccc([C:1])cc1 (2.52\%),
O[C:1] (2.20\%),
C1=C[C:1][C:1]C=C1 (2.03\%),
Br[C:1] (1.44\%);
the top-5 most popular fragments to be attached into {\moly} are:
C[C:1] (5.10\%),
O[C:1] (3.40\%),
CO[C:1] (2.69\%),
C[C:1]C (2.57\%),
c1cc[c:1]cc1 (2.55\%).
Overall, the attaching fragments are on average of 4.17 atoms and the removed fragments are of 6.78 atoms.
In addition, 17.62\% {\molx} molecules do not have new fragments attached and only have fragments removed, while only
5.02\% {\molx} molecules do not have fragments removed and only have fragments attached.
Unlike in {\plogp} and {\drd} training data, molecules in {\qed} training data tend to be smaller when they have better 
{\qed} properties. 
Fig.~{\ref{fig:qed_fragments}} presents the popular removal and attaching fragments among {\qed} training molecules. 
}

\subsection{{Baseline Methods}}
\label{appendix:drd2_qed_baseline}

{We used  {\jtnn}~{\cite{jin2019learning}} and {\hiergtog}~{\cite{jin2020hierarchical}} as the baseline methods to compare 
{\molmod} with. 
These two methods have been demonstrated to achieve the state-of-the-art performance on {\drd} and {\qed} 
optimization~{\cite{jin2020hierarchical}}. 
Note that in the original {\jtnn} and {\hiergtog} manuscripts, their training data are different from the training data we generated 
as in Section~{\ref{appendix:drd2_qed_data}}:  
in their training data, 
molecule pairs have a similarity difference not below 0.4 ($\text{sim}(\molx,\moly)\geq 0.4$); 
for the {\drd} task,  the {\drd} score of molecule $\mol_x$ is in $(0, 0.05]$ and $\mol_y$ in $[0.5,1)$;
for the {\qed} task, the {\qed} score of molecule $\mol_x$ is in $[0.7, 0.8]$ and $\mol_y$ to be $[0.9, 1)$. 
%
To allow a fair comparison, we also trained {\jtnn} and {\hiergtog} on {\molmod}'s training data, and the corresponding models are 
denoted as {\jtnnSTAR} and {\hiergtogSTAR}, respectively.}

{
Same as in the {\plogp} task, in each iteration, {\molmod} samples 20 times from its latent space and decodes 20 output molecules.  
In {\pipeline}, 
the best molecule which satisfies the corresponding similarity constraint and has positive improvement on {\drd} or {\qed}
at each iteration will be fed into the next iteration for further optimization. 	
Each test molecule will be optimized with at most five iterations.
In addition to {\pipeline}, {\pipelineF} is also applied to enable multiple optimized molecules as output. 
}


\begin{table*}
	\caption{{Overall Comparison on Optimizing {\drd}}}
	\label{tbl:drd2_overall}  
	\centering
	\vspace{-5pt}
	\begin{threeparttable}
		\scriptsize{
		\begin{tabular}{
				@{\hspace{3pt}}c@{\hspace{8pt}}
				@{\hspace{8pt}}l@{\hspace{8pt}}
				@{\hspace{3pt}}r@{\hspace{3pt}}
				@{\hspace{3pt}}c@{\hspace{3pt}}
				@{\hspace{3pt}}c@{\hspace{3pt}}
				@{\hspace{4pt}}c@{\hspace{4pt}}
				@{\hspace{3pt}}r@{\hspace{3pt}}
				@{\hspace{3pt}}c@{\hspace{3pt}}
				@{\hspace{3pt}}c@{\hspace{3pt}}
				@{\hspace{4pt}}c@{\hspace{4pt}}
				@{\hspace{3pt}}r@{\hspace{3pt}}
				@{\hspace{3pt}}c@{\hspace{3pt}}
				@{\hspace{3pt}}c@{\hspace{3pt}}
			}
			\toprule
			\multirow{2}{*}{$\delta$} & \multirow{2}{*}{model} & \multicolumn{3}{c}{\OM} &&  \multicolumn{3}{c}{\srone($\drd(\moly)\geq0.5$)} && \multicolumn{3}{c}{\srtwo(imprv $\geq$ 0.2)} \\
			\cmidrule(r){3-5} \cmidrule(r){7-9} \cmidrule(r){11-13}
			& & rate\% & imprv$\pm$std & sim$\pm$std && rate\% & imprv$\pm$std & sim$\pm$std && rate\% & imprv$\pm$std & sim$\pm$std\\ 
			\midrule
			\multirow{6}{*}{0.4} 
			&  \jtnn                & 78.90 & 0.83$\pm$0.18 & 0.44$\pm$0.05 && 78.10 & 0.83$\pm$0.17 & 0.44$\pm$0.05 && 78.30 & 0.83$\pm$0.17 & 0.44$\pm$0.05 \\ 
			& \hiergtog          & 85.40 & 0.81$\pm$0.20 & 0.44$\pm$0.05 && 82.00 & 0.83$\pm$0.16 & 0.44$\pm$0.05 && 84.00 & 0.82$\pm$0.18 & 0.44$\pm$0.05 \\
			\cmidrule(r){2-13}
			& \jtnnSTAR        & 84.40 & 0.49$\pm$0.32 & 0.49$\pm$0.08 && 43.50 & 0.77$\pm$0.15 & 0.49$\pm$0.08 && 61.60 & 0.65$\pm$0.24 & 0.49$\pm$0.08 \\
			& \hiergtogSTAR & 91.50 & 0.53$\pm$0.32 & 0.49$\pm$0.08 && 51.80 & 0.78$\pm$0.15 & 0.49$\pm$0.08 && 70.20 & 0.66$\pm$0.24 & 0.49$\pm$0.08 \\
			& \pipeline           & 98.80 & 0.69$\pm$0.29 & 0.48$\pm$0.07 && 74.90 & 0.83$\pm$0.14 & 0.48$\pm$0.07 && 89.00 & 0.75$\pm$0.22 & 0.48$\pm$0.07 \\
			& \pipelineF         & \bf{99.10} & 0.82$\pm$0.22 & 0.46$\pm$0.05 && \textbf{88.60} & 0.88$\pm$0.12 & 0.46$\pm$0.05 && \bf{95.90} & 0.84$\pm$0.18 & 0.46$\pm$0.05 \\
			\midrule
			\multirow{6}{*}{0.5} 
			& \jtnn     		   & 13.40 & 0.73$\pm$0.23 & 0.54$\pm$0.05 && 12.50 & 0.78$\pm$0.17 & 0.54$\pm$0.05 && 12.60 & 0.77$\pm$0.17 & 0.54$\pm$0.04 \\ 
			& \hiergtog           & 22.40 & 0.63$\pm$0.32 & 0.55$\pm$0.05 && 16.20 & 0.80$\pm$0.16 & 0.54$\pm$0.05 && 18.50 & 0.75$\pm$0.21 & 0.54$\pm$0.05 \\
			\cmidrule(r){2-13}
			& \jtnnSTAR  	   & 64.90 & 0.35$\pm$0.32 & 0.57$\pm$0.07 && 22.50 & 0.74$\pm$0.15 & 0.57$\pm$0.07 && 35.80 & 0.59$\pm$0.23 & 0.57$\pm$0.06 \\
			& \hiergtogSTAR  & 78.10 & 0.39$\pm$0.33 & 0.57$\pm$0.07 && 30.80 & 0.75$\pm$0.15 & 0.56$\pm$0.06 && 45.80 & 0.61$\pm$0.24 & 0.57$\pm$0.06 \\
			& \pipeline            & 94.20 & 0.49$\pm$0.34 & 0.56$\pm$0.05 && 47.70 & 0.79$\pm$0.15 & 0.56$\pm$0.05 && 66.80 & 0.66$\pm$0.24 & 0.56$\pm$0.05 \\
			& \pipelineF          & \bf{97.80} & 0.63$\pm$0.32 & 0.54$\pm$0.04 && \bf{66.60} & 0.82$\pm$0.14 & 0.54$\pm$0.04 && \bf{83.20} & 0.72$\pm$0.23 & 0.54$\pm$0.04 \\	
			\midrule
			\multirow{6}{*}{0.6} 
			& \jtnn                  & 1.60 & 0.69$\pm$0.25 & 0.65$\pm$0.03 && 1.40 & 0.76$\pm$0.16 & 0.64$\pm$0.03 && 1.40 & 0.76$\pm$0.16 & 0.64$\pm$0.03 \\ 
			& \hiergtog           & 4.40 & 0.50$\pm$0.35 & 0.65$\pm$0.04 && 2.60 & 0.77$\pm$0.15 & 0.64$\pm$0.04 && 3.10 & 0.70$\pm$0.21 & 0.65$\pm$0.04 \\
			\cmidrule(r){2-13}
			& \jtnnSTAR         & 41.40 & 0.23$\pm$0.28 & 0.67$\pm$0.06 && 8.40 & 0.72$\pm$0.15 & 0.66$\pm$0.06 && 15.50 & 0.54$\pm$0.23 & 0.66$\pm$0.05 \\ 
			& \hiergtogSTAR  & 52.00 & 0.24$\pm$0.28 & 0.66$\pm$0.06 && 10.50 & 0.72$\pm$0.14 & 0.66$\pm$0.05 && 19.70 & 0.54$\pm$0.23 & 0.66$\pm$0.05 \\
			& \pipeline            & 79.60 & 0.27$\pm$0.31 & 0.65$\pm$0.04 && 19.70 & 0.75$\pm$0.15 & 0.64$\pm$0.03 && 32.40 & 0.59$\pm$0.24 & 0.64$\pm$0.03 \\
			& \pipeline-5         & \bf{87.50} & 0.34$\pm$0.33 & 0.64$\pm$0.04 && \bf{28.90} & 0.76$\pm$0.15 & 0.64$\pm$0.03 && \bf{45.30} & 0.61$\pm$0.24 & 0.63$\pm$0.03 \\		
			\bottomrule
		\end{tabular}
		}
		
		\begin{tablenotes}[normal,flushleft]
			\begin{scriptsize}			
				\item
				\!Columns represent: ``{\rate}": the percentage of optimized molecules in each group ({\OM}, {\srone}, {\srtwo})
				over all test molecules;
				``imprv": the average property improvement;  
				``std": the standard deviation; ``sim": the similarity between the original molecules {\molx} and 
				optimized molecules {\moly}. 
				Best {\rate} values are in \bf{bold}. 
				\par
			\end{scriptsize}
			
		\end{tablenotes}
	\end{threeparttable}
\end{table*}
%
\subsection{{Evaluation Metrics}}
\label{appendix:drd2_qed_metrics}

{We compared different methods by analyzing the following three groups of molecules:}
\begin{itemize}
	\item 
	{Optimized molecules ({\OM}): 
	If an input molecule {\molx} is optimized into {\moly} with any better properties, that is, 
	no constraints on how much better {\moly} is than {\molx}, and {\moly} also satisfies 
	the similarity constraint with {\molx} (i.e., $\text{sim}(\moly - \molx) > \delta$), this molecule is considered as optimized. }
	%
	%
	\item 
	{{\OM} under {\underline{p}}roperty {\underline{i}}mprovement {\underline{c}}onstraints ({\srone}): 
	We also measured the optimized molecules ({\OM} defined as above) that achieved a certain property improvement, 
	following {\jtnn} and {\hiergtog}'s criteria: }
	\begin{enumerate}[label=(\arabic*)]
	\item {For {\drd}, the optimized molecules {\moly} should have {\drd} score no less than 0.5, that is, }
	\begin{equation} 
	\drd(\moly)\geq0.5;  
	\end{equation}
	\item {For {\qed}, the optimized molecules {\moly} should have {\qed} score no less than 0.9, that is, }
	\begin{equation}
	\qed(\moly)\geq0.9. 
	\end{equation}
	\end{enumerate}
	\item 
	{{\OM} following {\underline{tr}}ai{\underline{n}}ing data property constraints ({\srtwo}): 
	a potential issue with {\srone} is that if in the training molecule pairs, the property improvement constraints are not satisfied, imposing 
	the constraints in {\OM} for evaluation is overkilling and tends to underestimate the model performance. 
	Therefore, we also measured the optimized molecules ({\OM} defined as above) that achieved property improvement in a similar 
	degree as in training data:}
	\begin{enumerate}[label=(\arabic*)]
	\item {For {\drd}, the optimized molecules {\moly} should have {\drd} scores such that}
	\begin{equation}
	\drd(\moly)-\drd(\molx)\geq0.2
	\end{equation}
	{that is, same criterion as in training data generation (Equation~{\ref{eqn:drd:training}}).}
	\item {For {\qed}, the optimized molecules {\moly} should have {\qed} scores such that}
	\begin{equation}
	\qed(\moly)-\qed(\molx)\geq0.1,
	\end{equation}
	{that is, same criterion as in training data generation (Equation~{\ref{eqn:qed:training}}).}
	\end{enumerate}
\end{itemize}

{Among each of these three groups of molecules, we measured 
1) the percentage of the optimized 
molecules from that group over all test molecules, referred to as success rate, denoted as \rate, 
2) within each group, the average property improvement of 
optimized {\moly} over {\molx}, and 
3) within each group, the average similarity values between {\molx} and {\moly}.
Note that {\rate} in {\srone} is identical to the success rate used in
Jin {\etal}~{\cite{jin2019learning}}({\jtnn})  
and Jin {\etal}~{\cite{jin2020hierarchical}}({\hiergtog})}.

%
\subsection{{Experimental Results}}
\label{appendix:drd2_qed_result}

	%

\subsubsection{{Experimental Results on Optimizing {\drd}}}
\label{appendix:drd2_qed_result:drd2}

{
Table~{\ref{tbl:drd2_overall}} presents the overall comparison between the baseline methods {\jtnn}, {\hiergtog}, {\jtnnSTAR}, {\hiergtogSTAR}
and our methods {\pipeline} and {\pipelineF} on optimizing the {\drd} property. 
Please note that {\jtnn} and {\hiergtog} used a training dataset generated based on their criteria~{\cite{jin2019learning}}; 
{\jtnnSTAR}, {\hiergtogSTAR}, {\pipeline} and {\pipelineF} used the training data generated as described in Section~{\ref{appendix:drd2_qed_data}};  
all the methods were tested on the same benchmark test set provided by Jin {\etal}~{\cite{jin2019learning}} 
that contains 1,000 molecules. 
We stratified {\jtnn}, {\jtnnSTAR}, {\hiergtog} and {\hiergtogSTAR}'s optimized molecules \moly's using their similarities with respect to 
corresponding \molx's. We used the similarity thresholds $\delta$=0.4, 0.5 and 0.6 for the stratification, 
which also define the similarity constraints in {\pipeline} and {\pipelineF}.
For {\pipelineF}, the property improvement and optimized molecule similarities to those before optimization are always considered on 
the best optimized molecule among all the optimized molecules. 
Under each similarity constraint $\delta$, the optimized molecules within each of the 
{\OM}, {\srone} and {\srtwo} groups are always at least $\delta$-similar to the molecules before optimization. 
%


Table~{\ref{tbl:drd2_overall}} shows that {\pipelineF} consistently achieves the highest success rates of optimized molecules, 
when considering molecules that are ever optimized (in {\OM}), or optimized with a certain property improvement (in {\srone} and {\srtwo}), 
under all the similarity constraints ($\delta$=0.4, 0.5 and 0.6). 
With similarity constraint $\delta$=0.4, that is, each optimized molecule {\moly} has a similarity at least 0.4 with its corresponding {\molx} 
before optimization, {\pipelineF} is able to improve the {\drd} property for 99.10\% of all the test molecules in {\OM}; 
this is a 8.31\% improvement over the best baseline method {\hiergtogSTAR}, which achieves a 91.5\% success rate. 
Among the optimized molecules in {\OM}, the average {\drd} property improvement (``imprv$\pm$std") from {\pipelineF} 
(0.82$\pm$0.22) is among the best compared to the results from other baseline methods (e.g., 0.83$\pm$0.18 from {\jtnn}); the average 
pair-wise similarities between the test molecules and their optimized molecules from {\pipelineF} (0.46$\pm$0.05) are also comparable 
to those of other methods. 
Compared to {\OM} with $\delta$=0.4, the optimized molecules in {\srone} and {\srtwo} have a lower success rate because not all the 
optimized molecules in {\OM} have \mbox{$\drd(\moly)\ge 0.5$} or {\drd} improvement no less than 0.2. 
However, {\pipelineF} still achieves the best success rates (88.60\% in {\srone} and 95.90\% in {\srtwo}, respectively) compared to 
the baseline methods. 
In terms of the property improvement among the optimized molecules in {\srone}, {\pipelineF} achieves the best (0.88$\pm$0.12); 
in terms of pair-wise similarities, {\pipelineF} has very comparable performance (0.46$\pm$0.05) compared to the baseline methods. 
The same conclusions hold true in {\srtwo}. 

With similarity constraint $\delta$=0.5 and 0.6, we observed the same trends as those with $\delta$=0.4:
consistently in {\OM}, {\srone} and {\srtwo}, {\pipelineF} achieves the best success rates ({\rate}),  best or competitive property improvement 
(imprv$\pm$std), and competitive pair-wise similarities (sim$\pm$std) between the test molecules and their optimized molecules. 
Particularly, with $\delta$=0.5 and 0.6, {\pipelineF}'s success rates in {\OM}, {\srone} and {\srtwo} were substantially higher than those of the 
baseline methods:
for example, with $\delta$=0.5 and 0.6 in {\srone}, {\pipelineF} has a 116.23\% and 175.24\%, respectively, higher {\rate} than that of {\hiergtogSTAR} 
(66.60\% vs 30.80\%; 28.90\% vs 10.50\%); 
in {\srtwo}, {\pipelineF}'s {\rate} is 81.66\% and 129.95\%, respectively, higher than that of {\hiergtogSTAR} (83.20\% vs 45.80\%; 45.30\% vs 19.70\%).  
This demonstrates the strong capability of {\pipelineF} in optimizing molecules and improving {\drd} properties.  
}

\begin{table*}
	\caption{{Overall Comparison on Optimizing {\qed}}}
	\label{tbl:qed_overall}  
	\centering
	\vspace{-5pt}
	\begin{threeparttable}
		\scriptsize{
		\begin{tabular}{
				@{\hspace{3pt}}c@{\hspace{8pt}}
				@{\hspace{8pt}}l@{\hspace{8pt}}
				@{\hspace{3pt}}r@{\hspace{3pt}}
				@{\hspace{3pt}}c@{\hspace{3pt}}
				@{\hspace{3pt}}c@{\hspace{3pt}}
				@{\hspace{4pt}}c@{\hspace{4pt}}
				@{\hspace{3pt}}r@{\hspace{3pt}}
				@{\hspace{3pt}}c@{\hspace{3pt}}
				@{\hspace{3pt}}c@{\hspace{3pt}}
				@{\hspace{4pt}}c@{\hspace{4pt}}
				@{\hspace{3pt}}r@{\hspace{3pt}}
				@{\hspace{3pt}}c@{\hspace{3pt}}
				@{\hspace{3pt}}c@{\hspace{3pt}}
			}
			\toprule
			\multirow{2}{*}{$\delta$} & \multirow{2}{*}{model} & \multicolumn{3}{c}{\OM} &&  \multicolumn{3}{c}{\srone($\qed(\moly)\geq$0.9)} && \multicolumn{3}{c}{\srtwo(imprv$\geq$0.1)} \\
			\cmidrule(r){3-5} \cmidrule(r){7-9} \cmidrule(r){11-13}
			& & rate\% & imprv$\pm$std & sim$\pm$std && rate\% & imprv$\pm$std & sim$\pm$std && rate\% & imprv$\pm$std & sim$\pm$std\\ 
			%
			\midrule
			\multirow{6}{*}{0.4} 
			& \jtnn        & 71.00 & 0.16$\pm$0.04 & 0.47$\pm$0.06 && 60.50 & 0.17$\pm$0.03 & 0.47$\pm$0.06 && 67.38 & 0.17$\pm$0.03 & 0.47$\pm$0.07 \\ 
			& \hiergtog   & 86.50 & 0.17$\pm$0.04 & 0.46$\pm$0.06 && \textbf{75.12} & 0.18$\pm$0.03 & 0.46$\pm$0.06 && 82.38 & 0.17$\pm$0.03 & 0.46$\pm$0.06 \\
			\cmidrule(r){2-13}
			& \jtnnSTAR  & 93.50 & 0.13$\pm$0.06 & 0.55$\pm$0.10 && 40.50 & 0.17$\pm$0.03 & 0.54$\pm$0.09 && 68.50 & 0.15$\pm$0.03 & 0.54$\pm$0.09  \\
			& \hiergtogSTAR & 91.75 & 0.13$\pm$0.06 & 0.52$\pm$0.10 && 37.12 & 0.17$\pm$0.03 & 0.52$\pm$0.09 && 65.88 & 0.15$\pm$0.03 & 0.53$\pm$0.10\\
			& \pipeline   & 96.38 & 0.13$\pm$0.05 & 0.52$\pm$0.09 && 40.00 & 0.17$\pm$0.03 & 0.51$\pm$0.08 && 70.00 & 0.16$\pm$0.03 & 0.51$\pm$0.08 \\
			& \pipeline-5  & \textbf{99.12} & 0.15$\pm$0.04 & 0.48$\pm$0.07 && 66.25 & 0.18$\pm$0.03 & 0.48$\pm$0.07 && \textbf{87.62} & 0.17$\pm$0.03 & 0.48$\pm$0.07  \\
			\midrule
			\multirow{6}{*}{0.5} 
			& \jtnn        & 42.38 & 0.15$\pm$0.04 & 0.56$\pm$0.05 && 30.25 & 0.17$\pm$0.03 & 0.55$\pm$0.05 && 37.88 & 0.16$\pm$0.03 & 0.56$\pm$0.05 \\ 
			& \hiergtog    & 55.00 & 0.16$\pm$0.04 & 0.55$\pm$0.05 && 40.38 & 0.17$\pm$0.03 & 0.55$\pm$0.05 && 50.38 & 0.16$\pm$0.03 & 0.55$\pm$0.05 \\
			\cmidrule(r){2-13}
			& \jtnnSTAR & 86.62 & 0.12$\pm$0.06 & 0.60$\pm$0.08 && 30.50 & 0.17$\pm$0.03 & 0.60$\pm$0.07 && 56.12 & 0.15$\pm$0.03 & 0.60$\pm$0.07 \\
			& \hiergtogSTAR & 84.25 & 0.11$\pm$0.06 & 0.59$\pm$0.08 && 26.75 & 0.17$\pm$0.03 & 0.59$\pm$0.06 && 53.25 & 0.15$\pm$0.03 & 0.60$\pm$0.07 \\
			& \pipeline   & 89.25 & 0.12$\pm$0.05 & 0.59$\pm$0.07 && 27.25 & 0.17$\pm$0.03 & 0.58$\pm$0.06 && 53.62 & 0.15$\pm$0.03 & 0.58$\pm$0.06 \\
			& \pipeline-5  & \textbf{98.62} & 0.13$\pm$0.05 & 0.56$\pm$0.06 && \textbf{43.38} & 0.17$\pm$0.03 & 0.56$\pm$0.05 && \textbf{71.25} & 0.16$\pm$0.03 & 0.56$\pm$0.05  \\	
			\midrule
			\multirow{6}{*}{0.6} 
			&  \jtnn        & 17.62 & 0.13$\pm$0.05 & 0.65$\pm$0.05 && 10.12 & 0.16$\pm$0.03 & 0.65$\pm$0.04 && 13.38 & 0.15$\pm$0.03 &   0.65$\pm$0.04 \\ 
			& \hiergtog    & 20.25 & 0.14$\pm$0.06 & 0.65$\pm$0.05 && 12.00 & 0.17$\pm$0.03 & 0.65$\pm$0.04 && 15.75 & 0.16$\pm$0.03 & 0.65$\pm$0.04 \\
			\cmidrule(r){2-13}
			& \jtnnSTAR  & 73.62 & 0.10$\pm$0.07 & 0.68$\pm$0.07 && \textbf{18.88} & 0.17$\pm$0.03 & 0.66$\pm$0.06 && 38.62 & 0.15$\pm$0.03 & 0.66$\pm$0.05 \\ 
			& \hiergtogSTAR & 70.88 & 0.10$\pm$0.07 & 0.67$\pm$0.06 && 17.38 & 0.17$\pm$0.03 & 0.66$\pm$0.05 && 37.25 & 0.15$\pm$0.03 & 0.66$\pm$0.05 \\
			& \pipeline    & 66.25 & 0.09$\pm$0.06 & 0.66$\pm$0.05 && 12.25 & 0.16$\pm$0.03 & 0.65$\pm$0.04 && 29.62 & 0.14$\pm$0.03 & 0.65$\pm$0.04 \\
			& \pipeline-5  & \textbf{89.25} & 0.09$\pm$0.07 & 0.66$\pm$0.05 && 18.62 & 0.17$\pm$0.03 & 0.65$\pm$0.04 && \textbf{39.50} & 0.15$\pm$0.03 & 0.65$\pm$0.04  \\		
			\bottomrule
		\end{tabular}
		}
		
		\begin{tablenotes}[normal,flushleft]
			\begin{scriptsize}			
				\item
				{\!Columns represent: ``{\rate}": the percentage of optimized molecules in each group ({\OM}, {\srone}, {\srtwo})
					over all test molecules;
					``imprv": the average property improvement;  
					``std": the standard deviation; ``sim": the similarity between the original molecules {\molx} and 
					optimized molecules {\moly}. 
					Best {\rate} values are in \bf{bold}. }
				\par
			\end{scriptsize}
			
		\end{tablenotes}
	\end{threeparttable}
\end{table*}

\subsubsection{{Experimental Results on Optimizing {\qed}}}
\label{appendix:drd2_qed_result:qed}

{
Table~{\ref{tbl:qed_overall}} presents the overall comparison between the baseline methods {\jtnn}, {\hiergtog}, {\jtnnSTAR}, {\hiergtogSTAR} 
and our methods {\pipeline} and {{\pipelineF}} on optimizing the {\qed} property. 
Overall, the trends are very similar with what we observed in {\drd} optimization: {\pipelineF} achieves the best success rates 
in {\OM}, {\srone} and {\srtwo} with all the similarity constraints ($\delta$=0.4, 0.5 and 0.6), except in {\srone} with $\delta$=0.4 and 0.6, 
where {\pipelineF} has the second best success rates (with $\delta$=0.6, the difference with the best success rate is very minimum). 
}

%


%
%

\subsubsection{Parameters for Reproducibility of Optimizing \drd and \qed}
\label{appendix:param:drd2qed}


{
We tuned the hyper-parameters of our models for {\drd} and {\qed} optimization tasks with the grid-search algorithms in the parameter 
spaces as presented in Table~{\ref{tbl:drd_qed_hyper}}.
We determined the optimal hyper-parameters according to the success rate {\rate} of {\pipeline} in {\srone} under $\delta$=0.4 
over the validation molecules provided by Jin {\etal}~{\cite{jin2019learning}}.
In each iteration, we randomly sampled $K$=20 latent vectors and output 20 optimized candidates.
}

{
For {\drd}, the optimal dimension of all the hidden layers is 320 and the dimension of latent embedding {\latent} is 64 (i.e., 32 for $\latent^-$ and $\latent^+$, respectively).
The optimal iterations of graph message passing {\GMPN} and tree message passing {\TMPN} are 6 and 5, respectively.
The optimal {\molmod} for {\drd} has 3.09M parameters and can achieve the best performance with 3 epochs of training.
For {\qed}, the optimal dimension of all the hidden layers is 256 and the dimension of latent embedding {\latent} is 32 (i.e., 16 for $\latent^-$ and $\latent^+$, respectively).
The optimal iterations of graph message passing {\GMPN} and tree message passing {\TMPN} is 4.
The optimal {\molmod} for {\qed} has 1.81M parameters and can achieve the best performance with 4 epochs of training.
Other training details such as learning rate and KL regularization weight are the same as demonstrated in Section~{\ref{appendix:logp:param}}. 
}


\begin{table*}
	\centering
	\caption{{Hyper-Parameter Space for {\drd} and {\qed} Optimization}}
	\label{tbl:drd_qed_hyper}
	\begin{threeparttable}
		\scriptsize{
		\begin{tabular}{
				@{\hspace{2pt}}l@{\hspace{5pt}} 
				@{\hspace{2pt}}l@{\hspace{2pt}}         
			}
			\toprule
			Hyper-parameters &  Space\\
			\midrule
			hidden layer dimension         & \{64, 128, 256, 320\} \\
			atom/node embedding dimension &  \{64, 128, 256, 320\} \\
			$\latent^{\add}$/$\latent^{\delete}$ dimension        & \{8, 16, 32\} \\
			\# iterations of \GMPN  & \{4, 5, 6, 7\} \\
			\# iterations of \TMPN  & \{3, 4, 5\} \\
			\# sampling             & 20          \\
			\bottomrule
		\end{tabular}
	}
		%
		%
	\end{threeparttable}
\end{table*}


\section{Experimental Results on Multi-Property Optimization of \drd and \qed}
\label{appendix:multi_drd2_qed}
%
Here we demonstrate the effectiveness of \molmod for multi-property optimization.
We conduct experiments to simultaneously improve two molecular properties (\qed and \drd).
Specifically, given a molecule that is not much drug-like (i.e., with a low \drd score) and doesn't bind well to the \drd receptor (i.e., with a low \qed score),
the objective of the task is to modify this molecule to a drug-like molecule bound well to the \drd (i.e., with both high 
\drd and \qed scores).
The assessment of {\drd} and {\qed} properties is as described in the Section~\ref{appendix:drd2_qed}.

\subsection{Training Data Generation for Multi-Property Optimization}
\label{appendix:multi_drd2_qed_data}
%
We used the training molecule pairs for \drd task derived from CheMBL dataset as in Section~\ref{appendix:drd2_qed_data} to 
construct the training data for this multi-property optimization task.
Recall that each molecule pair $(\molx, \moly)$ for the {\drd} task has similarity $\text{sim}(\molx, \moly)\geq 0.6$ and satisfies the following property constraint:
\begin{equation}
\drd(\moly) - \drd(\molx) \geq 0.2;
\end{equation}
Among these molecule pairs, we first selected the pairs that also have differences on \qed scores, that is,
\begin{equation}
\qed(\molx) < 0.6 \ \text{AND} \ \qed(\moly) \geq 0.6.   
\end{equation}
Among the molecule pairs that satisfied the above two property constraints, 
we first identified the pairs in which two molecules are 
different only at one disconnection site.
We finally identified 14,230 training pairs for this task.
The test set for this task includes 800 molecules that do not appear in the training pairs and have the property 
$\drd(\mol)<0.5$ and 
$\qed(\mol)< 0.6$.
Table~\ref{tbl:multi_drd_qed_stats} presents the data statistics for this multi-property optimization task.

\begin{table*}
	\centering
	\caption{Data Statistics for Multi-Property Optimization of \drd and \qed}
	\label{tbl:multi_drd_qed_stats}
	\begin{threeparttable}
		\scriptsize{
		\begin{tabular}{
				@{\hspace{2pt}}p{0.4\linewidth}@{\hspace{5pt}}
				@{\hspace{2pt}}r@{\hspace{2pt}}      
			}
			\toprule
			description & \drd \& \qed \\
			\midrule
			\texttt{\#}training molecules        & 10,121 \\
			\texttt{\#}training (\molx, \moly) pairs  & 14,230 \\
			\texttt{\#}validation molecules           & 200\\
			\texttt{\#}test molecules              & 800\\
			\midrule
			average similarity of training (\molx, \moly) pairs   & {0.6721}\\
			average pairwise similarity between training and test molecules & {0.1257}\\
			\midrule
			average training molecule size & 30.08\\
			average training $\{\molx\}$ size &  32.42\\
			average training $\{\moly\}$ size &  27.16 \\
			average test molecule size     & 32.05\\
			\midrule
			average $\{\drd(\molx)\}$ score            & 0.1788\\
			average $\{\drd(\moly)\}$ score            & 0.4957\\
			average $\{\qed(\molx)\}$ score            & 0.5033\\
			average $\{\qed(\moly)\}$ score            & 0.6963\\
			average \drd score of test molecules     & 0.1102\\
			average \qed score of test molecules     & 0.4241\\
			average \drd score improvement in training \mbox{(\molx, \moly)} pairs   & {0.4356}\\
			average \qed score improvement in training \mbox{(\molx, \moly)} pairs   & {0.2240}\\
			\bottomrule
		\end{tabular}
	}
	\end{threeparttable}
\end{table*}

%
\subsubsection{Fragment and Molecule Size Analysis for Multi-Property Optimization}
\label{appendix:multi_drd2_qed_data_analysis}
%
Among the training molecules for the multi-property optimization task, the top-5 most popular fragments that have been removed from 
$\molx$ are: 
O=[N+]([O-])[C3:1] (5.17\%),
c1cc[c:1]cc1 (1.91\%),
c1ccc([C3:1])cc1 (1.90\%),
c1ccc2c(c1)ccc[c:1]2 (1.25\%),
CO[C3:1] (1.15\%);
the top-5 most popular fragments to be attached into \moly are:
F[C3:1] (5.14\%),
Cl[C3:1] (3.32\%),
C[C3:1] (2.99\%),
O[C3:1]  (2.31\%),
c1cc[c:1]cc1 (2.14\%).
In addition, 16.75\% \molx do not have new fragments attached and only have fragments removed, 
while only a few \molx do not have fragments removed and only have fragments attached.
This is likely due to the fact that smaller molecules have better \qed properties.

\subsection{Baseline Methods}
\label{appendix:multi_drd2_qed_baseline}
%
We used the same baselines \jtnn and \hiergtog as in the \drd and \qed tasks in Section~\ref{appendix:drd2_qed}. 
These two methods have been demonstrated to achieve the state-of-the-art performance on \drd and \qed optimization, respectively, and thus are
strong baselines on the multi-property optimization task on \drd and \qed.
Note that the original \jtnn and \hiergtog did not have the experiments on this multi-property optimization task
in their manuscripts.
Therefore, we can only train \jtnn and \hiergtog with our dataset, and the corresponding models are denoted as 
$\jtnn(m)$ and $\hiergtog(m)$, respectively.

Same as in the single-property optimization task, in each iteration, \molmod sampled 20 times from 
its latent space and decodes 20 output molecules.
At each iteration, \pipeline and \pipelineF used the sum of \drd and \qed scores
to select the best optimized molecules under the corresponding similarity constraint.
The selected molecules were fed into the next iteration for further optimization.
Each test molecule was optimized in at most five iterations.

\begin{table*}
	\captionof{table}{Overall Comparison on Multi-Property Optimization of \drd and \qed}
	\label{tbl:multi_drd2_pip}  
	\centering
	\vspace{-5pt}
		\begin{threeparttable}
			\scriptsize{
			\begin{tabular}{
					@{\hspace{10pt}}c@{\hspace{8pt}}
					@{\hspace{20pt}}l@{\hspace{20pt}}
					@{\hspace{3pt}}r@{\hspace{3pt}}
					@{\hspace{3pt}}c@{\hspace{3pt}}
					@{\hspace{3pt}}c@{\hspace{3pt}}
					@{\hspace{3pt}}c@{\hspace{25pt}}
					@{\hspace{3pt}}r@{\hspace{3pt}}
					@{\hspace{3pt}}c@{\hspace{3pt}}
					@{\hspace{3pt}}c@{\hspace{3pt}}
					@{\hspace{3pt}}c@{\hspace{3pt}}
					%
					@{\hspace{3pt}}l@{\hspace{3pt}}
					@{\hspace{20pt}}r@{\hspace{3pt}}
					@{\hspace{3pt}}c@{\hspace{3pt}}
					@{\hspace{3pt}}c@{\hspace{10pt}}
				}
				\toprule
				\multirow{2}{*}{$\delta$} & \multirow{2}{*}{model} & \multicolumn{3}{c}{\srone($\drd\ge0.5$)} &&  \multicolumn{3}{c}{{\srone($\qed\ge0.6$)}} && & \multicolumn{3}{c}{\hspace{-25pt}\srone({$\drd\ge0.5$ \&\& $\qed\ge0.6$})} \\
				\cmidrule(r){3-5} \cmidrule(r){7-9} \cmidrule(r){11-14}
				& & rate\% & imprv$\pm$std & sim$\pm$std && rate\% & imprv$\pm$std & sim$\pm$std &&& rate\% & imprv$\pm$std & sim$\pm$std\\ 
				\midrule
				\multirow{4}{*}{0.4} 
				& \jtnnSTAR        & 10.62 & 0.53$\pm$0.20 & 0.54$\pm$0.11 && 21.00 & 0.27$\pm$0.13 & 0.55$\pm$0.12 &&& 8.62 & 0.82$\pm$0.22 & 0.55$\pm$0.11 \\
				& \hiergtogSTAR    & 15.62 & 0.58$\pm$0.18 & 0.51$\pm$0.10 && 26.00 & 0.30$\pm$0.13 & 0.51$\pm$0.10 &&& 14.50 & 0.85$\pm$0.21 & 0.51$\pm$0.10 \\
				& \pipeline        & 48.88 & 0.63$\pm$0.17 & 0.51$\pm$0.10 && 54.00 & 0.32$\pm$0.13 & 0.49$\pm$0.08 &&& 24.62 & 0.92$\pm$0.20 & 0.49$\pm$0.08 \\
				& \pipelineF       & \textbf{67.75} & 0.71$\pm$0.15 & 0.47$\pm$0.07 && \textbf{82.25} & 0.36$\pm$0.13 & 0.48$\pm$0.07 &&& \textbf{38.88} & 0.95$\pm$0.21 & 0.48$\pm$0.06 \\
				\midrule
				\multirow{4}{*}{0.5} 
				& \jtnnSTAR  	  & 6.88 & 0.53$\pm$0.19 & 0.60$\pm$0.09 && 13.12 & 0.26$\pm$0.12 & 0.62$\pm$0.10 &&& 5.88 & 0.82$\pm$0.21 & 0.60$\pm$0.09 \\
				& \hiergtogSTAR   & 8.38 & 0.55$\pm$0.17 & 0.59$\pm$0.08 && 15.25 & 0.24$\pm$0.12 & 0.60$\pm$0.08 &&& 7.88 & 0.80$\pm$0.21 & 0.60$\pm$0.08 \\
				& \pipeline       & 41.25 & 0.58$\pm$0.17 & 0.59$\pm$0.07 && 44.75 & 0.27$\pm$0.12 & 0.59$\pm$0.07 &&& 16.62 & 0.82$\pm$0.20 & 0.59$\pm$0.07 \\
				& \pipelineF      & \textbf{58.38} & 0.66$\pm$0.15 & 0.56$\pm$0.06 && \textbf{69.25} & 0.32$\pm$0.12 & 0.57$\pm$0.06 &&& \textbf{27.25} & 0.84$\pm$0.23 & 0.57$\pm$0.07 \\	
				\midrule
				\multirow{4}{*}{0.6}
				& \jtnnSTAR       & 2.88 & 0.49$\pm$0.15 & 0.68$\pm$0.06 && 7.25 & 0.24$\pm$0.11 & 0.70$\pm$0.07 &&& 2.62 & 0.75$\pm$0.18 & 0.68$\pm$0.06 \\ 
				& \hiergtogSTAR   & 4.38 & 0.53$\pm$0.15 & 0.66$\pm$0.05 && 8.12 & 0.23$\pm$0.12 & 0.67$\pm$0.06 &&& 4.12 & 0.75$\pm$0.20 & 0.66$\pm$0.05  \\
				& \pipeline       & 26.88 & 0.53$\pm$0.16 & 0.67$\pm$0.06 && 29.12 & 0.24$\pm$0.11 & 0.67$\pm$0.05 &&& 9.12 & 0.75$\pm$0.19 & 0.67$\pm$0.05 \\
				& \pipelineF      & \textbf{34.12} & 0.58$\pm$0.17 & 0.66$\pm$0.05 && \textbf{43.88} & 0.27$\pm$0.11 & 0.66$\pm$0.05 &&& \textbf{13.00} & 0.78$\pm$0.22 & 0.66$\pm$0.06\\		
				\bottomrule
			\end{tabular}
			}
			
			\begin{tablenotes}[normal,flushleft]
				\begin{scriptsize}			
					\item
					\!Columns represent: 
					``{\srone($\drd\ge0.5$)}": the optimized molecules that have a {\drd} score no less than 0.5;
					``{\srone($\qed\ge0.5$)}": the optimized molecules that have a {\qed} score no less than 0.6;
					``{\srone($\drd\ge0.5$ \&\& $\qed\ge0.6$)}": the optimized molecules that have a {\drd} score no less than 0.5 and at the same time 
					a {\qed} score no less than 0.6;
					``{\rate}": the percentage of optimized molecules in each group
					over all test molecules;
					``imprv": the average property improvement on {\drd} or {\qed} for the group {\srone($\drd\ge0.5$)} or {\srone($\qed\ge0.6$)}, or the average of sum of property improvements on {\drd} and {\qed} scores
					for the group {\srone($\drd\ge0.5$ \&\& $\qed\ge0.6$)};  
					``std": the standard deviation; ``sim": the similarity between the original molecules {\molx} and 
					optimized molecules {\moly}. 
					Best {\rate} values are in \bf{bold}.
					\par
				\end{scriptsize}
			\end{tablenotes}
		\end{threeparttable}
\end{table*}

\subsection{Evaluation Metrics}
\label{appendix:multi_drd2_qed_metric}
%
Similarly to that in Section~\ref{appendix:drd2_qed_metrics}, we compared different methods by 
analyzing \srone on different groups of molecules as follows in the test set 
with respect to different property constraints:
\begin{itemize}
	\item \srone for \drd:
	The optimized molecules \moly should have \drd score no less than 0.5, that is,
	\begin{equation}
	\drd(\moly) \geq 0.5. 
	\end{equation}
	\item \srone for \qed:
	The optimized molecules \moly should have \qed score no less than 0.6, that is,
	\begin{equation}
	\qed(\moly) \geq 0.6. 
	\end{equation}
	\item \srone for \drd and \qed:
	The optimized molecules \moly should have \drd score no less than 0.5 and \qed score 
	no less than 0.6, that is,
	\begin{equation}
	\drd(\moly) \geq 0.5 \ \&\& \ \qed(\moly) \geq 0.6. 
	\end{equation}
\end{itemize}
	Note that in this multi-property optimization task, molecules that have been optimized to have
	$\drd(\moly) \geq 0.5$ or
	$\qed(\moly) \geq 0.6$ may or may not satisfy the \qed or \drd constraint.
	The group \srone for \drd or \srone for \qed only includes the optimized molecules that satisfy the 
	\drd or \qed constraint alone. 
	The group \mbox{\srone for \drd and \qed} includes the optimized molecules that satisfy both the 
	\drd and \qed constraints. In other words, this group is the intersection of the above two groups. 
	Among each of these three groups of molecules, we used the success rate \rate, the average property improvement 
of \moly over \molx, and the average similarity values between \molx and \moly to compare the different 
results.

\subsection{Experimental Results}
\label{appendix:multi_drd2_qed_result}
%
Table~\ref{tbl:multi_drd2_pip} presents the overall comparison between the baseline methods $\jtnn(m)$ and 
$\hiergtog(m)$ and our methods \pipeline and \pipelineF on the multi-property optimization problem.
All the methods were trained for 30 epochs on the same training data as generated in Section~\ref{appendix:multi_drd2_qed_data}.
We then tuned the hyper-parameters and found the best model for each method using a validation set with 200 molecules.
The selected best models were tested on the test set that contains 800 molecules.
Following Section~\ref{appendix:drd2_qed_result}, we used the similarity thresholds $\delta$=0.4, 0.5 and 0.6 to stratify 
the optimized molecules of all the methods.

Table~\ref{tbl:multi_drd2_pip} shows that \pipeline and \pipelineF significantly outperform two baseline methods $\jtnn(m)$ and $\hiergtog(m)$
on the success rates of optimized molecules under all the similarity constraints ($\delta$=0.4, 0.5 and 0.6).
With similarity constraint $\delta$=0.4, that is, each optimized molecule \moly has a similarity at least 0.4 with its corresponding 
\molx before optimization, \pipeline and \pipelineF is able to achieve 24.62\% and 38.88\% success rates on the group \srone($\drd\ge0.5~$\&\&$~\qed\ge0.6$), respectively;
these are 69.79\% and 168.14\% better than the best baseline method $\hiergtog(m)$
(14.50\% in \srone($\drd\ge0.5~$\&\&$~\qed\ge0.6$)), respectively.
With other similarity constraints $\delta$=0.5 and 0.6, we observed the same trend that \pipeline and \pipelineF consistently outperformed two baseline 
methods on the results of group \srone($\drd\ge0.5~$\&\&$~\qed\ge0.6$).
This demonstrates the strong capability of \pipeline and \pipelineF on the multi-property optimization problem.

In addition, Table~\ref{tbl:multi_drd2_pip} shows that \pipeline and \pipelineF also outperformed the two baseline methods $\jtnn(m)$ and $\hiergtog(m)$ 
with a wide margin on the success rates in group $\srone(\drd\geq0.5)$ and $\srone(\qed\geq0.6)$.
With similarity constraint $\delta$=0.4, \pipeline and \pipelineF were able to achieve 54.00\% and 82.25\% success rates on the group $\srone(\qed\ge0.6)$, respectively;
these are 107.69\% and 216.34\% better the best baseline method $\hiergtog(m)$ (26.00\% in \srone($\drd\ge0.5~$\&\&$~\qed\ge0.6$)), respectively.
This demonstrates that \pipeline and \pipelineF have stronger capabilities to optimize molecules under the similarity constraint towards better properties.
Compared to these \rate values on the group in single-property optimization as in Table~\ref{tbl:multi_drd2_pip}
(54.00\% for \pipeline and 82.25\% for \pipelineF on \qed optimization), the smaller success rates in the 
multi-property optimization (24.62\% for \pipeline and 38.88\% for \pipelineF) also indicate that multi-property optimization is a challenging problem.

\subsubsection{Parameters for Reproducibility of Multi-Property Optimization}
\label{appendix:param:multi_drd2qed}
%
We tuned the hyper-parameters of our models with the grid-search algorithms in the parameter spaces as presented
in Table~\ref{tbl:drd_qed_hyper}.
The optimal hyper-parameters were determined according to the success rates \rate of \pipeline in the group
\srone($\drd\ge0.5~$\&\&$~\qed\ge0.6$) under $\delta$=0.4 over the 200 validation molecules.
The optimal dimension of all the hidden layers is 128 and the dimension of the latent embedding \latent is 64 (i.e., 32 for $\latent^-$ and $\latent^+$, respectively).
The optimal iterations of graph message passing \GMPN and tree message passing \TMPN is 4.
The optimal \molmod model has 0.51M parameters and can achieve the best performance with 8 epochs of training.

\section{Additional Discussions}
\label{appendix:discussion}
%
\subsection{Local Greedy Optimization}
\label{appendix:discussion:greedy}
%

A limitation of {\pipeline} is that it 
employs a local greedy optimization strategy: in each iteration, 
the input molecules to \molmod will be optimized to the best, and if the optimized molecules 
do not have better properties, they will not go through additional \molmod iterations. 
Table~\ref{tbl:pipeline2} shows that at \mbox{$\delta$=0.4}, about {4\%} 
of such molecules in total (under ``\#n\%") stop 
in the middle of 5-iteration optimization. 
However, it is possible that such optimized molecules with declined properties might be further optimized 
in the later \molmod iterations. 
Fig.~\ref{fig:example_logp}d in the main manuscript shows an example of such molecules, where 
$\mol^{(2)}_x$ has worse properties than $\mol^{(1)}_x$ after the second 
iteration of optimization, due to the replacement of two chlorine atoms with one fluorine atom which 
decreases \logp but slightly increases SA.
However, $\mol^{(2)}_x$ can be further optimized into $\mol^{(3)}_x$
of better properties in the third iteration, due to the addition of a more hydrophobic chlorophenyl group.
Note that in our experiments (Table~\ref{tbl:pipeline2}), \pipeline always stops if no property improvement
is observed.
Instead, we can exhaustively optimize each molecule through the entire \pipeline and identify the global optimality.  
In \pipelineF, the top results from each iteration, regardless whether their properties are improved or not, will be 
further optimized in the next iteration. Therefore, it mitigates the local optimal issue.

\subsection{Multi-Property Optimization}
\label{appendix:discussion:multi-property}

In addition to partition coefficient,
there are a lot of factors that need to be considered in order to optimize a lead, or in general, 
to develop a molecule into a drug. For example, toxicity and synthesizability are another two important 
factors for a promising drug candidate; potency, metabolism, cell permeability and side effects 
are also important for a successful drug. 
Almost all the existing computational methods can only optimize one single property, 
or a few properties that can be simply (linearly) combined as one objective~\cite{li2018,jin2020multiobj} 
(e.g., \plogp is a linear combination of \logp, synthesis accessibility and ring size; 
in our experiments presented in the Supplementary Information Section~\ref{appendix:multi_drd2_qed}, it is a linear combination of \drd and \qed scores). 
These methods can be used to optimize one property after another (i.e., in a sequential order) 
by training and applying one model for each property of interest.
However, this sequential optimization is typically suboptimal as later optimization on one 
property could alter the property optimized earlier. 
Unfortunately, to develop one computational model 
that can best optimize multiple properties simultaneously toward global optimality is non-trivial~\cite{Marle2004, Nicolaou2013}, 
and represents a challenging future research direction for generative models. 
There are many reasons contributing to the difficulty of multi-property optimization for computational drug
development. 
First of all, it could be not trivial to set up appropriate mathematical objective or loss functions for the machine learning process.  
Existing methods use simple combinations (e.g., linear) of multiple objective or loss functions, each with respect to
one specific desired drug property. Such simple combinations (e.g., via arithmetic mean or geometric mean) may 
already impose biases on the multiple properties to be optimized. For example, in \plogp, the implicit bias is that 
hydrophobicity and synthesis accessibility are independent, which may or may not be true~\cite{ertl2009estimation},  
since the estimated hydrophobicity (\logp) tends to be larger on larger molecules but the synthesis accessibility score 
would be penalized by larger molecule size.
Learning with respect to a biased objective or loss function may lead to incorrect solutions. 
Even with the loss function accurately provided, it may still be highly non-trivial to solve or approximate 
the underlying mathematical or machine learning problems given their combinatorial nature. 
It is very likely the loss function is non-convex and its components 
cannot be easily decoupled, and therefore, it is very challenging to utilize existing learning strategies 
(e.g., alternating minimization~\cite{boyd2011}, teacher's forcing~\cite{Williams1989}). Meanwhile, 
it is in general very challenging to develop new, effective learning algorithms for multi-objective optimization~\cite{Coello2007}. 
Limited training data is another barrier. To train a truly multi-property optimization model, which is very likely more 
complex than a single-property optimization model, molecules 
satisfying concurrent constraints on multiple properties are needed. This would lead to a small set of eligible molecules
to train a complex model, and thus the model cannot be well trained.

\subsection{Target-Specific Molecule Optimization}
\label{appendix:discussion:target}
Most of the existing molecule optimization methods focus 
on optimizing properties that are general for all the drug candidates, regardless of 
the protein targets that these drug candidates are desired to bind to. 
There are quite a few such
common properties that successful drugs need to exhibit, 
such as high solubility, low toxicity and small size, which ensure the general usability of the existing 
molecule optimization methods. 
However, it is also important that molecule optimization can be tailored to a given protein target, pathway or 
disease. 
There exist some studies on molecule optimization with respect to Dopamine receptor D2 
(\drd)~\cite{jin2019learning,jin2020hierarchical} and Epidermal Growth Factor Receptor (EGFR)~\cite{winter2019}, but they do not 
consider the target information in the modeling, and also are limited by a small number of training molecules. 
Our methods have demonstrated good performance on optimizing with respect to \drd (Supplementary Information 
Section~\ref{appendix:drd2_qed_result}), but the limitation is that they cannot be easily adapted to incorporate additional information 
from protein targets, or in general any information rather than molecule structures
that could be helpful for drug development purposes (e.g., dose response, toxicity profiles).
To optimize a molecule with respect to a certain protein target via deep learning methods, 
new techniques are needed to deal with the target information (e.g., 3D structures of the binding pockets) and 
modify molecule function groups (and molecule 3D structures) accordingly, particularly when the training data 
(e.g., known, accurate binding conformations and binding pocket structures) are much less than what we have for 
general molecule property optimization.

\subsection{Optimization in Other Areas}
\label{appendix:discussion:other_areas}
%
In addition for drug development, the \molmod framework could also be used for compounds or substance property
optimization in other application areas, such as band gaps for solar cells, 
organic light emitting diodes and dyes~\cite{Lile2020}, melting or boiling points for volatiles, materials and plastics~\cite{Sivaraman2020}, and solubility for batteries~\cite{Sorkun2019}.
This could be done by replacing the properties and their measurements of molecules in the loss function 
of \molmod with those of compounds or substances in the respective application areas. 
Given the fact that \molmod is data-driven, 
learns only from pairs of molecules which have difference in the interested properties, 
and does not require any application or domain-specific knowledge,
it is flexible enough to be adapted to other applications where compound or substance properties can be measured and 
compared. 

\section{Algorithms of \molmod}
\label{appendix:alg}

Algorithm~\ref{alg:encoding} describes the encoding process of \molmod. 
\encoder takes a pair of molecules as input and encodes the difference of the molecules into vector $\latent_{xy}$.
Algorithm~\ref{alg:decoding} describes the decoding process of \molmod. \decoder modifies \molx into \moly 
according to the latent difference vector $\latent_{xy}$.
Specifically, in the modification procedure, \decoder first identifies the scaffolds of molecules 
which should be retained after the optimization, and 
removes fragments that are not in the scaffolds to get the intermediate representation.
\decoder then modifies the intermediate molecule representation $\mol^*$ into 
\moly by sequentially attaching new nodes to $\tree^*$ in a breadth-first order as in Algorithm~\ref{alg:nfa}.
%
Algorithm~\ref{alg:pipeline} describes \pipeline optimization. 
Given a molecule \molx, similarity constraint $\delta$, the maximal number of samplings $K$, 
and the maximum number of iterations allowed maxIters, 
our \pipeline iteratively optimizes \molx into a new molecule \moly with better property ($\plogp(\mol_y)>\plogp(\mol_x)$) 
under the similarity constraint ($\text{sim}(\molx,\moly)\geq\delta$).
%
%
{%
Algorithm~{\ref{alg:pipelineF}} describes {\pipelineF} optimization. 
Given a molecule {\molx}, similarity constraint $\delta$, the maximal number of samplings $K$, 
the maximum number of iterations allowed maxIters, the maximum number of input molecules in each iteration $m$
and the maximum number of output molecules of {\pipelineF}, in the $t$-th iteration,
our {\pipelineF} optimizes each input molecule, denoted as {$\mol_{x}^{(t)}(i)$}, into $K$ decoded molecules, denoted as 
 \mbox{$\{\moly^{(t)}(i,k)|k=1, \cdots, K\}$} 
under the similarity constraint.
{\pipelineF} then selects no more than $m$ unique best molecules from \mbox{$\cup_i\{\moly^{(t)}(i, k)|k=1, \cdots, K\}$} for further
 optimization in the next $t$+1 iteration.
The best $b$ molecules among all decoded molecules \mbox{$\cup_t\cup_i\{\moly^{(t)}(i, k)|k=1, \cdots, K\}$} 
will be returned as the final output of {\pipelineF}.}
%

%
\begin{algorithm}
	\caption{\encoder}
	\label{alg:encoding}
	\begin{algorithmic}[1]
		\Require \mbox{\molx = (\graphx, \treex)}, \mbox{\moly = (\graphy, \treey), \nodeb} 
		\vspace{0.3em}
		\LineComment{Atom embedding}
		\State  $\{\atomEmb_x\}$ = $\GMPN(\graphx)$
		\Statex $\{\atomEmb_y\}$ = $\GMPN(\graphy)$ 
		%
		\vspace{0.3em}
		\LineComment{Node embedding}
		\State $\{\nodeEmb_x\}$ = $\TMPN(\treex, \{\atomEmb_x\})$
		\Statex $\{\nodeEmb_y\}$ = $\TMPN(\treey, \{\atomEmb_y\})$ 
		%
		\vspace{0.3em}
		\LineComment{Difference embedding}
		\State $\latent_{xy}$ = $\DE(\treex, \{\nodeEmb_x\}, \treey, \{\nodeEmb_y\}, \nodeb)$ 
		\State \Return $\latent_{xy}$
	\end{algorithmic}
\end{algorithm}


\begin{algorithm}
	\caption{\decoder}
	\label{alg:decoding}
	\begin{algorithmic}[1]
		\Require \mbox{\molx = (\graphx, \treex)}, $\latent_{xy}=[\latent_{xy}^\add, \latent_{xy}^\delete]$
		%
		\vspace{0.3em}
		\LineComment{Disconnection site prediction}
		\State \nodeb = \bpp(\treex, $\latent_{xy}$) 
		\vspace{0.3em}
		\LineComment{Removal fragment prediction}	
		\State $\node_r$ = $\rfp(\{\nodeEmbu |\edge_{ub} \in \vertices_x\}, \latent_{xy}^{\delete})$ 
		\vspace{0.3em}
		\LineComment{Intermediate representation}
		\State $\mol^*$ = $\imr(\node_r, \molx)$
		\vspace{0.3em}
		\LineComment{New fragment attachment} 
		\State $\moly$ = $\nfa(\mol^*, \nodeb, \latent_{xy}^\add)$ 
		\vspace{0.3em}
		\State \Return \moly
	\end{algorithmic}
\end{algorithm}

\begin{algorithm}[!h]
	\caption{\molmod New Fragment Attacher \nfa}
	\label{alg:nfa}
	\begin{algorithmic}[1]
		\Require $\mol^*=(\graph^*, \tree^*)$, \nodeb, $\latent_{xy}^\add$
		\State $t = 0$
		\State {$\mol^{*(0)}=\mol^*$}
		\State $Q$ = emptyQueue()
		\State $Q$.push(\nodeb)
		\vspace{0.3em}
		\While{!Q.isEmpty()}
		\State $\node^{*(t)}$ = $Q$.pop()
		\vspace{0.3em}
		\ShortLineComment{Child prediction}
		\While{\cp($\node^{*(t)}$, $\latent_{xy}^\add$,{$\mol^{*(t)}$})}                   
		\vspace{0.3em}
		\ShortShortLineComment{	Node type prediction}
		\State ($\node_c$, $\ntype_c$)= $\ntp(\node^{*(t)}, \latent_{xy}^\add, \mol^{*(t)})$	 
		%
		\vspace{0.3em}	
		\ShortShortLineComment{Attachment point prediction}
		\State $(\atom_p^*, \atom_c^*)$ = $\app(\node^{*(t)}, \node_c,  \graph^{*(t)}, \latent_{xy}^\add)$
		\vspace{0.3em}	
		\ShortShortLineComment{Attach $\node_c$ to $\mol^{*(t)}$ at $\node^{*(t)}$ }
		\State $\mol^{*(t)}$ = attach($\mol^{*(t)}$, $\node^{*(t)}$, $\atom_p^*$, $\node_c$, $\atom_c^*$)
		%
		\vspace{0.3em}	
		\ShortShortLineComment{Breath-first child expansion}
		\State Q.push($\node_c$)
		%
		%
		\EndWhile
		\vspace{0.3em}
		\State $\mol^{*(t+1)}$ = $\mol^{*(t)}$
		\State $t=t+1$
		
		\EndWhile
		\State \Return $\mol^{*(t)}$
	\end{algorithmic}
\end{algorithm}
%

\begin{algorithm}[!h]
	\caption{Molecule Optimization via \pipeline}
	\label{alg:pipeline}
	\begin{algorithmic}[1]
		\Require \molx, $\delta$, $K$, maxIters, {{\prop}}
		\State $\mol_y^{(0)} = \molx$
		\vspace{0.3em}
		\For{$t=$ 1 to maxIters}
		\vspace{0.3em}
		\ShortLineComment{input molecule to the $t$-th \molmod module}
		\State $\mol_x^{(t)}=\mol_y^{(t-1)}$
		\vspace{0.3em}
		\ShortLineComment{best decoded molecule from this module}
		\State $\mol_y^{\star(t)}$ = $\mol_x^{(t)}$
		\vspace{0.3em}
		\ShortLineComment{multiple samplings and decoding}
		\For{$k=$ 0 to $K$}
		\State $\latent^{(k)} $ = sample from \latent
		\State $\mol = \decoder(\mol_x^{(t)}, \latent^{(k)})$
		\vspace{0.3em}
		\ShortShortLineComment{the best decoded molecule under constraints}
		\If{{$\prop$}$(\mol)\!\!>${$\prop$}$(\mol_y^{\star(t)})$ and \\
			~~~~~~~~~~~~~~~~$\text{sim}(\mol, \mol_x)${$\geq$}$\delta$}  
		\State $\mol_y^{\star(t)}=\mol$
		\EndIf
		\vspace{0.3em}
		\EndFor
		\vspace{0.3em}
		\If{isSame($\mol_y^{\star(t)}, \mol_x^{(t)}$)}
		\ShortShortLineComment{no more optimization and \pipeline stops}
		\State $t = t - 1$
		\State \textbf{break}
		\Else
		\ShortShortLineComment{output molecule from the $t$-th module}
		\State $\mol_y^{(t)} = \mol_y^{\star(t)}$
		\EndIf
		\vspace{0.3em}
		\EndFor
		
		\State $\moly=\mol_y^{(t)}$
		\State \Return \moly
	\end{algorithmic}
\end{algorithm}

\begin{algorithm}[!h]
	\caption{{Molecule Optimization via \pipelineF}}
	\label{alg:pipelineF}
	\begin{algorithmic}[1]
		\Require \molx, $\delta$, $K$, maxIters, $m$, $b$
		\State $\{\mol_y^{\star(0)}\} = \{\molx\}$
		
		\vspace{0.3em}
		\For{$t=$ 1 to maxIters}
		\vspace{0.3em}
		\ShortLineComment{input $m$ molecules to the $t$-th \molmod module}
		\State $\{\mol_x^{(t)}(i)|i=1,...,m\}=\{\mol_y^{\star(t-1)}\}$
		%
		%
		\vspace{0.3em}
		\For{$i=$ 0 to $m$-1}
		\vspace{0.3em}
		\ShortShortLineComment{multiple samplings and decoding}
		\For{$k=$ 0 to $K$}
		\State sample \latent from latent space
		\vspace{0.3em}
		\State $\mol_y^{(t)}(i,k)\!\!=\!\!\decoder(\mol_{x}^{(t)}(i), \latent)$
		\ShortShortShortLineComment{the decoded molecule is unique and similar}
		\If{$\text{sim}(\mol_y^{(t)}(i,k), \mol_x)\!\!\geq\!\delta$}
		\State add $\mol_y^{(t)}(i,k)$ into $\{\mol_y^{(t)}(i,k)\}$
		\EndIf
		%
		%
		\vspace{0.3em}
		\EndFor
		\EndFor
		\vspace{0.3em}
		\ShortLineComment{select top-$m$ molecules from $\cup_i\{\moly^{(t)}(i, k)\}$}
		\State $\{\mol_y^{\star(t)}\}\!\!=\!\text{top}(\cup_i\{\moly^{(t)}(i, k)\}, m)$
		\vspace{0.3em}
		\EndFor
		\vspace{0.3em}
		\LineComment{select $b$ best molecules from $\cup_t\cup_i\{\mol_y^{(t)}(i, k)\}$}
		\State $\{\mol_y^{\star}\}\!\!=\!\text{top}(\cup_t\cup_i\{\mol_y^{(t)}(i, k)\}, b)$
		\State \Return  $\{\mol_y^{\star}\}$
	\end{algorithmic}
\end{algorithm}

\clearpage
\bibliographystyle{naturemag}
\bibliography{abbreviated}


\end{document}